\documentclass[10pt]{article} 
\usepackage[preprint]{tmlr}

\usepackage{amsmath,amsfonts,bm}

\def\eqref#1{equation~\ref{#1}}

\def\1{\bm{1}}

\DeclareMathAlphabet{\mathsfit}{\encodingdefault}{\sfdefault}{m}{sl}
\SetMathAlphabet{\mathsfit}{bold}{\encodingdefault}{\sfdefault}{bx}{n}

\usepackage[colorlinks, citecolor=teal, backref=page]{hyperref}
\usepackage{url}

\usepackage{xcolor}
\usepackage{amsmath,amsfonts,amsthm,amssymb,mathtools}

\usepackage{subfigure} 
\usepackage{multirow} 
\usepackage{xfrac}

\usepackage{algorithm}
\let\AND\relax
\usepackage{algorithmic}
\theoremstyle{plain}
\newtheorem{theorem}{Theorem}[section]

\theoremstyle{definition}

\theoremstyle{remark}

\title{FourStat: A Dual AutoEncoder Leveraging Frequency and Stationarity for Time Series Anomaly Detection}

\author{\name Saman Hadian \email saman.hadian@sharif.edu \\
      \addr Department of Computer Engineering\\
      Sharif University of Technology
      \AND
      \name Ramtin Moslemi \email ramtin.moslemi@sharif.edu \\
      \addr Department of Computer Engineering\\
      Sharif University of Technology
      \AND
      \name Mohammad Mohammadi \email mohammad.mohammadi97@sharif.edu \\
      \addr Department of Mathematical Sciences\\
      Sharif University of Technology
      \AND
      \name Sarina Heshmati \email sarinaheshmati@aut.ac.ir \\
      \addr Department of Mathematics and Computer Science\\
      Amirkabir University of Technology
      \AND
      \name Siavash Ahmadi \email s.ahmadi@sharif.edu\\
      \addr Department of Electrical Engineering \\
      Sharif University of Technology
      \AND
      \name Babak Hossein Khalaj \email khalaj@sharif.edu\\
      \addr Department of Electrical Engineering \\
      Sharif University of Technology
      \AND
      \name Mohammad Hossein Rohban \email rohban@sharif.edu\\
      \addr Department of Computer Engineering \\
      Sharif University of Technology
      }

\def\month{MM}  
\def\year{YYYY} 
\def\openreview{\url{https://openreview.net/forum?id=XXXX}} 

\begin{document}

    \maketitle
    
    \begin{abstract}

    While numerous studies have focused on multivariate time series anomaly detection, many existing methods perform poorly or are not directly applicable to univariate settings.
    We introduce FourStat, a dual-model framework for univariate time series anomaly detection that combines two complementary approaches. 
    The Fourier Autoencoder (FAE) captures global patterns like seasonality through frequency-domain analysis, while the Stationary Autoencoder (SAE) focuses on local structures by enforcing statistical stationarity. 
    We also propose a new evaluation metric called Perfect Point that assesses whether perfect F1-scores can be achieved directly from anomaly scores, eliminating thresholding subjectivity. 
    Results demonstrate that FourStat outperforms existing baselines, particularly on challenging anomaly types, while the new metric provides clearer insights into model performance gaps and advantages.

\end{abstract}

    \section{Introduction}\label{sec:intro}

 Time series anomaly detection (TSAD) is of considerable significance given its numerous applications across diverse areas: health monitoring \citep{healthcare}, industrial manufacturing \citep{industrial}, and finance \citep{finance}, to name a few \citep{Review2}.
There are several broad approaches to detect anomalous subsequences in time series. 
Model-based approaches commonly use forecasting or reconstruction errors to identify deviations from learned temporal behavior \citep{Review}.

Univariate TSAD is challenging because the significance of a pattern often depends on temporal context.
Changes in a series's mean or variance can complicate the learning of normal behavior \citep{blazquez2021review}, while an unusual cycle may only be recognizable when compared with longer-range structure.
A model whose input window is shorter than the relevant temporal pattern may therefore miss contextual anomalies \citep{ermshaus2023window}.
Additionally, the evaluation metrics commonly used in the field are prone to subjective interpretation, underscoring the need for a metric specifically designed for the TSAD problem. 
Furthermore, the field must consider the existence of different anomaly types \citep{IsItWorthIt}. 
This naturally calls for group robustness analysis, as many models may perform well under conventional evaluations but fail to detect certain types of anomalies, which limits their generalizability.

In this paper, we demonstrate the inherent limitations of using relatively small window sizes in TSAD models. 
Building on this insight, we introduce FourStat, a dual-model framework composed of two complementary subnetworks designed to address key challenges of univariate TSAD in parallel. 
These subnetworks are the Fourier AutoEncoder (FAE) and the Stationary AutoEncoder (SAE). 
FAE incorporates large window sizes and leverages the Fast Fourier Transform (FFT) to represent the input signal in the frequency-domain. 
This allows it to effectively capture long-range dependencies and periodic patterns characteristic of seasonal components, without losing information from distant time steps. 
To address the limitations of fixed window lengths, we also introduce an adaptive window selection mechanism that adjusts the input window size based on the characteristics of the time series.
In parallel, SAE operates on a shorter input segment and regularizes the mean and standard deviation of its latent representation toward zero and one, respectively.
Its reconstruction contributes to the final portion of the FAE reconstruction, providing a complementary view of local structure.

Additionally, beyond evaluating our method using conventional metrics, we introduce a group-based evaluation strategy that emphasizes robustness across different types of anomalies. 
Since each anomaly type has distinct characteristics, many models struggle to detect certain groups due to a mismatch between the anomaly’s nature and the model’s architecture. 
We also propose Perfect Point, a binary metric that tests whether the anomaly scores admit a threshold with complete event recall and no point-wise false positives.
It evaluates score separability using ground-truth labels without requiring the researcher to choose an operating threshold; it does not solve the problem of selecting a threshold in deployment.
Together with results grouped by anomaly type, this provides a complementary view of performance under our univariate evaluation protocol.

Our main contributions are listed below:

\begin{enumerate}
    \item We demonstrate the limitations of small and fixed window sizes in TSAD models, particularly in capturing long-range dependencies and seasonal patterns, and propose an adaptive window selection mechanism to address these limitations.
    \item We introduce \textbf{FourStat}, a dual-model framework that combines FAE and SAE to jointly learn the trend and seasonal components of time series more effectively.
    \item We design a composite loss function that combines Mean Squared Error (MSE) and Relative Squared Error (RSE), and we provide theoretical justification for its effectiveness in improving reconstruction-based anomaly detection.
    \item We introduce \textbf{Perfect Point}, a threshold-free evaluation metric, along with a group-based evaluation strategy to better assess robustness across different anomaly types and reduce reliance on subjective threshold selection.
\end{enumerate}

\section{Contextual Blindness of Small Windows}\label{sec:theory}

\begin{figure}[ht]
    \centering
    \subfigure[Missing Peak]{\label{fig:missing_peak}\includegraphics[width=0.48\textwidth]{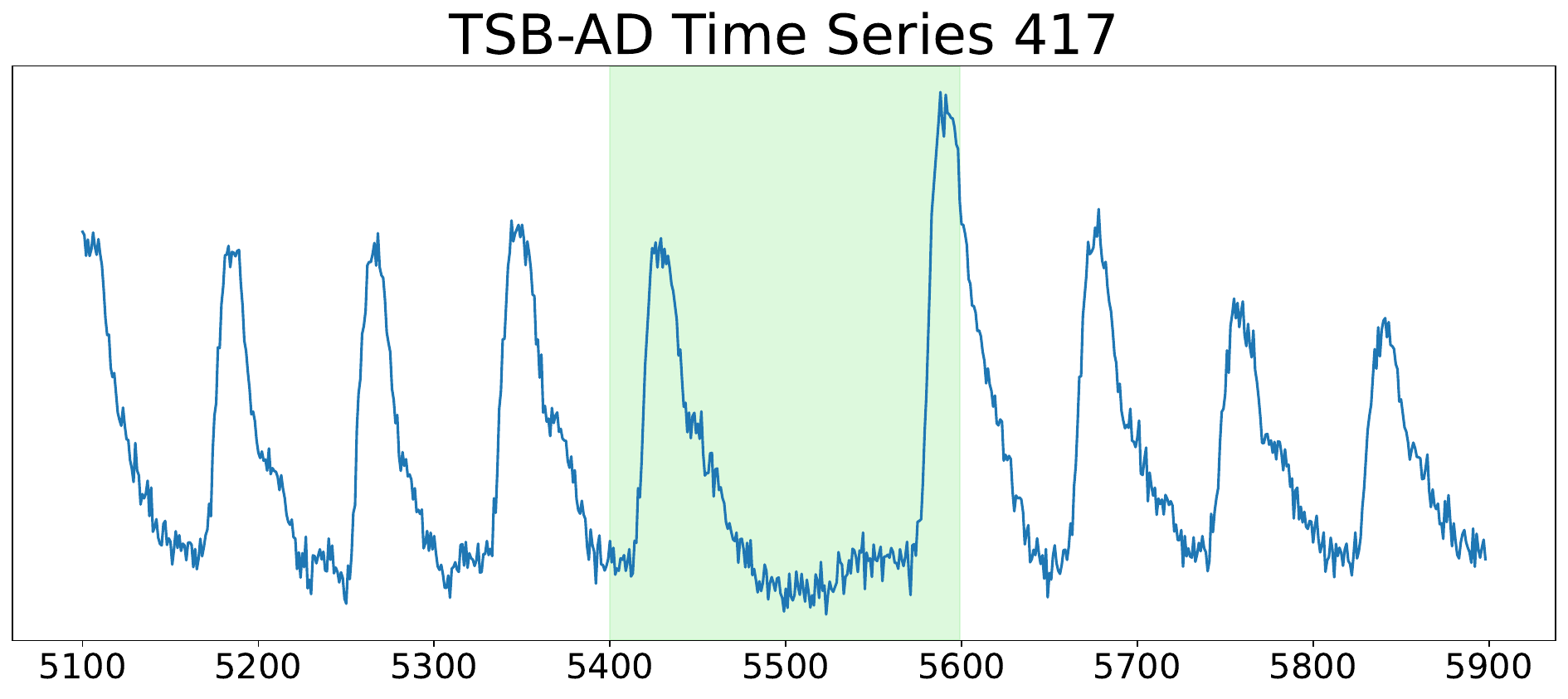}}
    \subfigure[Time Shift]{\label{fig:time_shift}\includegraphics[width=0.48\textwidth]{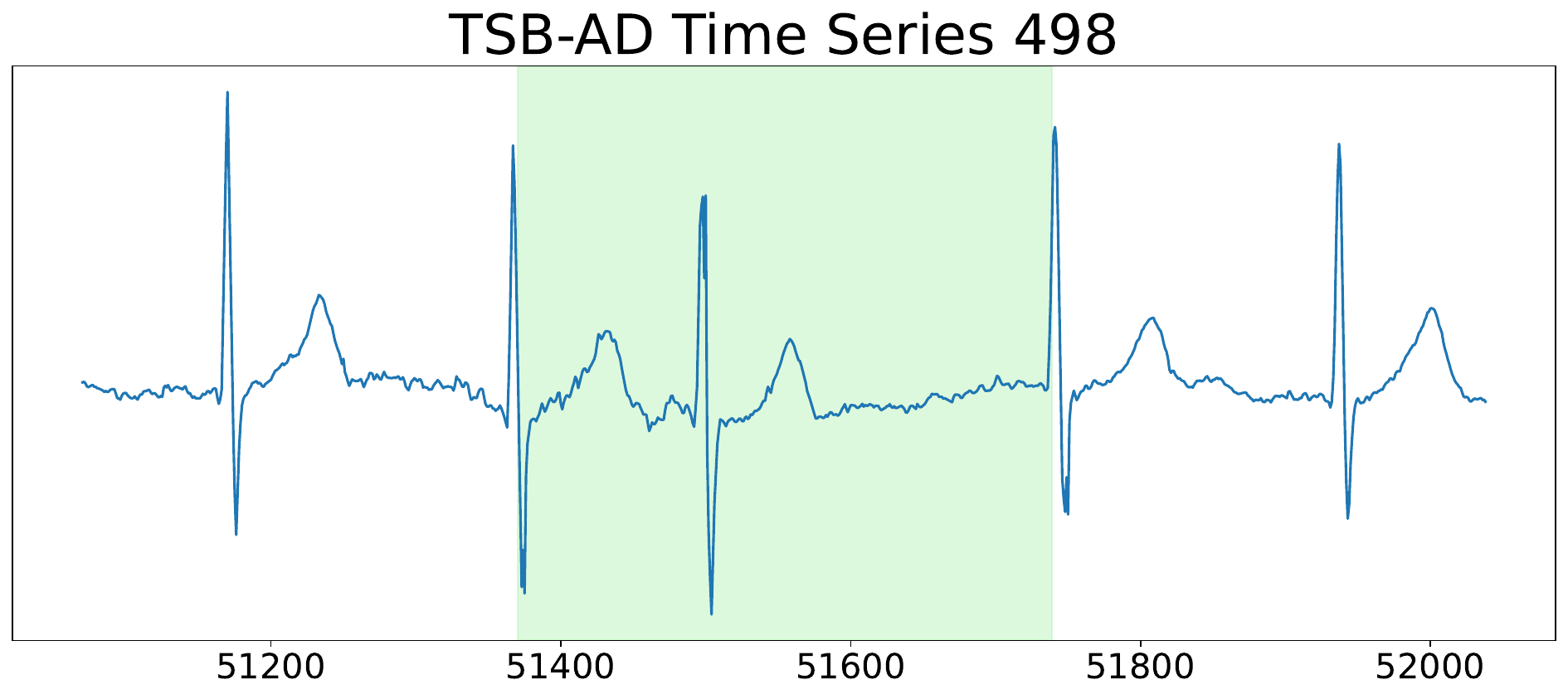}}
    \caption{
    Two common anomaly types where small windows falter.
    Anomalous regions have been highlighted.
    In \subref{fig:missing_peak} a peak was removed and in \subref{fig:time_shift} the pause between the two peaks was increased.
    Small windows in these regions are not anomalies themselves and the anomalous patterns emerge only when we use sufficiently large windows that cover full cycles.}
    \label{fig:small_window}
\end{figure}

\paragraph{Small windows fail to capture sufficient context.}
Most TSAD methods operate on small windows, typically segmenting long sequences into smaller ones. 
While this approach is widely adopted, particularly in reconstruction-based methods that assess the normality of each window independently, the choice of window size is critical. 
It is common in prior work to use relatively small window sizes for evaluation \citep{DAGMM, MAD-GAN, USAD, CAE-M}, yet we argue that this practice is often inadequate. 
Many existing methods struggle with limited window lengths, as smaller windows may fail to capture sufficient contextual information necessary for accurate detection \citep{ermshaus2023window}. 
This issue is especially pronounced in time series with inherent long seasonality, and when the window size is shorter than the seasonal cycle, it becomes fundamentally incapable of representing the structure needed to identify such anomalies.
Illustrative cases of this limitation are shown in Figure~\ref{fig:small_window}, where anomalies are effectively undetectable by models operating on narrow temporal windows.

\paragraph{Fixed window sizes are inherently limited.}
While using larger windows may alleviate these issues, relying on a fixed large window is often suboptimal. 
Different anomalies, particularly those driven by seasonality, rely on different amounts of temporal context. 
As a result, a single, manually chosen window size cannot adequately capture all relevant behaviors.
Beyond this, the reliance on small windows also raises concerns about the group robustness of detection methods. 
In many benchmark datasets, most anomalies belong to a single dominant type. 
As a result, models with limited temporal context may appear to perform well overall, while in reality failing to detect other types of anomalies that require broader contextual understanding. 
This lack of robustness across anomaly groups is particularly problematic in real-world deployments, where anomaly types are typically more diverse. 
These observations highlight that window size must be chosen in a data-dependent manner rather than fixed globally.
While fixed window sizes may perform adequately on individual datasets, they fundamentally limit the generalizability of TSAD models across diverse anomaly types, a behavior that is clearly reflected in our experimental results in Section~\ref{sec:experiments}.

\paragraph{Large windows introduce architectural and computational challenges.}
These limitations do not imply that long context is unnecessary; rather, they indicate that long windows should be used selectively, only when the underlying structure demands it.
Despite the intuitive appeal of larger windows for capturing longer temporal dependencies, they have historically been avoided in TSAD for several reasons.
First, recurrent architectures such as LSTM exhibit well-known practical limitations when applied to very large input windows (e.g., beyond 100 time steps), where vanishing gradients and degraded effective memory reduce performance \citep{NOLSTM}. 
It is important to emphasize that we do not argue that LSTM-based models are inherently unsuitable for temporal modeling; in fact, we also employ LSTM components within our architecture, where they are applied to shorter input windows for which they remain effective and computationally efficient.
Moreover, training efficiency is compromised as large windows drastically increase computational costs and reduce the number of unique training samples, especially when data is scarce. 
Transformer-based models, while powerful, are known to be data-hungry and therefore not suitable for low-data regimes \citep{hungrytransformer}. 
Fully connected networks also fail to generalize well due to their inappropriate inductive bias in handling sequential data.

Another critical issue arises in point-wise anomaly detection.
When using large windows, a single anomalous point can be \textit{diluted} within mostly normal segments, making detection through window-based scoring less effective.
Rather than addressing this issue directly within large-window models, our architecture separates global and local modeling.
Global temporal structure is captured using large windows in the Fourier AutoEncoder (FAE), while sensitivity to localized deviations is handled by a dedicated Stationary AutoEncoder (SAE) operating on short temporal contexts.
The SAE is trained using the \textbf{MaxDiff} loss introduced in Equation~\ref{eq:maxdiff}, inspired by localization techniques in image anomaly detection \citep{MaxDiff-SimpleNet}, which enhances contrast between normal and anomalous local patterns.
This design enables effective point-wise anomaly detection without relying on large-window localization.
    \section{Our Approach}\label{sec:method}

\begin{figure*}[ht]
    \centering
    \includegraphics[width=\textwidth]{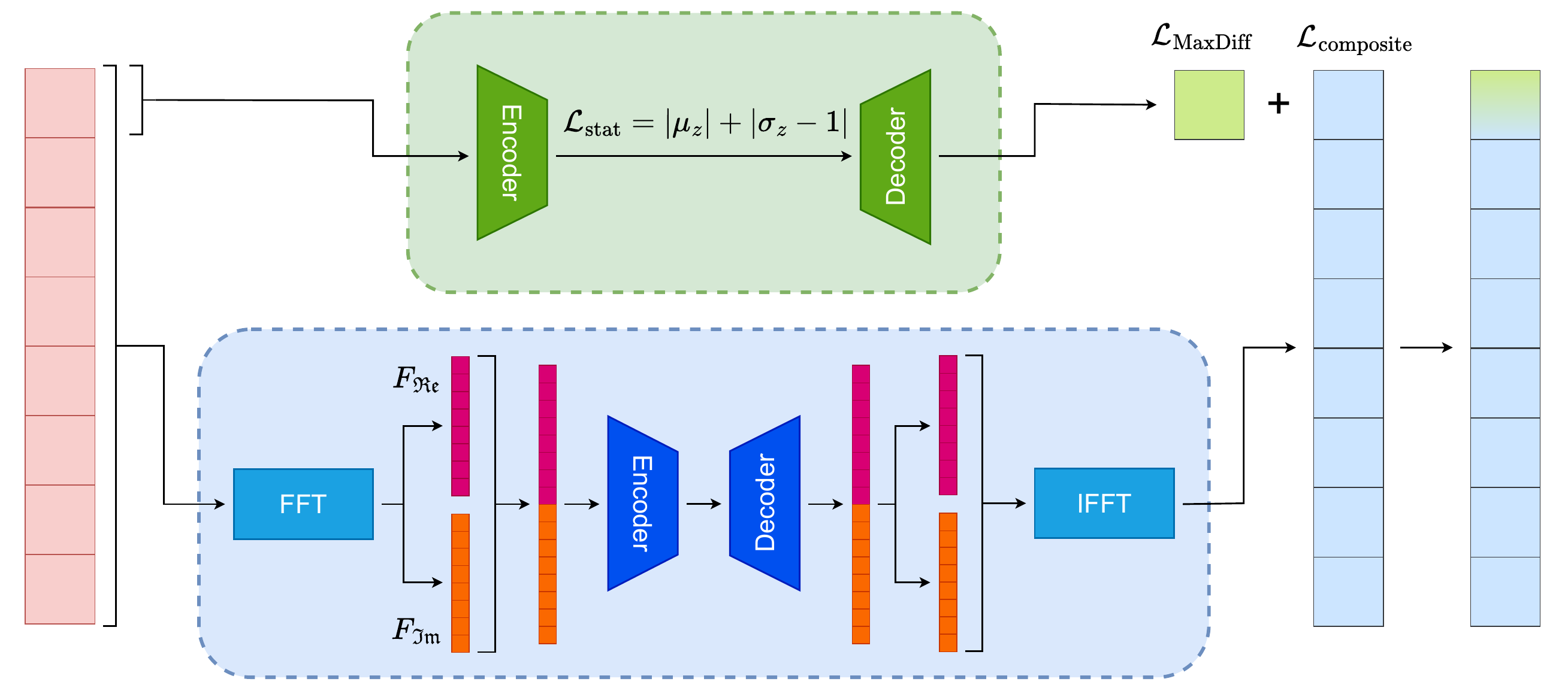}
    \caption{\textbf{The FourStat Architecture.} The FAE operates on the entire input window, applying the FFT to obtain a complex-valued vector $F'$. By concatenating the real and imaginary components of $F'$, we construct a real-valued vector $F$, which is then passed through a fully connected AE to produce the latent representation $Z$. To reconstruct the original signal via the IFFT, $Z$ is split into two equal-length sequences representing the real and imaginary parts, forming the complex vector $Z'$, which yields the reconstructed signal $X'$ after applying the IFFT.
The SAE, on the other hand, processes the final 32 values of the input window using an LSTM-based AE to produce the reconstruction $Y'$. The stationarity loss, defined in Equation~\ref{eq:stat-loss}, is imposed on its latent representation to enhance temporal consistency. 
The final reconstruction is formed by adding the SAE reconstruction to the corresponding final 32 time steps of the FAE output, while the remaining time steps are taken directly from $X'$.
}
    \label{fig:fourstat}
\end{figure*}

Given that time series data can be naturally decomposed into trend, seasonality, and noise components, we propose FourStat, as depicted in Figure~\ref{fig:fourstat}, a modular architecture in which each sub-network is dedicated to modeling a specific component, excluding the noise. 
To capture the periodic structure of the seasonal component, we employ FAE, which operates on large temporal windows and leverages frequency-domain representations, making it well-suited for learning recurring global patterns.
For modeling long-term trends and sequential dependencies, we incorporate a recurrent network based on Long Short-Term Memory (LSTM) \citep{LSTM} units. 
To capture localized structures and transient patterns, we include SAE, which operates on smaller windows and learns statistically stable representations of short-term dynamics.

\paragraph{Adaptive context selection (window size).}
Before training the full FourStat architecture, we determine an appropriate temporal window length for each time series using an efficient adaptive context selection procedure.
While an exhaustive sweep over candidate window sizes is possible, it is computationally expensive and unnecessary.
Instead, we leverage the FAE as a lightweight proxy model. 
A standalone FAE is briefly trained for a small update budget on different candidate window sizes, and the behavior of its reconstruction loss is monitored.

Due to the frequency-domain representation induced by the FFT, this proxy converges sufficiently fast to reveal whether a given window provides adequate temporal context.
We perform a coarse-to-fine search over window sizes, in a manner analogous to binary search, starting from a large candidate window and progressively shrinking it.
The selected window corresponds to the smallest window for which the proxy reconstruction loss exhibits a clear early drop followed by stabilization.
Intuitively, windows shorter than this point lack sufficient context, while larger windows yield diminishing returns and increase computational cost and anomaly dilution.
The resulting window size is then fixed and used by the full FourStat model for training and inference.
This adaptive selection avoids the inherent trade-off of fixed window designs discussed in Section~\ref{sec:theory}, while keeping the overall training cost modest.

\subsection{Fourier AutoEncoder}\label{sec:fae}
The Fourier AutoEncoder (FAE) is designed to process larger temporal windows, enabling it to effectively capture long-range dependencies and recurring seasonal patterns. 
Although processing larger windows increases sample complexity, the FAE benefits from the global view provided by frequency-domain representations, allowing it to learn periodic structures more efficiently than sequential models like LSTMs, which are inherently limited by their recursive nature \citep{yi2023frequency}.

The FAE operates by first applying the Fast Fourier Transform (FFT) to the input time series, transforming it from the time-domain to the frequency-domain. 
Since FFT outputs are complex-valued, we separate the real and imaginary parts and represent them as two separate real-valued vectors of equal length. 
These vectors are concatenated to form a single input vector, which is then passed through a standard AutoEncoder architecture composed of fully connected layers. 
This bottlenecked representation allows the model to learn a compressed frequency-domain encoding that emphasizes dominant spectral components while potentially filtering out noise.

The decoder reconstructs the frequency representation by producing an output vector that is again split into two parts, corresponding to the real and imaginary components. 
These are recombined to form a complex-valued output vector, which is subsequently transformed back into the time-domain using the Inverse Fast Fourier Transform (IFFT). 
The result is a time-domain reconstruction of the input signal, ideally preserving the periodic structures while discarding irrelevant fluctuations.

This architecture enables the model to learn a compact, interpretable, and frequency-aware representation of seasonal components in time series data. 
The internal structure of the FAE is illustrated in Figure~\ref{fig:fourstat}.
Furthermore, its reliance on global transformations like the FFT allows the FAE to generalize across windows of varying patterns and frequencies, making it a powerful module for isolating and reconstructing seasonal behavior.
Since we are working in the frequency-domain, we no longer need recurrent network for processing sequences and can instead utilize fully connected layers.

\subsection{Loss functions for capturing seasonality}\label{sec:loss}

Aside from the limitations of large window sizes discussed in Section~\ref{sec:theory}, a notable drawback of using the standard MSE loss in isolation is its tendency to bias the model toward reconstructing low-variance components of the signal, most often manifesting as a flat, mean-valued output.
This is particularly problematic when the underlying time series exhibits meaningful periodic or seasonal structure. 
To address this issue in the FAE, which operates on large temporal windows, we introduce a hybrid loss function, defined in Equation~\ref{eq:composite_loss}, which combines the absolute fidelity of MSE with the scale-awareness of the RSE.

To better understand the intuition behind our proposed loss, consider a simple target function \( s(t) = a \sin(bt)\), and let the model's prediction be \(\hat{s}(t) = x \sin(yt) \) where \( x \) and \( y \) denote the predicted amplitude and frequency, respectively.

\begin{figure*}[t]
    \centering
    \subfigure[MSE Loss Surface]{\label{fig:mse3d_inv}\includegraphics[width=0.32\textwidth]{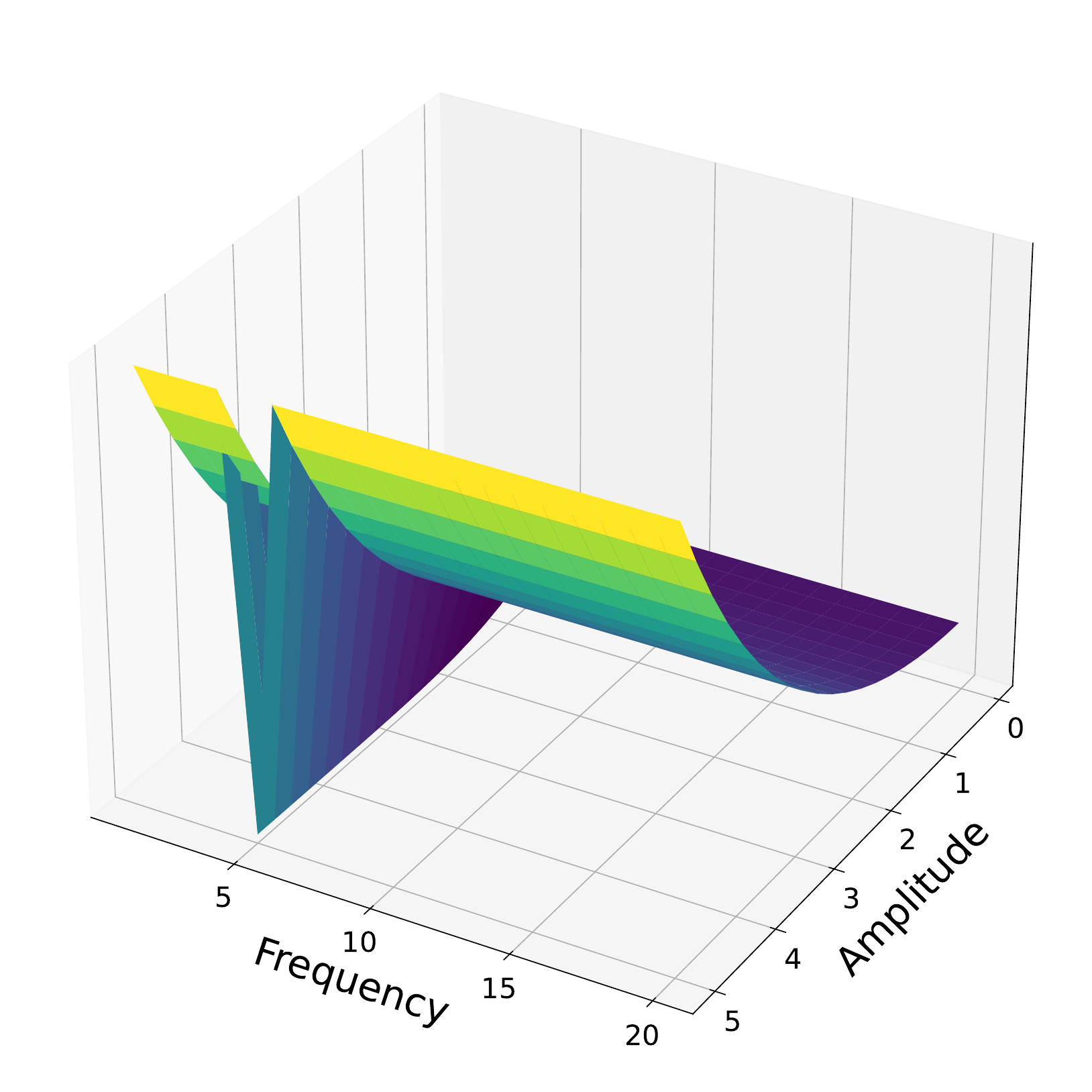}}
    \subfigure[RSE Loss Surface]{\label{fig:rse3d_inv}\includegraphics[width=0.32\textwidth]{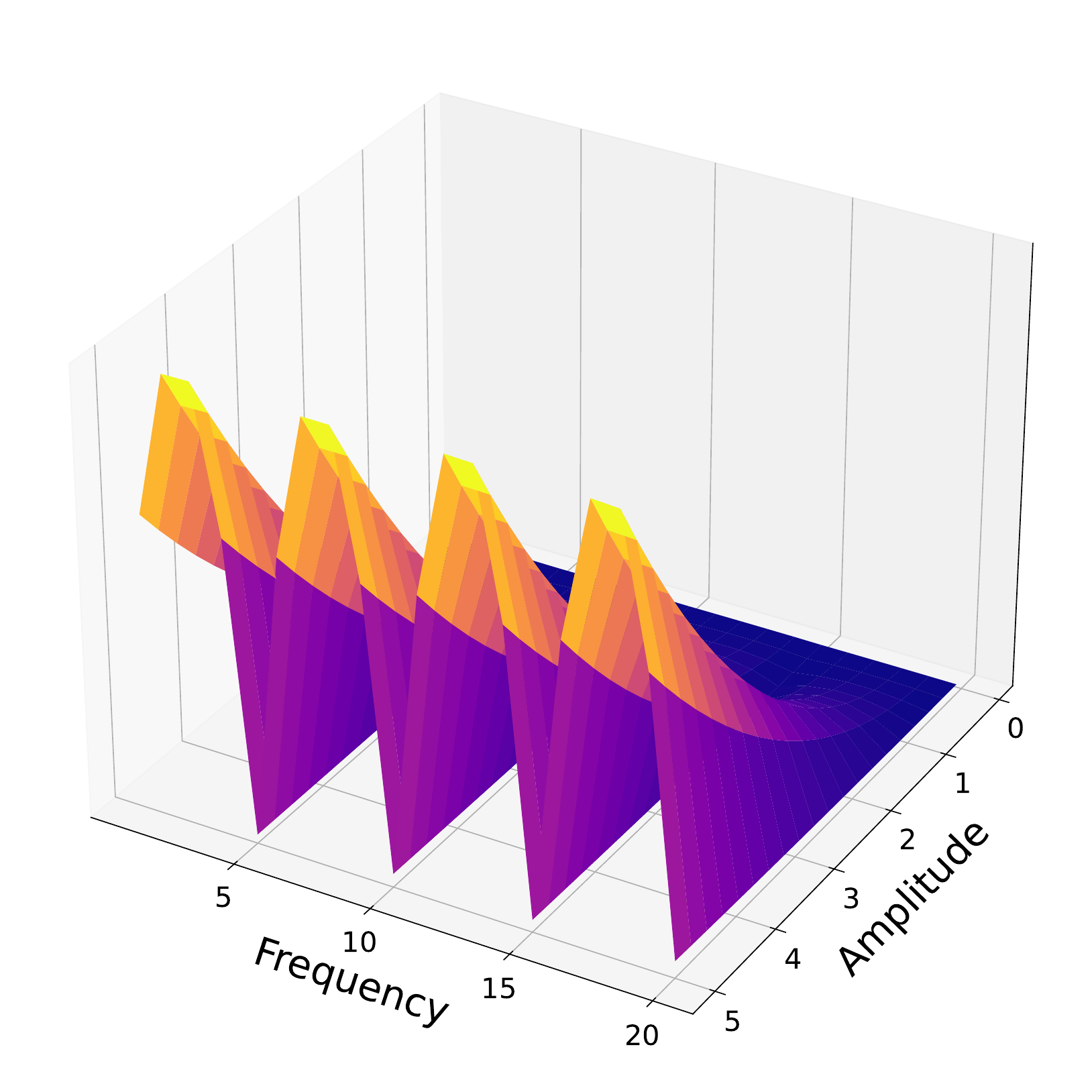}}
    \subfigure[$(\text{MSE} + \text{RSE})/2$ Loss Surface]{\label{fig:mse_rse3d_inv}\includegraphics[width=0.32\textwidth]{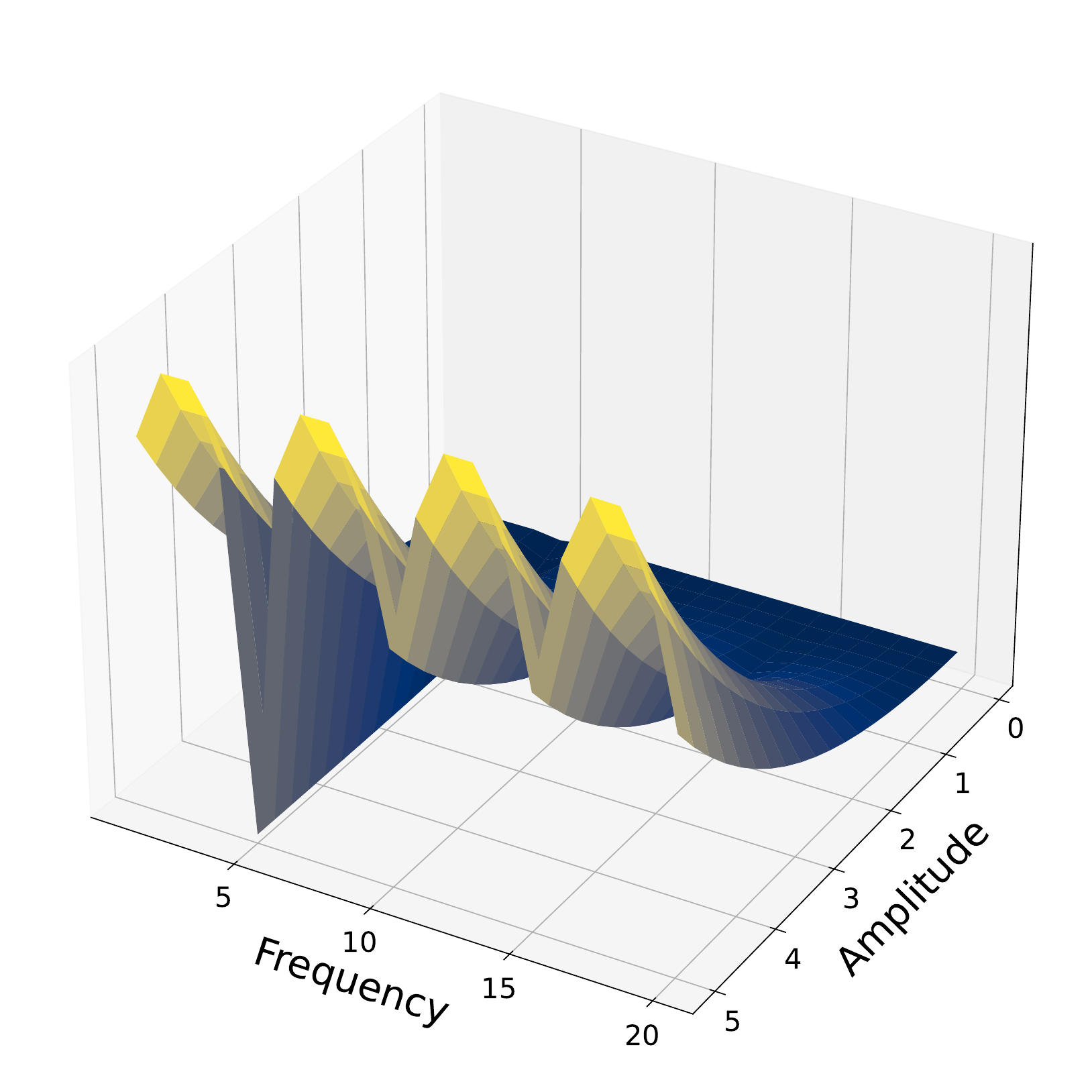}}
    \caption{
    3D surface plots of loss metrics comparing a reference signal \( s(t) = 4\sin(10\pi t) \) with test signals \( \hat{s}(t) = A\sin(2\pi F t) \), where amplitude \( A \) and frequency \( F \) vary over a grid.
    }
    \label{fig:loss_surface_inv}
\end{figure*}

We begin with a visual exploration. 
As illustrated in Figure~\ref{fig:loss_surface_inv}, we plot the loss surfaces of MSE, RSE, and their combination across a grid of amplitude and frequency values. 
The MSE loss surface flattens near zero amplitude, offering no gradient signal when the predicted amplitude \( x \) is close to zero, regardless of the frequency. 
This plateau makes it difficult for optimization to escape from poor initializations. 
The RSE loss surface, in contrast, is more sensitive to relative discrepancies and introduces useful local minima around harmonic frequencies (\( y = k b \), for integer \( k \)), reflecting the nature of Fourier harmonics.
The composite loss yields a more informative and smoother landscape, mitigating the individual shortcomings of its components and enabling more reliable optimization.

These insights are supported by a theoretical analysis of the gradients. When \( x = 0 \), the MSE loss becomes a constant:
\begin{equation}
\mathcal{L}_{\text{MSE}} = \frac{1}{N} \sum_{t=1}^N \left[ a \sin(bt) - x \sin(yt) \right]^2,
\end{equation}
yielding gradients with negligible magnitude with respect to both $x$ and $y$

\begin{equation}
\frac{d}{dx} \mathcal{L}_{\text{MSE}} \approx 0, \quad \frac{d}{dy} \mathcal{L}_{\text{MSE}} = 0.
\end{equation}
In contrast, the RSE loss is defined as:
\begin{equation}
\mathcal{L}_{\text{RSE}} = \frac{1}{N} \sum_{t=1}^N \left( \frac{a \sin(bt) - x \sin(yt)}{a \sin(bt) + \epsilon} \right)^2.
\end{equation}
In this case, the gradients of the RSE loss do not vanish exactly, but can become very small when the predicted signal aligns closely with the target or its harmonics, i.e., when $x \approx a$ and $y \approx kb$.

\begin{equation}
\left| \frac{d}{dx} \mathcal{L}_{\text{RSE}} \right| \ll 1, \quad
\left| \frac{d}{dy} \mathcal{L}_{\text{RSE}} \right| \ll 1.
\end{equation}

To mitigate these individual weaknesses, we propose the following composite loss:
\begin{equation}
\mathcal{L}_{\text{composite}} = \lambda \cdot \mathcal{L}_{\text{MSE}} + (1 - \lambda) \cdot \mathcal{L}_{\text{RSE}},
\label{eq:composite_loss}
\end{equation}
where \( \lambda \in [0, 1] \) balances the trade-off between absolute and relative error sensitivity. Empirically, we find that this hybrid formulation yields better convergence and more reliable reconstructions, especially when recovering oscillatory or seasonal signals.
A full derivation of the gradients and behavior of the loss functions is provided in Appendix~\ref{sec:more-proofs}.

The benefits of this composite loss are not limited to simple sine waves. By the Fourier theorem, any time series signal can be decomposed into a sum of sine and cosine functions. Therefore, capturing the amplitude and frequency components of sinusoidal basis functions is essential for accurately modeling a wide class of signals, especially those exhibiting periodic structure.
In the context of our FAE, applying the composite loss helps the model learn richer frequency components via its latent representation. 
The MSE captures fidelity in time, while the RSE penalizes underestimation of high-amplitude components in a scale-aware fashion.

\subsection{Stationary AutoEncoder}\label{sec:sae}
The Stationary AutoEncoder (SAE) is designed to capture local, short-term structures by operating on smaller input windows, making it well-suited for detecting transient patterns and local anomalies. 
This localized focus complements the broader temporal modeling performed by the FAE and addresses the common assumption in time series analysis that local segments exhibit stationarity, i.e., their statistical properties remain constant over short durations.

In particular, this refers to weak stationarity, where the mean and variance of the segment are constant over time, and the autocovariance function depends only on the time lag between observations rather than their absolute positions in the sequence.

A key innovation in the SAE lies in its dual-loss formulation.
The first loss enforces statistical stationarity by penalizing deviations in the mean and standard deviation of the bottleneck (latent) representation from standardized values (mean = 0, standard deviation = 1), inspired by the definition of weak stationarity and principles from z-score normalization.
Formally, this loss minimizes the absolute difference between the empirical mean and standard deviation of the latent vector and the target values, i.e.,
\begin{equation}\label{eq:stat-loss}
\mathcal{L}_{\text{stat}} = \left| \mu_z \right| + \left| \sigma_z - 1 \right|
\end{equation}
where $\mu_z$ and $\sigma_z$ denote the  mean and standard deviation of the latent representation z.
This encourages the model to focus on learning patterns that are invariant to scale and level shifts.
As a result the encoder learns to remove trend and small seasonality from the inputs and the decoder learns to add them once more.

The second loss is a max difference loss (MaxDiff loss), which captures the largest point-wise deviation between the reconstructed sequence and the original input.
Specifically, it computes the maximum absolute difference across all time steps, emphasizing worst-case reconstruction errors.
Formally, the loss is defined as:
\begin{equation}
\mathcal{L}_{\text{MaxDiff}} = \max_t \left| x_t - \hat{x}_t \right|
\label{eq:maxdiff}
\end{equation}
where $x_t$ is the input signal at time step t, and $\hat{x}_t$ is the corresponding reconstructed value.
Unlike MSE, which averages errors across all time steps and may dilute sharp anomalies, the max difference loss is highly sensitive to localized deviations.
This makes it particularly effective for detecting point-wise anomalies such as sudden spikes or drops and helps localize them even when operating over relatively large window sizes.
Together, these losses guide the SAE to learn stable, localized patterns while remaining sensitive to small scale deviations traits that are particularly useful for anomaly detection in non-seasonal components of the signal.

\section{Perfect Point Evaluation}
\label{sec:pp_eval}

Current research in TSAD faces two key limitations that our work aims to address. 
First, many popular benchmark datasets exhibit a \textbf{high density of anomalies}, which can distort performance evaluations. 
In such cases, models may learn to exploit superficial patterns or dataset-specific biases instead of genuinely understanding the distinction between normal and anomalous behavior. 
This often results in inflated metrics, where even weak models can appear to perform well, misleading both researchers and practitioners. 
Second, the field suffers from \textbf{inconsistent and often unreliable evaluation metrics}. 
Different studies employ a wide variety of metrics and evaluation setups, making it difficult to compare models fairly. 
This issue is similar to the \emph{unit test} problem in other machine learning domains, where overly narrow or brittle evaluation criteria cause results to fluctuate significantly with small changes in setup. 
As a consequence, progress in the field is impeded by a lack of standardized, meaningful performance benchmarks.

\subsection{Existing Metrics}
The \textbf{F1-composite score}, as proposed by~\citet{zhang2024benchmarking}, proves particularly informative and is included in our evaluation protocol. 
We also report the \textbf{Area Under the Precision-Recall Curve (AUC-PR)}, a robust metric for imbalanced datasets that evaluates how well the model’s score prioritizes true anomalies over normal points.
Additionally, we include an \textbf{event-based AUC}, which measures whether anomalous events, not just individual anomalous points, receive higher scores than normal regions. 
In this formulation, an event is considered correctly ranked if at least one of its constituent points scores higher than the highest score in any normal region. 
This event-centric view aligns more closely with real-world applications, where identifying the occurrence of an anomaly is often more critical than detecting every single anomalous point.

\subsection{Moving Beyond Threshold Predictions}
Our goal is to move beyond binary predictions at arbitrary thresholds and instead evaluate the quality of the anomaly score itself. 
A meaningful evaluation should assess whether the model effectively ranks anomalies higher than normal points and whether the score signal captures the underlying structure of anomalous events.

\subsection{The Perfect Point Metric}
We introduce the \textbf{Perfect Point (PP)} metric, which illustrates whether a model can achieve perfect detection performance at any decision threshold. 
Specifically, it asks whether there exists a threshold where the model attains both 100\% precision and 100\% event recall, indicating that a perfect F1-score is achievable at some threshold. 
This implies that all anomaly events are detected without any false positives. 
We record this as a binary metric: a \emph{Yes} indicates that the anomaly score distribution is sufficiently well-separated to allow flawless classification at some operating point.

This threshold-independent, instance-level perspective aligns better with real-world deployment scenarios, where even a single missed anomaly can have critical consequences. 
It also incentivizes models to produce sharp and interpretable anomaly scores, rather than merely optimizing threshold-dependent metrics. 
In Appendix~\ref{sec:ppt}, we further investigate how this metric relates to common thresholding strategies.


\paragraph{Definition and Scope.}
A {Perfect Point} exists if and only if there is at least one decision threshold where \textit{event-based recall} = 100\% and \textit{point-based precision} = 100\%. 
Unlike continuous metrics such as F1 or AUC, PP is a \textbf{binary indicator} that tests whether perfect classification is achievable at any threshold. 
Its threshold-free nature makes it particularly suitable for \textbf{safety-critical domains}, where interpretability and strict guarantees are preferred over average-case scores.

Formally, our definition follows the standard formulations of event-based recall and point-based precision introduced by \citet{zhang2024benchmarking} under the \textit{Detailed Evaluation Metrics} section. 
While these metrics quantify detection quality at fixed thresholds, PP asks a simpler yet more stringent question: 
\begin{quote}
\emph{Does any threshold exist where all anomaly events are detected correctly and no false alarms occur?}
\end{quote}

\paragraph{Motivation.}
The motivation for introducing PP is twofold. 
First, existing F1-based evaluations in TSAD are \textbf{threshold-dependent and inconsistent}. 
Different studies use point-wise F1, adjusted F1, or composite F1 variants, often producing incomparable rankings. 
Second, the F1 score can be artificially inflated even by random or poorly performing models. 
For example, in datasets where anomaly events are long and rare, a random model marking 5\% of points as anomalies could achieve event recall $\approx$ 100\% and point precision $\approx$ 95\%, leading to an apparently high composite F1 of $\sim$97\%. 
However, such a model fails to meaningfully separate anomalies from normal points. 
In contrast, PP would return $0$, as no threshold yields perfect precision and recall simultaneously. 

\paragraph{Conceptual Distinction.}
The PP and F1 metrics serve complementary purposes:
\begin{itemize}
    \item \textbf{F1 Score:} continuous, threshold-sensitive; measures average-case performance at a selected threshold.
    \item \textbf{Perfect Point:} binary, threshold-free; tests for the existence of any perfectly separable decision boundary.
\end{itemize}
This distinction allows PP to act as a \textbf{unit-test-style validation}, a pass/fail diagnostic that reveals whether the model ever reaches flawless discrimination, regardless of threshold tuning. 
This is especially valuable for debugging, reproducibility, and assessing reliability in high-stakes applications.

\paragraph{Statistical Validation.}
To quantify the relationship between PP and model performance, we conducted a \textbf{Mann–Whitney U test} comparing the distributions of F1 scores for the \texttt{PP=True} and \texttt{PP=False} groups. 
The resulting $p$-value was extremely small, indicating a highly significant difference between the two groups. 
This confirms that the presence of a Perfect Point is \textbf{strongly associated with superior F1 performance}, yet not redundant with it. 
Importantly, since PP does not require fixing a decision threshold, it directly addresses the concern raised in the Introduction regarding brittle, threshold-dependent evaluation. 
Full statistical results and corresponding visualizations are included in Appendix~\ref{sec:ppt}.

\paragraph{Robustness and Generalizability.}
To assess generalizability, we further performed \textbf{group robustness analysis} across multiple datasets. 
This analysis confirmed that PP consistently reflects true model quality, even under varying anomaly densities and temporal characteristics. 
Such validation mirrors a \textit{unit-test paradigm}, ensuring consistent metric behavior across experimental conditions.

\paragraph{Practical Advantages.}
Beyond statistical correlation, PP provides practical interpretability benefits:
\begin{itemize}
    \item Identifies misleading cases where F1 appears high but model reconstructions are clearly flawed.
    \item Eliminates dependence on post-hoc threshold tuning and arbitrary decision boundaries.
    \item Encourages the development of sharper, well-separated anomaly score distributions.
\end{itemize}

\paragraph{Conclusion.}
The PP metric complements rather than replaces F1. 
It offers a binary, interpretable, and threshold-free validation mechanism that supports reproducible, unambiguous evaluation. 
By aligning with the paper’s broader goal of moving beyond threshold-dependent metrics, PP provides a principled foundation for fairer and more robust assessment of time series anomaly detection models.

    \section{Experiments}\label{sec:experiments}
We present a comprehensive empirical evaluation of our proposed framework, designed to rigorously assess performance while addressing key limitations in current TSAD research. 
Our experiments are structured to highlight critical gaps in existing benchmarks and metrics, introduce novel evaluation protocols, and provide actionable insights into model behavior across diverse anomaly types.
We break down and discuss the anomaly types in Appendix~\ref{sec:anomaly-types}.
You can find a detailed description of our experimental settings in Appendix~\ref{sec:settings}.

\begin{table*}[t]
    \caption{
        Number of Perfect Points based on anomaly type. 
        The best results are in \textbf{bold} and the second best are \underline{underlined}.
        Forecasting and reconstruction methods are separated with a horizontal line.
    }
    \label{tab:results-by-type}
    \centering
    
    \resizebox{\textwidth}{!}{
    \begin{tabular}{l c c c c c c c c c c c}
        \hline
        \textbf{Model}
        & \rotatebox{0}{Amplitude} 
        & \rotatebox{0}{Flat} 
        & \rotatebox{0}{Frequency}
        & \rotatebox{0}{Inverted}
        & \rotatebox{0}{Local}
        & \rotatebox{0}{Missing}
        & \rotatebox{0}{Noise} 
        & \rotatebox{0}{Outlier} 
        & \rotatebox{0}{Reversed}
        & \rotatebox{0}{Time} 
        & \textbf{All} \\
        \hline
        TFT & \underline{42} & \underline{12} & 3 & 8 & 30 & 2 & 5 & 19 & \underline{10} & 10 & 141 \\
        Informer & 40 & \textbf{17} & \underline{7} & 12 & 31 & \underline{4} & \textbf{12} & \underline{22} & \textbf{13} & \textbf{16} & \underline{174} \\
        Autoformer & 37 & 11 & 4 & \underline{13} & \underline{33} & \underline{4} & \underline{10} & \underline{22} & 8 & \underline{13} & 155 \\
        FEDformer & 22 & 8 & 3 & 11 & 28 & 2 & 8 & 13 & 5 & 8 & 108 \\
        PSformer & \textbf{44} & 8 & \textbf{9} & \textbf{15} & \textbf{34} & \textbf{7} & \textbf{12} & \textbf{27} & \underline{10} & 10 & \textbf{176} \\
        \hline
        DAGMM     & 29 & 9 & 0 & 7 & 27 & 0 & 2 & 13 & 0 & 3 & 90 \\
        MAD-GAN   & 27 & 10 & 0 & 7 & 25 & 0 & 2 & 12 & 0 & 3 & 86 \\
        USAD      & 29 & 9 & 0 & 8 & 27 & 0 & 2 & 13 & 0 & 3 & 91 \\
        CAE-M     & 29 & 9 & 2 & 7 & 27 & 0 & 2 & 14 & 0 & 5 & 95 \\
        LSTM-AE   & 30 & 10 & 2 & \underline{9} & \underline{28} & 0 & 2 & 10 & 0 & 5 & 96 \\
        LSTM-VAE  & 29 & 8 & 1 & 1 & 27 & 0 & 2 & 12 & 0 & 4 & 84 \\
        LSTM-SAE  & 29 & 8 & 3 & 1 & 27 & 0 & 2 & 13 & 1 & 5 & 89 \\
        FAE       & \underline{43} & \underline{13} & \underline{6} & 8 & \underline{28} & \underline{5} & \textbf{11} & \underline{24} & \underline{7} & \underline{18} & \underline{163} \\
        FourStat  & \textbf{48} & \textbf{18} & \textbf{8} & \textbf{10} & \textbf{30} & \textbf{6} & \textbf{11} & \textbf{26} & \textbf{11} & \textbf{20} & \textbf{188} \\
        \hline
        Adaptive FourStat & \textbf{49} & \textbf{19} & \textbf{11} & \textbf{10} & \textbf{30} & \textbf{7} & \textbf{11} & \textbf{27} & \textbf{12} & \textbf{20} & \textbf{196} \\
        \hline
        \textbf{TS count} & 57 & 28 & 14 & 19 & 45 & 17 & 20 & 53 & 23 & 35 & 311  \\
        \hline
    \end{tabular}
    }
\end{table*}

Table~\ref{tab:results-by-type} highlights the strengths and limitations of different architectural choices in TSAD. \textbf{FourStat} achieves the highest total score, demonstrating that combining frequency-aware, locally stable, and point-sensitive modules is essential for effective detection.

FourStat excels at detecting subtle or uniform anomaly signatures (e.g., amplitude changes, flat segments) that elude conventional models. Its SAE, with a stationarity-enforcing loss and MaxDiff objective, enhances sensitivity to localized deviations and short-term dynamics.

The FAE is a critical component, enabling strong global modeling through large windows and frequency-domain analysis. Its ability to capture periodic structure and spectral misalignments is central to detecting anomalies like frequency shifts and time-warped patterns.

FourStat is particularly adept at detecting complex structural distortions like temporal shifts and reversals. These often confound models with limited context, but FAE's long-range frequency modeling allows FourStat to identify disruptions in temporal continuity.

Beyond traditional metrics, we employ the strict, threshold-free \textbf{Perfect Point} metric, which tests whether a model can achieve perfect precision and recall simultaneously. FourStat satisfies this criterion more often than any other model, confirming the superior separability of its anomaly scores.

Despite FourStat's strong performance, univariate anomaly detection remains challenging. Certain subtle anomaly types reveal the persistent limitations of generic losses and monolithic architectures.

    \section{Conclusion}\label{sec:conclusion}



FourStat introduces a dual-model framework for time series anomaly detection, combining a frequency-based AutoEncoder (FAE) for global patterns and a standardized AutoEncoder (SAE) for localized anomalies. The FAE handles nonstationarity and long-range dependencies, while the SAE uses a novel loss function to detect transient anomalies. We propose Perfect Point, a threshold-free evaluation metric, and demonstrate FourStat's superiority over state-of-the-art methods across benchmarks. Experiments confirm its robustness, especially in detecting diverse anomaly types where other models fail.



    
    

    \bibliography{main}
    \bibliographystyle{tmlr}
    

    \newpage 
    \appendix
    
    \tableofcontents
    
    \clearpage
    \section{Proofs of MSE and RSE Surfaces}\label{sec:more-proofs}

\paragraph{Motivation and Problem Statement.}
In regression tasks involving oscillatory or periodic signals, the choice of loss function significantly affects convergence and generalization. A common loss is the Mean Squared Error (MSE), which penalizes deviations in predicted values from the ground truth. However, it suffers from an undesirable bias toward predicting the mean of the target, particularly in noisy regimes or early training. This leads to convergence toward a trivial solution where the model fails to capture essential structure in the data.

To mitigate this, we propose a composite loss combining MSE with the Relative Squared Error (RSE), which penalizes errors relative to the magnitude of the target signal itself. The RSE encourages the model to explain variations in the target more faithfully, especially when the absolute error alone is insufficient to guide optimization.

We analyze the gradient behavior of MSE and RSE with respect to the amplitude and frequency parameters in a simple sine-based signal model. Our goal is to characterize their loss surfaces and understand their optimization dynamics.

\paragraph{Model Setup.}
Let the ground truth signal be given by:
\[
s(t) = a \sin(bt),
\]
and the predicted signal be parameterized as:
\[
\hat{s}(t) = x \sin(yt),
\]
where \( a, b \) are ground truth amplitude and frequency, and \( x, y \) are the model's learnable parameters. Let \( N \) denote the number of sampled time points.

\begin{theorem}[Behavior of MSE and RSE gradients at initialization]
\label{thm:mse-rse-init}
Let \( \mathcal{L}_{\text{MSE}} \) and \( \mathcal{L}_{\text{RSE}} \) denote the mean squared error and relative squared error between the signals \( s(t) \) and \( \hat{s}(t) \), respectively. Then, at initialization when \( x = 0 \), the gradient of MSE with respect to \( x \) is approximately zero for generic values of \( b \) and \( y \), resulting in a plateau that can slow down learning. On the other hand, the gradient of RSE remains informative and encourages movement toward \( x = a \); however, it may present near‑flat regions that require careful optimization, especially when \( y = k b \).
\end{theorem}

\begin{proof}
We know that the \textbf{MSE} loss is defined as follows:
\[
\mathcal{L}_{\text{MSE}} = \frac{1}{N} \sum_{t=1}^N \left[ a \sin(bt) - x \sin(yt) \right]^2
\]

Taking the derivative with respect to $x$ we obtain:

\[
\frac{d}{dx} \mathcal{L}_{\text{MSE}} = -\frac{2}{N} \sum_{t=1}^N \left[ a \sin(bt) \sin(yt) - x \sin^2(yt) \right] 
\]

And by taking the derivative with respect to $y$ we obtain:

\[
\frac{d}{dy} \mathcal{L}_{\text{MSE}} = -\frac{2x}{N} \sum_{t=1}^N \left[ a \sin(bt) - x \sin(yt) \right] \cdot  \cos(yt) t
\]

At \( x = 0 \), obviously $\frac{d}{dy} \mathcal{L}_{\text{MSE}} = 0$ and for $\frac{d}{dx}$ we have:

\[
\left. \frac{d}{dx} \mathcal{L}_{\text{MSE}} \right|_{x = 0} = -\frac{2a}{N} \sum_{t=1}^N \sin(bt) \sin(yt)
\]

Using the identity \( \sin (A) \sin (B) = \frac{1}{2}[\cos(A - B) - \cos(A + B)] \), we have:
\[
\left. \frac{d}{dx} \mathcal{L}_{\text{MSE}} \right|_{x = 0} = -\frac{a}{N} \sum_{t=1}^N \left[ \cos((b - y)t) - \cos((b + y)t) \right]
\]
Using the identity for the sum of a finite cosine sequence:
\[
S(\omega) = \sum_{t=1}^N \cos(\omega t) = \frac{\sin\left( \frac{N \omega}{2} \right)}{\sin\left( \frac{\omega}{2} \right)} \cdot \cos\left( \frac{(N+1)\omega}{2} \right)
\]

we write:
\[
\sum_{t=1}^N \left[ \cos((b - y)t) - \cos((b + y)t) \right] 
= S(b - y) - S(b + y)
\]

Both \( S(b - y) \) and \( S(b + y) \) are of the same structure:
\[
S(\omega) = \frac{\sin\left( \frac{N \omega}{2} \right)}{\sin\left( \frac{\omega}{2} \right)} \cdot \cos\left( \frac{(N+1)\omega}{2} \right)
\]

Assuming that \( {b \pm y}/{\pi} \notin \mathbb{Q} \), the sine terms in the denominator do not vanish and the cosine terms oscillate in a bounded way. As \( N \to \infty \), the oscillations average out and the two sums become asymptotically equal:
\[
S(b - y) \approx S(b + y)
\quad \Rightarrow \quad
\sum_{t=1}^N \left[ \cos((b - y)t) - \cos((b + y)t) \right] \approx 0
\]

For general \( b \neq y \), the sum of oscillatory cosine terms approximately cancels out over the interval, implying:
\[
\left. \frac{d}{dx} \mathcal{L}_{\text{MSE}} \right|_{x = 0} \approx 0
\]

We can now proceed to the \textbf{RSE} loss which is defined as:
\[
\mathcal{L}_{\text{RSE}} = \frac{1}{N} \sum_{t=1}^N \left( \frac{a \sin(bt) - x \sin(yt)}{a \sin(bt) + \epsilon} \right)^2
\]

Taking the derivative with respect to $x$ we obtain:
\[
\frac{d}{dx} \mathcal{L}_{\text{RSE}} = -\frac{2}{N} \sum_{t=1}^N 
\frac{ \left(a \sin(bt) - x \sin(yt)\right) \cdot \sin(yt) }{ \left(a \sin(bt) + \epsilon\right)^2 } 
\]

Similarly by taking the derivative with respect to $y$ we have:

\[
\frac{d}{dy} \mathcal{L}_{\text{RSE}} = -\frac{2x}{N} \sum_{t=1}^N 
\frac{ \left(a \sin(bt) - x \sin(yt)\right) \cdot \cos(yt) t }{ \left(a \sin(bt) + \epsilon\right)^2 }
\]

When \( y = k b \) with \( k \neq 1 \), the terms \( \sin(yt) \) and \( \sin(bt) \) oscillate at harmonically related frequencies. For large \( N \), the cross‑correlation between these harmonically locked components leads to near‑cancellation. Consequently, the gradient becomes very small (though not identically zero) when \( x \) is close to \( a \). A more accurate approximation is obtained by noting the near‑orthogonality of the harmonic components:

\[
\frac{d}{dx} \mathcal{L}_{\text{RSE}} \bigg|_{y=kb} \approx -\frac{2}{N} \sum_{t=1}^N \frac{(a - x) \sin(bt) \sin(kbt)}{(a \sin(bt) + \epsilon)^2},
\]
\[
\frac{d}{dy} \mathcal{L}_{\text{RSE}} \bigg|_{y=kb} \approx -\frac{2x}{N} \sum_{t=1}^N \frac{(a - x) \sin(bt) \cos(kbt) \, t}{(a \sin(bt) + \epsilon)^2}.
\]

Both expressions are proportional to \( (a - x) \). Therefore, when \( x \) is near \( a \), the gradients become extremely small, creating a near‑flat region (a shallow local minimum) in the loss landscape. These regions are not true plateaus, because the gradient does not vanish exactly, but they can still slow down optimization. The flatness is most pronounced when \( y \) is an exact harmonic of \( b \); for non‑harmonic frequencies the gradient magnitude is typically larger, helping to escape the shallow basin.
\end{proof}

As a result, we know \textbf{MSE} tends to exhibit \textit{vanishing gradients} at initialization when predictions are small (\( x \approx 0 \)), providing little signal to move toward the true amplitude \( a \) or frequency \( b \).
In contrast, \textbf{RSE} retains sensitivity even when the prediction is zero, and promotes moving in the direction that captures the target signal's shape and scale.

\begin{figure*}[ht]
    \centering
    \subfigure[MSE Loss Contour]{\label{fig:mse2d}\includegraphics[width=0.32\textwidth]{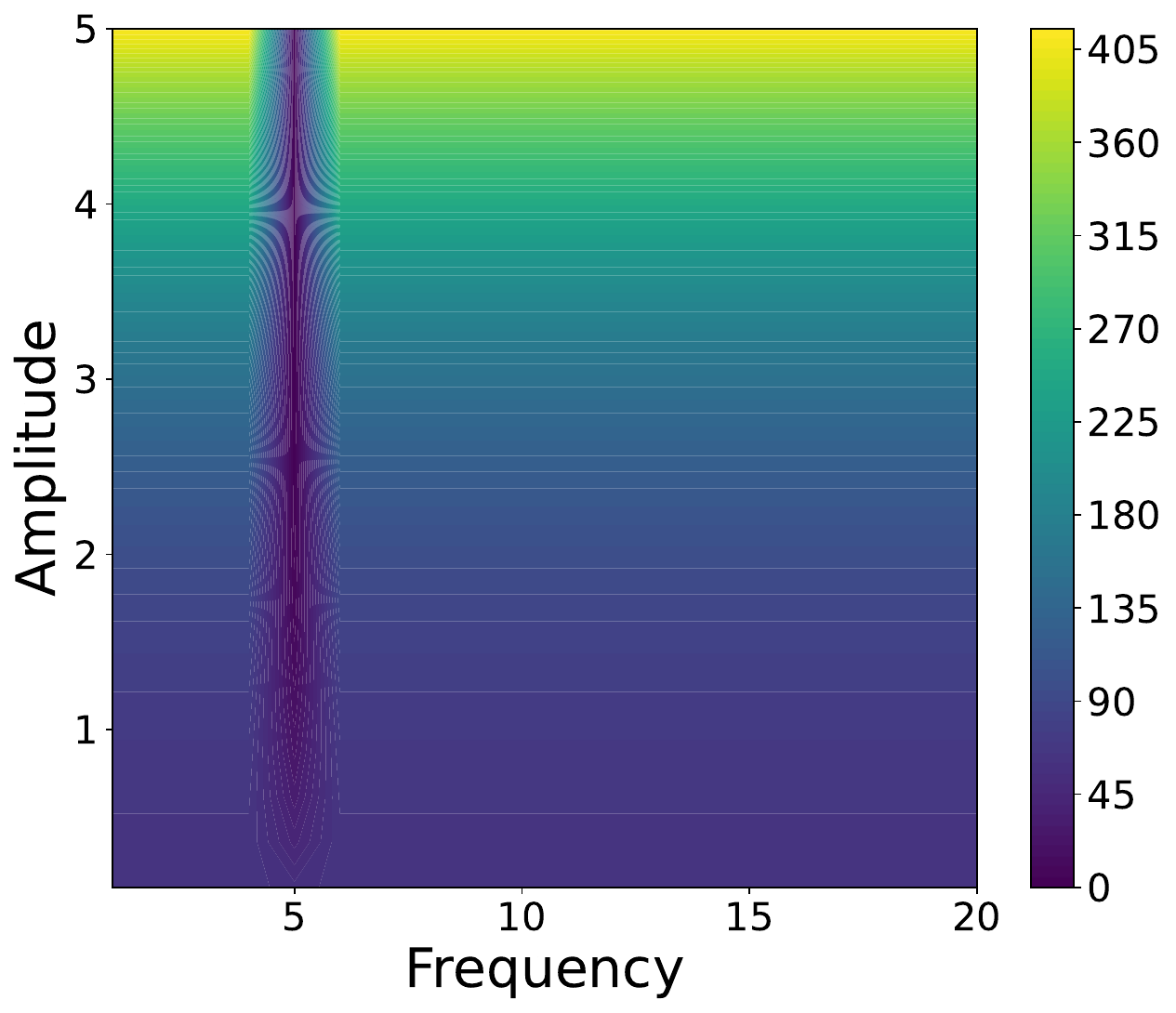}}
    \subfigure[RSE Loss Contour]{\label{fig:rse2d}\includegraphics[width=0.32\textwidth]{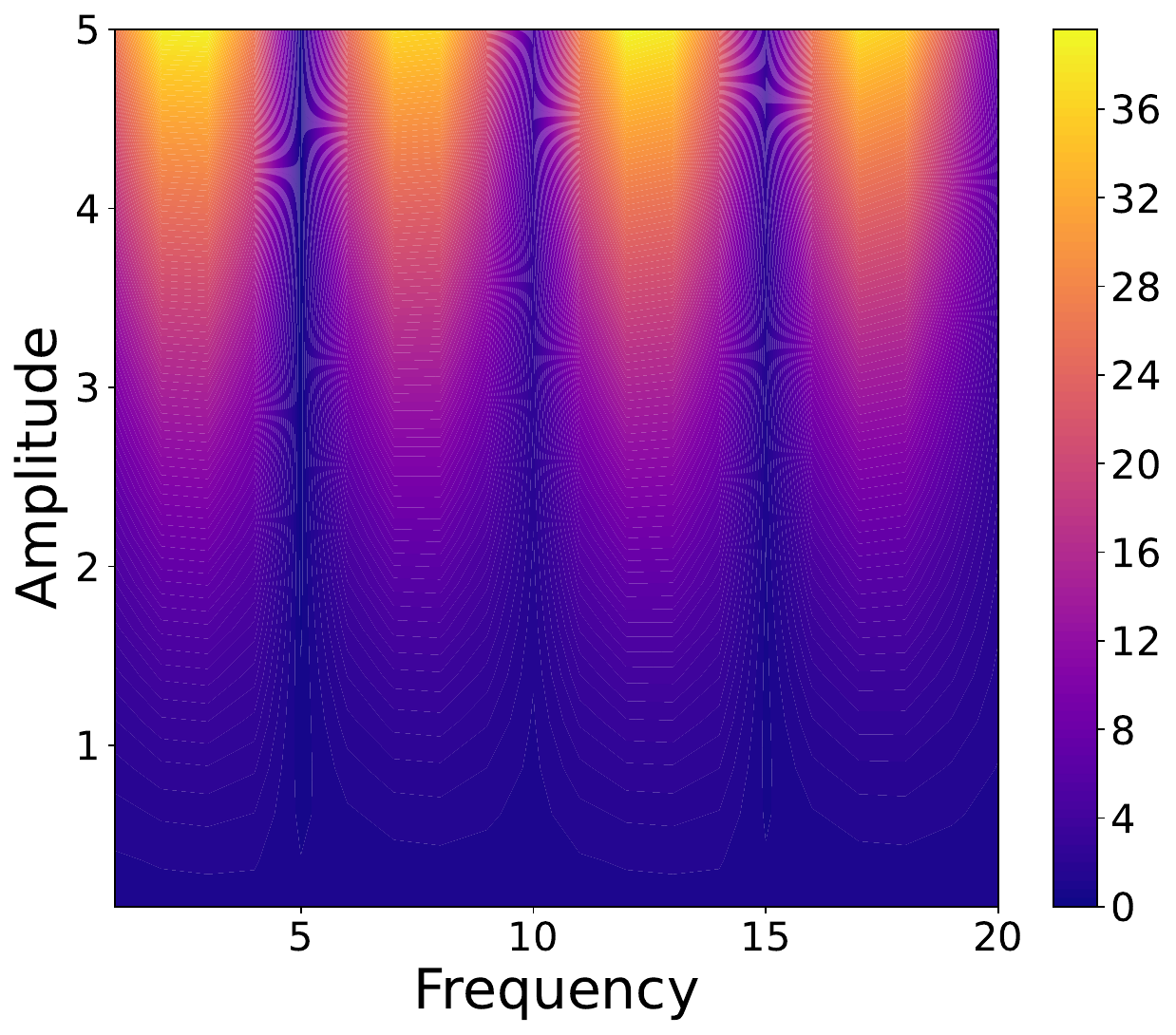}}
    \subfigure[$(\text{MSE} + \text{RSE})/2$ Loss Contour]{\label{fig:mse_rse2d}\includegraphics[width=0.32\textwidth]{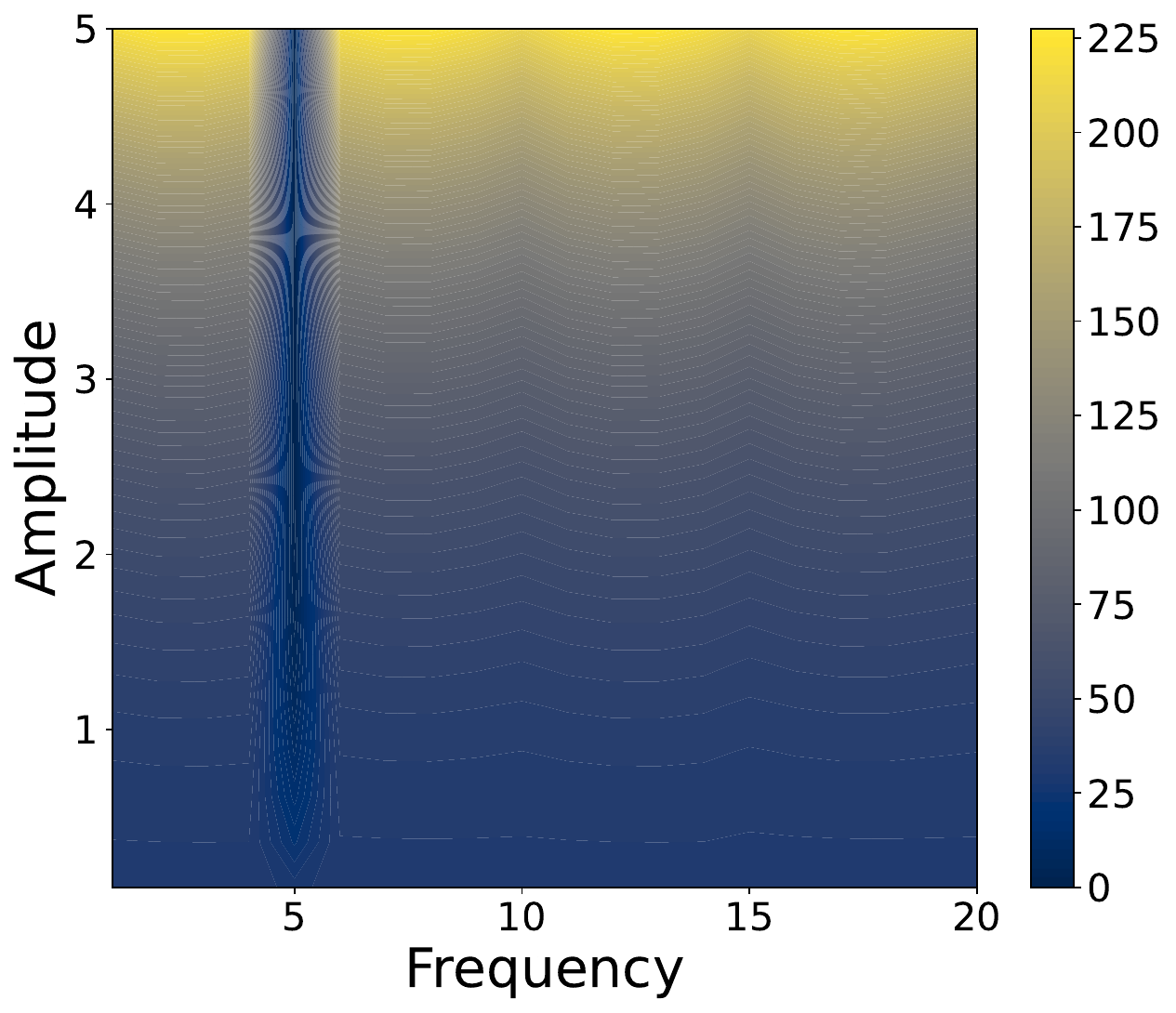}}
    \caption{
    2D contours of loss metrics comparing a reference signal \( s(t) = 4\sin(10\pi t) \) with test signals \( \hat{s}(t) = A\sin(2\pi F t) \), where amplitude \( A \) and frequency \( F \) vary over a grid.}
    \label{fig:loss_contour}
\end{figure*}

\begin{figure*}[ht]
    \centering
    \subfigure[MSE Loss Surface]{\label{fig:mse3d}\includegraphics[width=0.32\textwidth]{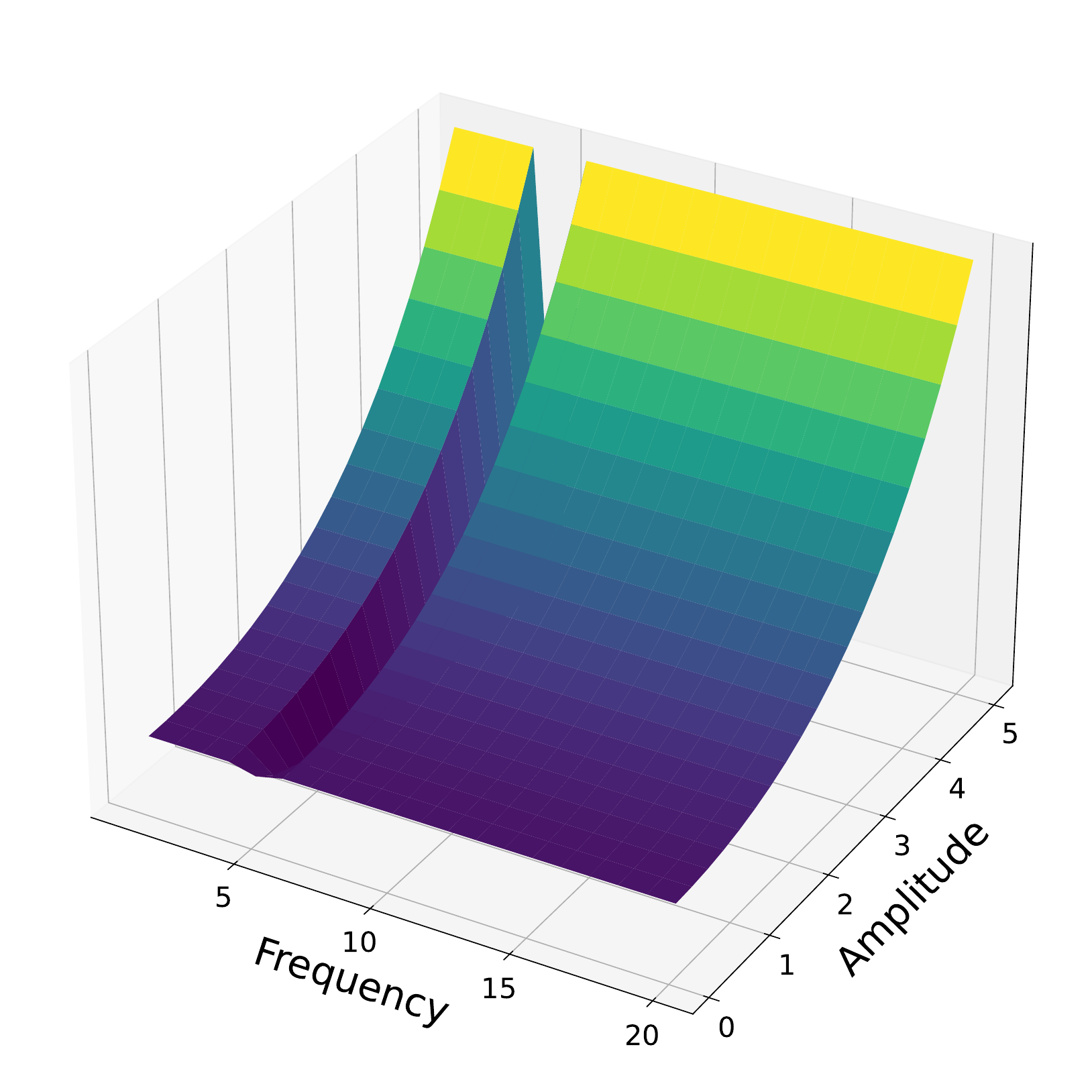}}
    \subfigure[RSE Loss Surface]{\label{fig:rse3d}\includegraphics[width=0.32\textwidth]{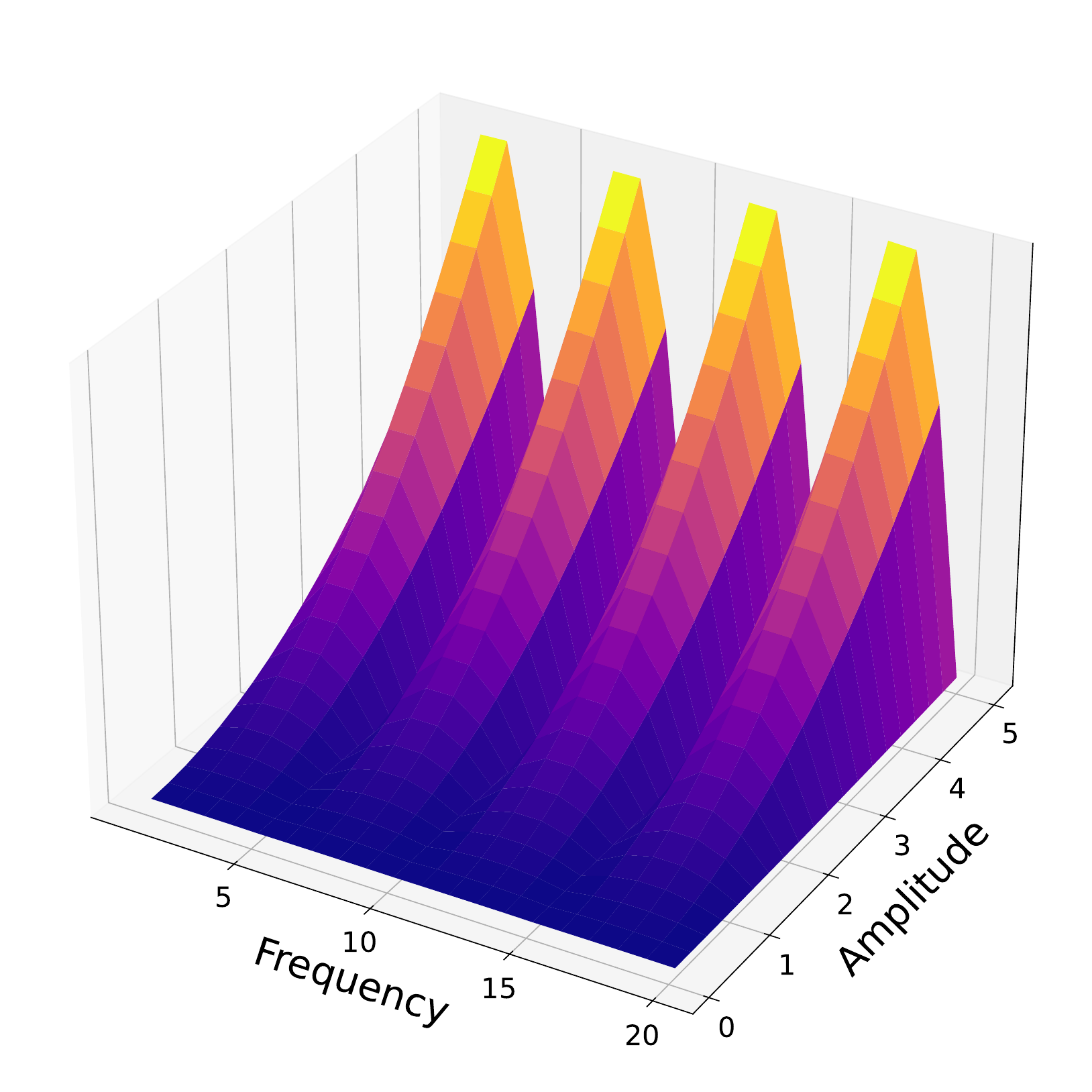}}
    \subfigure[$(\text{MSE} + \text{RSE})/2$ Loss Surface]{\label{fig:mse_rse3d}\includegraphics[width=0.32\textwidth]{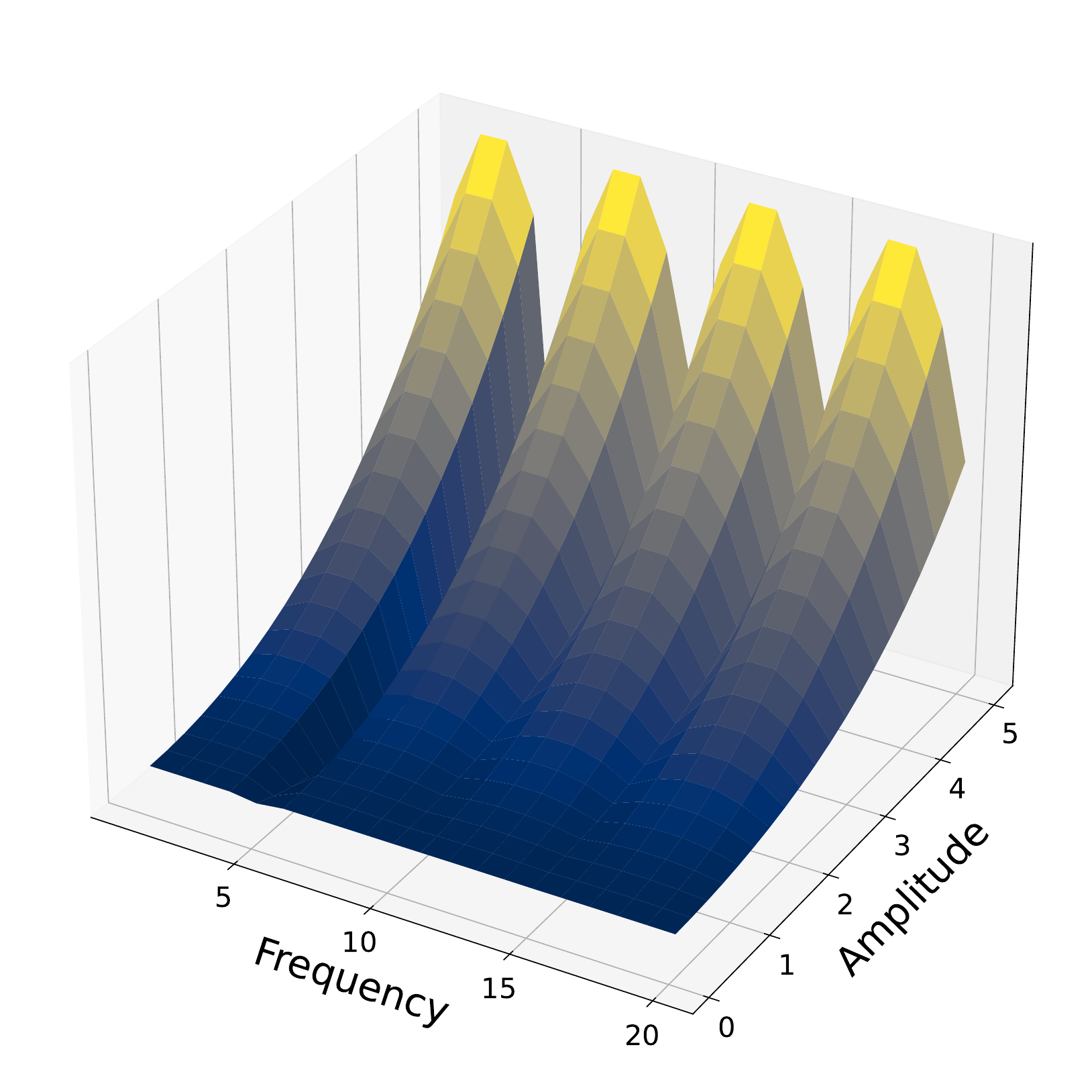}}
    \caption{
    3D surface plots of loss metrics comparing a reference signal \( s(t) = 4\sin(10\pi t) \) with test signals \( \hat{s}(t) = A\sin(2\pi F t) \), where amplitude \( A \) and frequency \( F \) vary over a grid.}
    \label{fig:loss_surface}
\end{figure*}

This theoretical analysis aligns with the empirical observations that MSE has a sharp, narrow minimum around the correct solution, whereas RSE exhibits multiple flat regions or local minima even when predictions match the shape, but not necessarily the scale of the target signal (e.g., \( y = kb \), \( x \approx a \) for integer \( k \)).
You can see the results of this combination for an example in Figure~\ref{fig:loss_contour} and Figure~\ref{fig:loss_surface}.

    \clearpage
    \section{Ablations}\label{sec:ablation}
In this section, we present a series of ablation studies to analyze the impact of different components of our model. First, we demonstrate how using a larger window size in conjunction with  FAE leads to significant improvements, particularly in groups associated with strong seasonality patterns. Next, we examine the performance benefits introduced by combining MSE and RSE as the loss function. We then compare the use of a SAE versus AE, and discuss the motivations for preferring SAE in our architecture. Finally, we discuss the motivation for introducing the \texttt{MaxDiff} loss and analyze its impact on model performance.

\subsection{FAE with Large Window Sizes}
As shown in Table~\ref{tab:detailed}, and as expected, the impact of this component is most evident in anomaly types related to \textit{frequency change} and \textit{time shift}. Additionally, another anomaly type, labeled \textit{reversed cycle}, can be interpreted as a form of frequency change, as it represents an inverted frequency pattern. Notably, our method achieved significant improvements across all three of these categories.

\begin{figure*}[ht]
    \centering
    \subfigure[Original test data for TSB-AD time series 209]{\label{fig:ts209}%
        \includegraphics[width=0.49\textwidth]{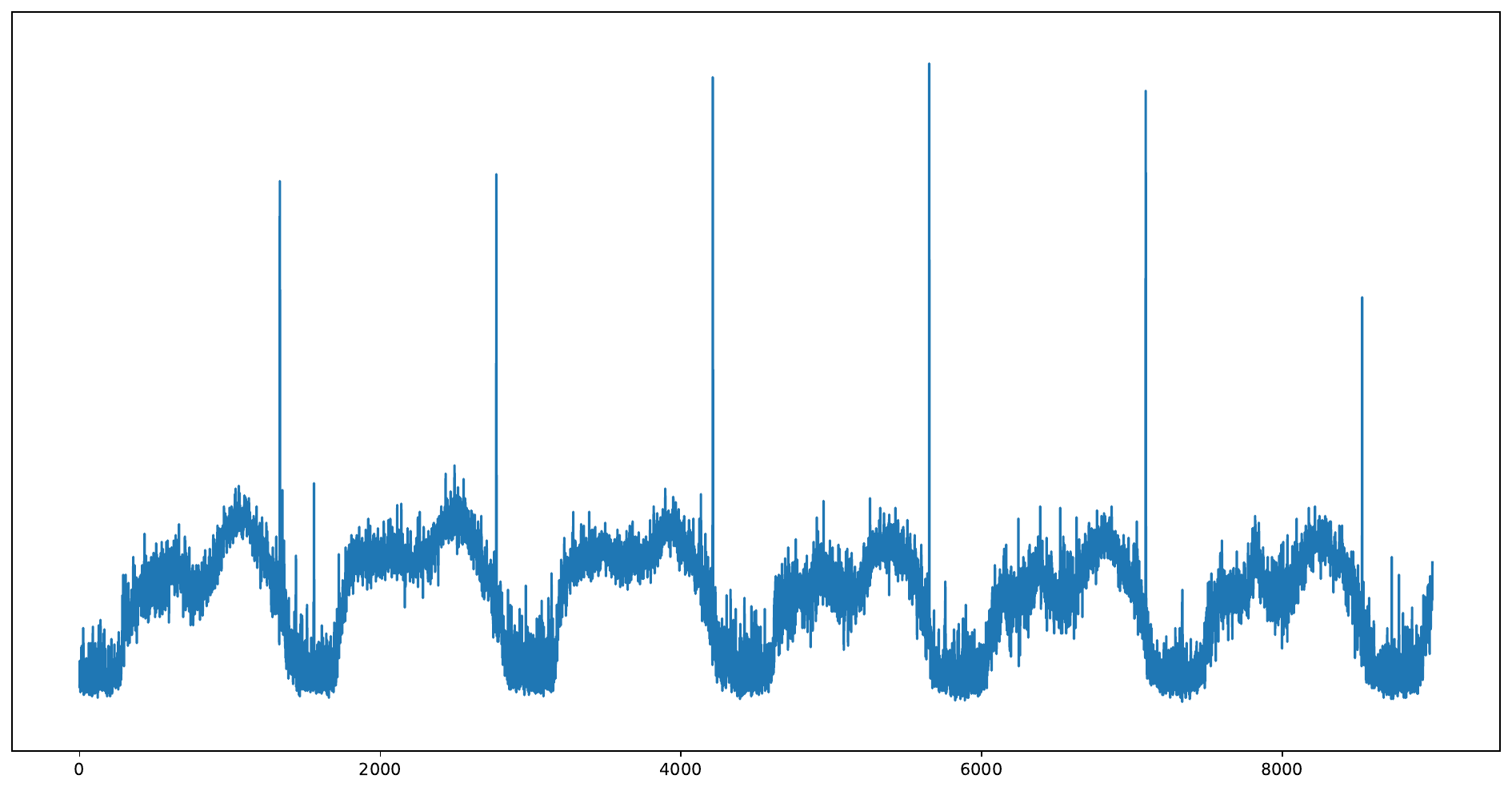}}
    \subfigure[FourStat reconstruction with window size 32]{\label{fig:rec32}%
        \includegraphics[width=0.49\textwidth]{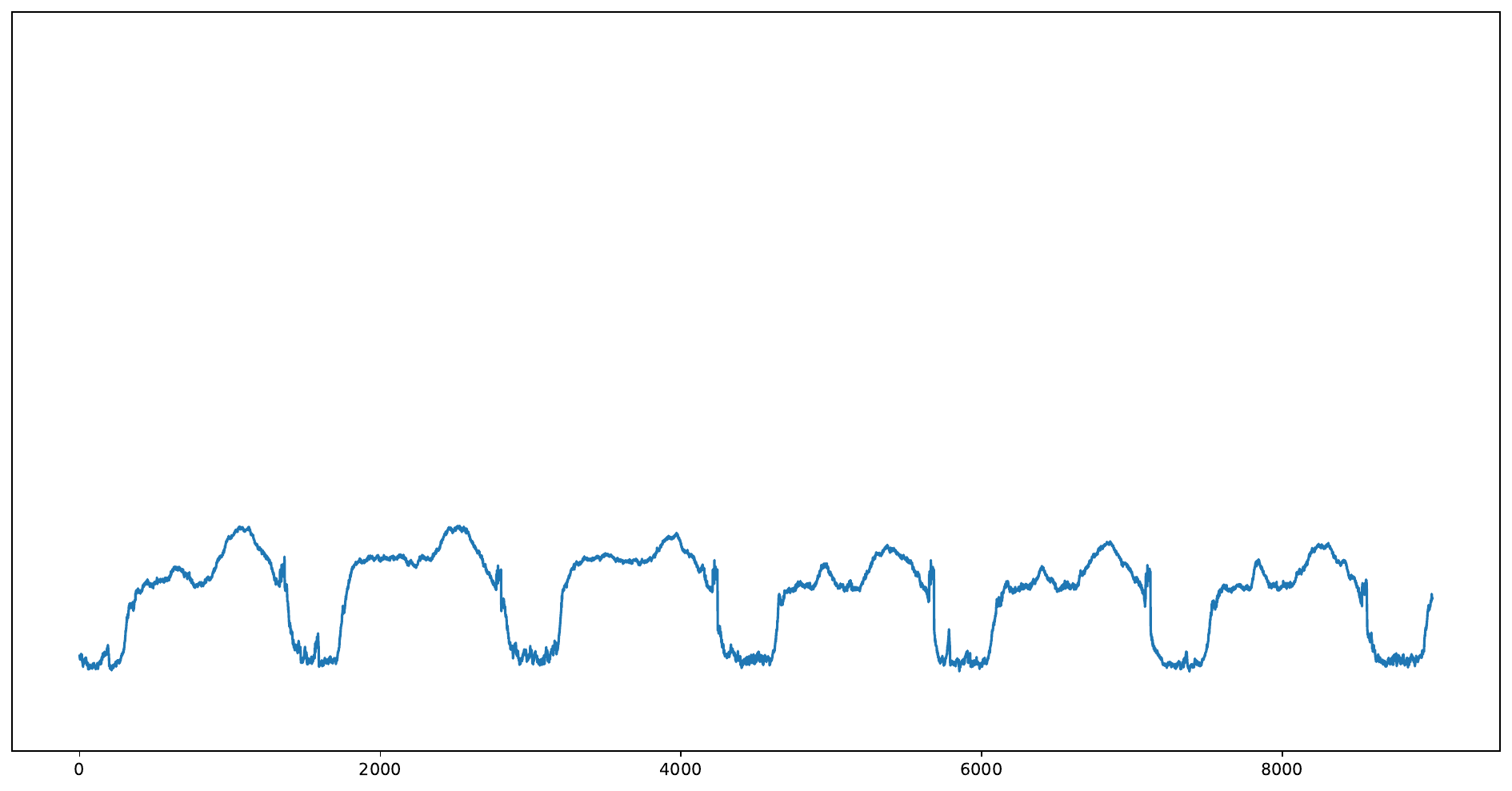}}
    \subfigure[FourStat reconstruction with window size 256]{\label{fig:rec256}%
        \includegraphics[width=0.49\textwidth]{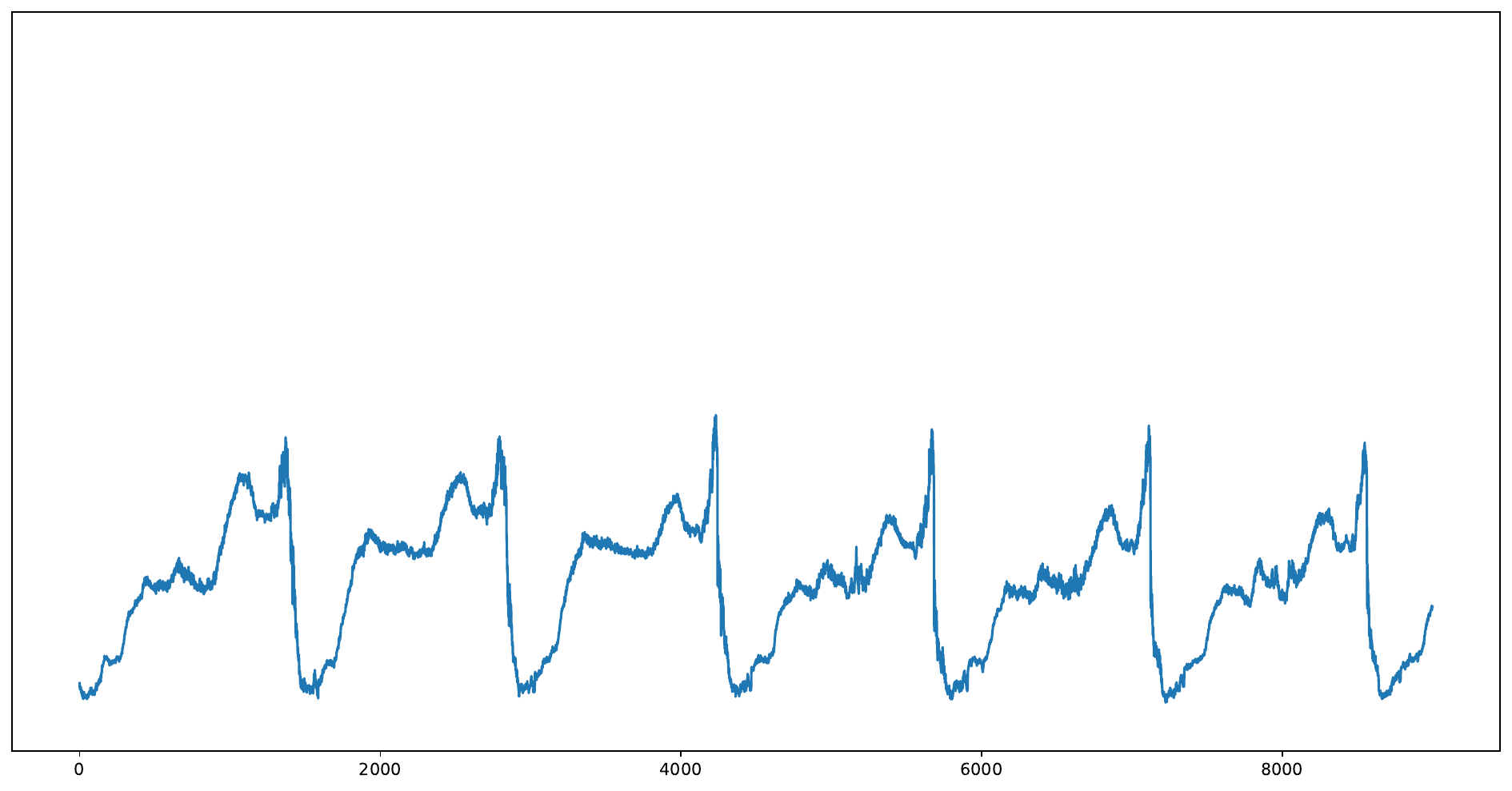}}
    \subfigure[FourStat reconstruction with window size 2048]{\label{fig:rec2048}%
        \includegraphics[width=0.49\textwidth]{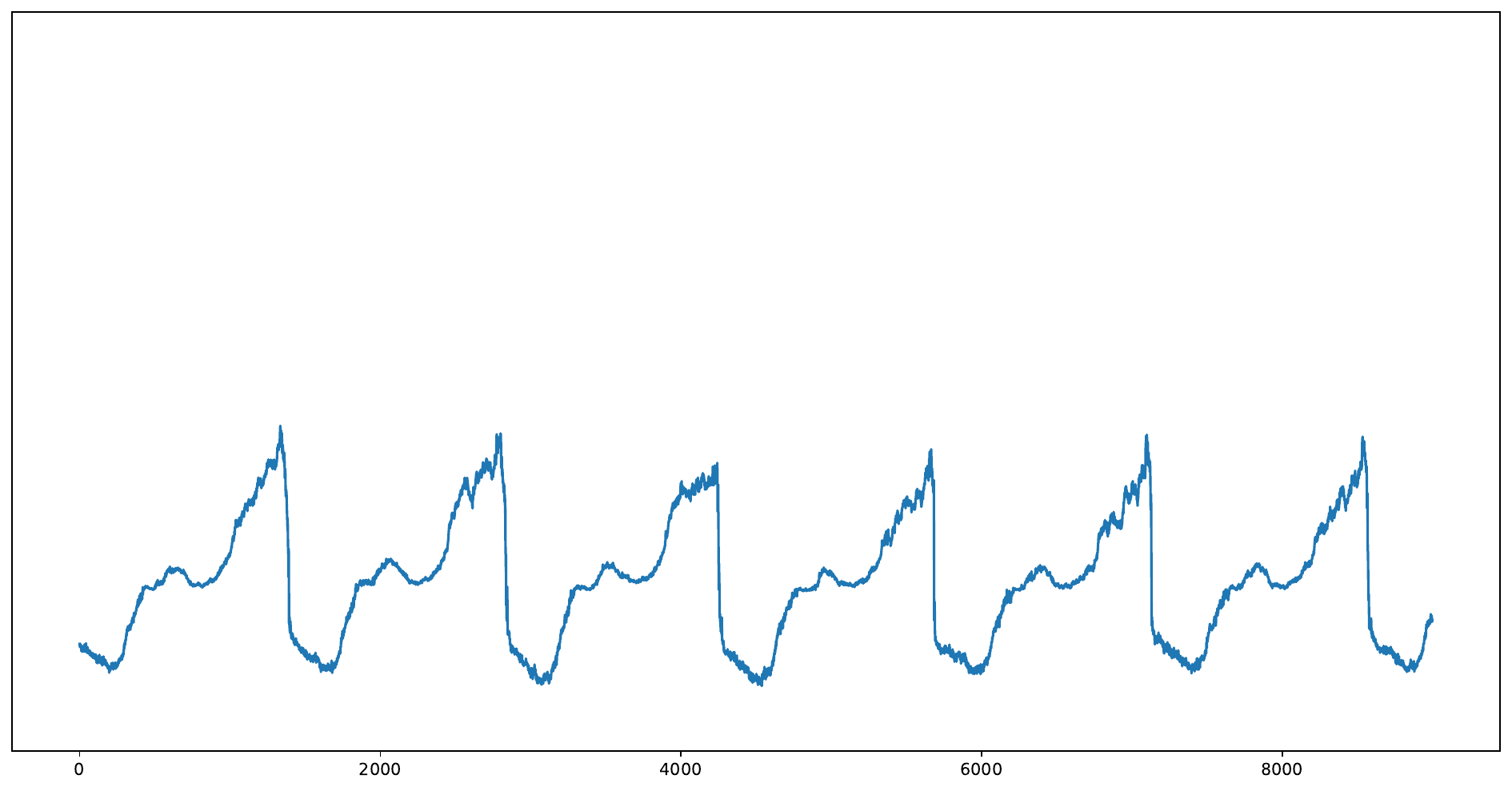}}
    \caption{Comparison of original and reconstructed time series with different window sizes.
    The original time series can be found in \subref{fig:ts209} and the reconstructions for FourStat models with different window sizes can be found in \subref{fig:rec32}, \subref{fig:rec256}, and \subref{fig:rec2048}.}
    \label{fig:window_size_plots}
\end{figure*}

It is important to mention that all experiments were conducted using a fixed window size of 128. 
However, we observed that certain anomalies, an example of which is illustrated in Figure~\ref{fig:window_size_plots}, could be more easily detected with larger window sizes.
This motivated our discussion on using adaptive window sizes as mentioned in Section~\ref{sec:method} and Appendix~\ref{sec:adaptive-window}.

\subsection{Fourstat Results: Fixed vs. Adaptive Window Comparison}
In Table~\ref{tab:results-by-type} we provide the results of our FourStat model with a fixed window size of 128, and an adaptive window size variant.
The adaptive window size variant uses a time series dependent window size whereas the fixed version uses a predefined window size.
The adaptive window size process is described in Section~\ref{sec:method} and Appendix~\ref{sec:adaptive-window}.

To provide more information about the overall consequences of choosing different window sizes, and the usefulness of the adaptive context selection, in Table~\ref{tab:fourstat-comparison} we provide the Perfect Point scores for the original FourStat model across window sizes ranging from 32 to 1024 against the adaptive variant.
The results are reported across 3 different seeds.
\begin{table}[ht]
    \small
    \caption{Comparison of Perfect Points (PP) for FourStat: Fixed vs. Adaptive Window Selection}
    \label{tab:fourstat-comparison}
    \centering
    \begin{tabular}{c c c c c}
        \hline
        \textbf{Method} 
        & \textbf{Window} 
        & \textbf{Seed 0} 
        & \textbf{Seed 123} 
        & \textbf{Seed 42} \\
        \hline
        \multirow{6}{*}{Fixed Window}
        & 32   & 168 & 167 & 170 \\
        & 64   & 183 & 184 & 185 \\
        & 128  & 185 & 188 & 188 \\
        & 256  & 174 & 177 & 185 \\
        & 512  & 175 & 179 & 171 \\
        & 1024 & 145 & 152 & 152 \\
        \hline
        \multirow{1}{*}{Adaptive Window}
        & --    & \textbf{196} & \textbf{197} & \textbf{195} \\
        \hline
    \end{tabular}
\end{table}

\subsection{Comparison of SAE and AE}

FourStat-AE and FourStat-SAE did not show a significant difference for the Perfect Point metric.
To make our motivation for choosing SAE more explicit, we designed an additional experiment. 
To further investigate our hypothesis regarding SAE’s ability to better capture the underlying data distribution, we conducted an additional experiment focusing on the statistical similarity between the reconstructed outputs and the original time series data.
While both models were trained using the same MSE loss, SAE introduces additional complexity through its multi-component loss function, making its training process more challenging and typically slower to converge.

For a fair comparison, we evaluated FourStat-AE and FourStat-SAE under identical settings: the same architecture, bottleneck size, optimizer, and number of training epochs (200). The only difference was the inclusion of an additional stationarity-enforcing loss term in SAE’s bottleneck. This test was conducted on 146 time series instances where the FourStat model had successfully detected the target pattern in previous evaluations.

To assess how well each model preserves the statistical characteristics of the data, we computed both the Jensen-Shannon Divergence (JSD) and the correlation coefficient between the reconstructed and original series. This setup allowed us to isolate the effect of the stationarity constraint in SAE and determine whether this added complexity translates to improved modeling of the true data structure.

\begin{table}[ht!]
    \caption{Number of test cases (out of 146) where each model outperformed the other based on Jensen-Shannon Divergence (lower is better) and Correlation (higher is better). LSTM-SAE consistently outperforms LSTM-AE across both metrics.}
    \label{tab:sae-results}
    \centering
    \begin{tabular}{c cc}
        \hline
        \textbf{Model} & \textbf{Jensen-Shannon Divergence (JSD)} & \textbf{Correlation}  \\
        \hline
        LSTM-AE     & 22 & 30  \\
        LSTM-SAE    & 124 & 116   \\
        \hline
    \end{tabular}
\end{table}

\subsection{Impact of MSE+RSE as the loss function}

We observe that many time series in the dataset exhibit seasonal patterns, with dominant periods typically between 32 and 128. This justifies the use of frequency-aware models such as FAE, which are capable of capturing such periodicity within the fixed window.
It should be noted that this benefit is limited to time series exhibiting seasonal behavior with period lengths shorter than the window size, and may not be applicable to all datasets.

\begin{figure}[!h]
    \centering
    \subfigure{\label{fig:tsb509}\includegraphics[width=0.49\textwidth]{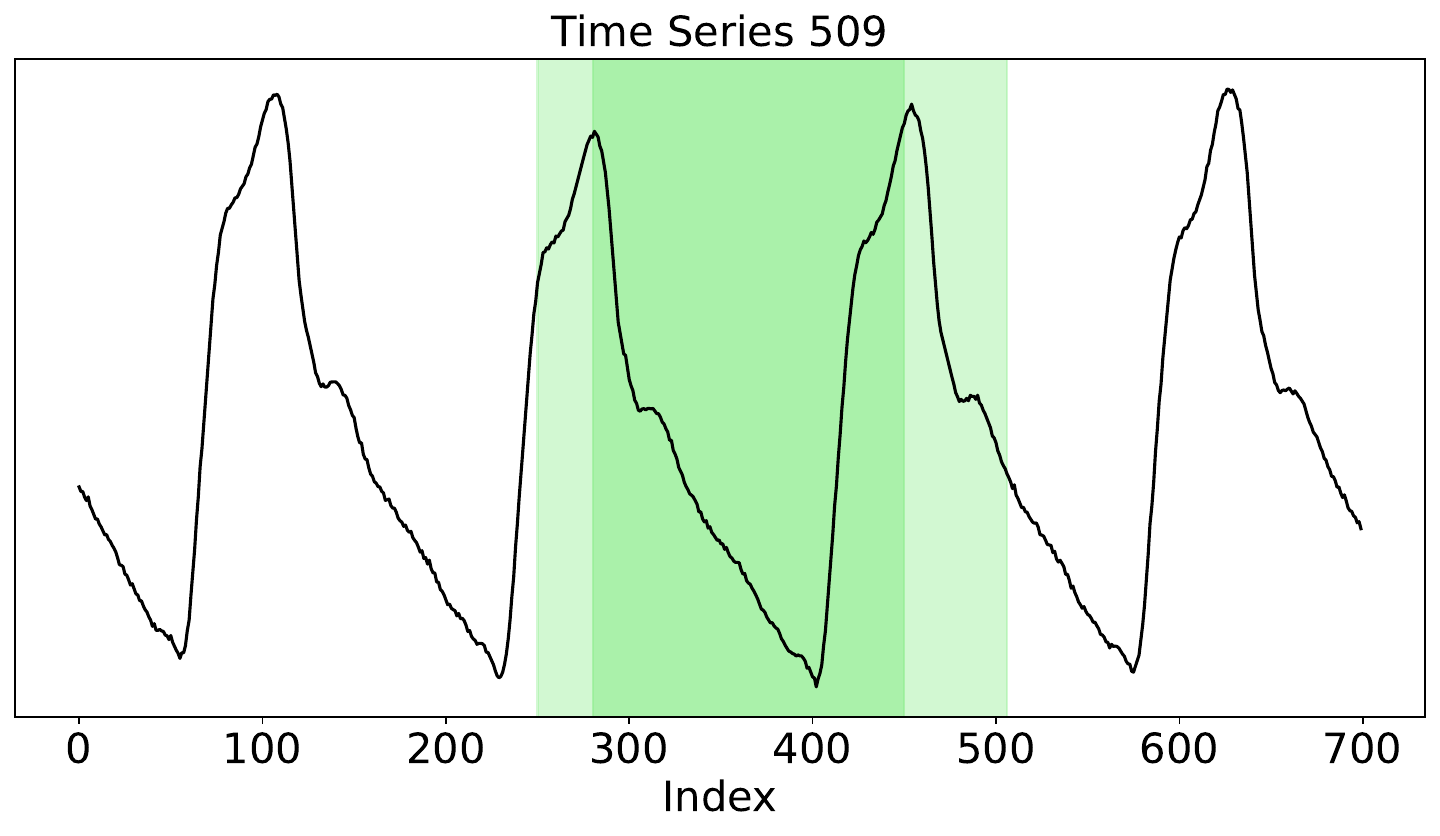}}
    \subfigure{\label{fig:tsb509conv}\includegraphics[width=0.49\textwidth]{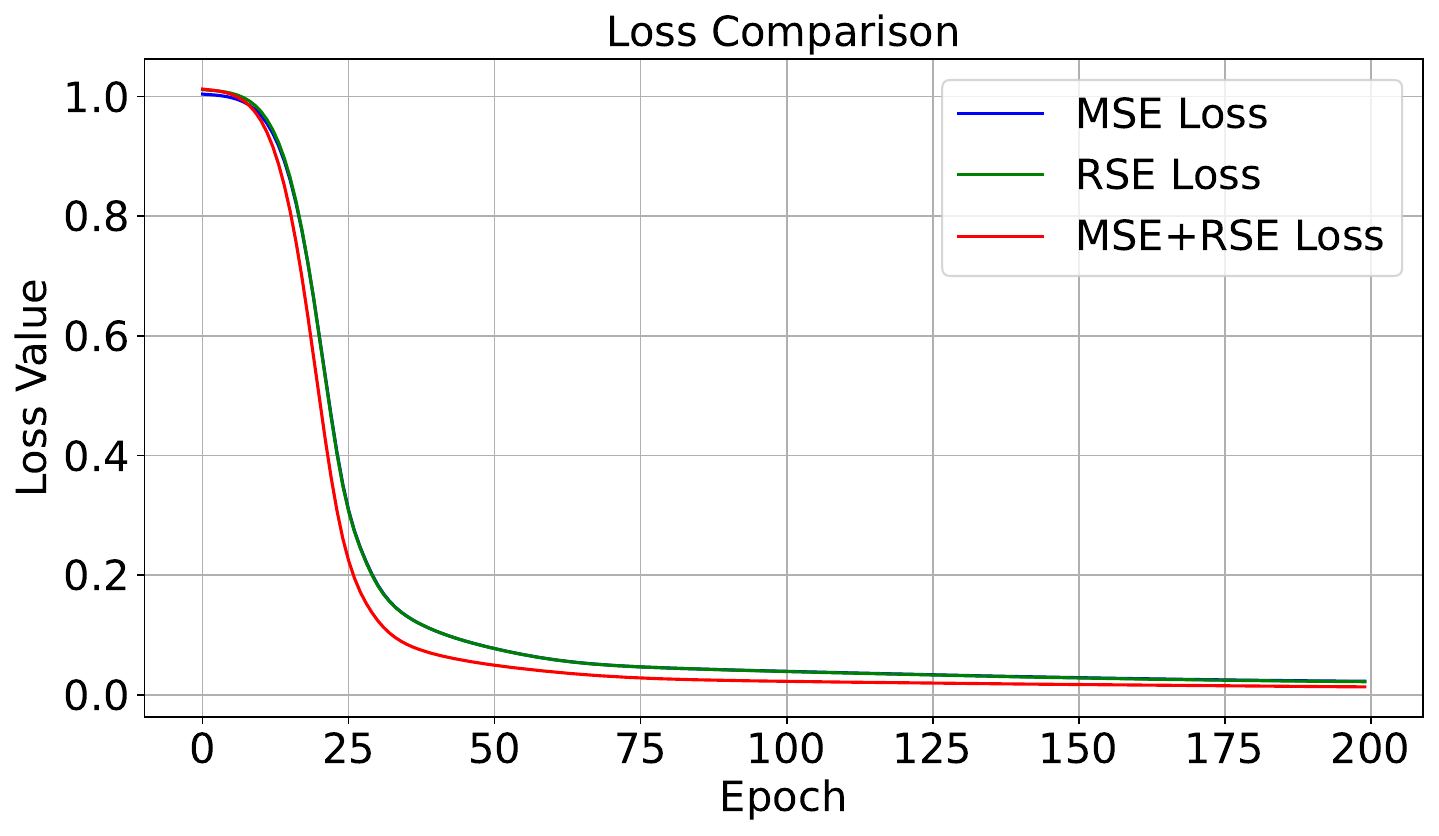}}
    \subfigure{\label{fig:tsb524}\includegraphics[width=0.49\textwidth]{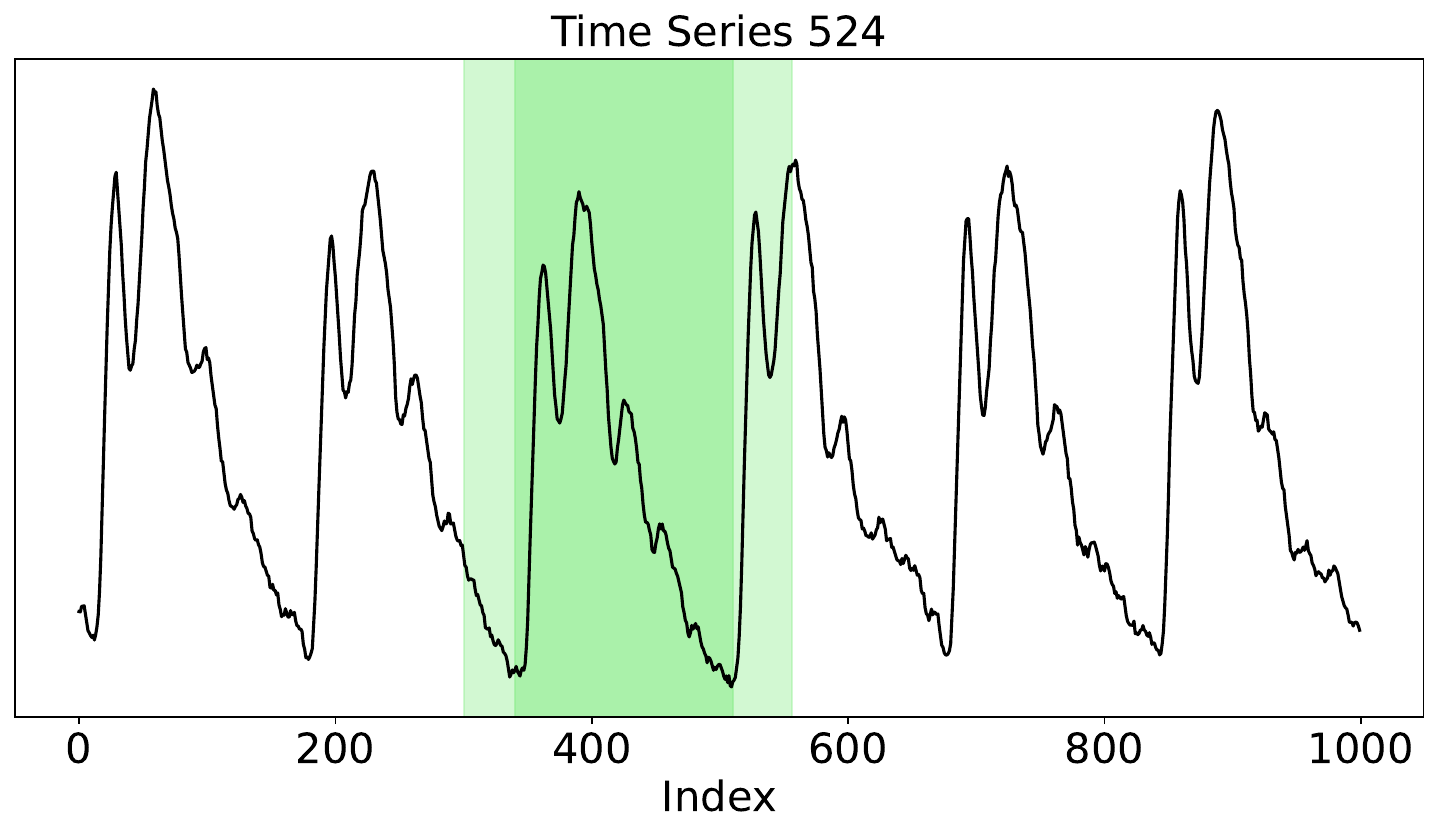}}
    \subfigure{\label{fig:tsb524conv}\includegraphics[width=0.49\textwidth]{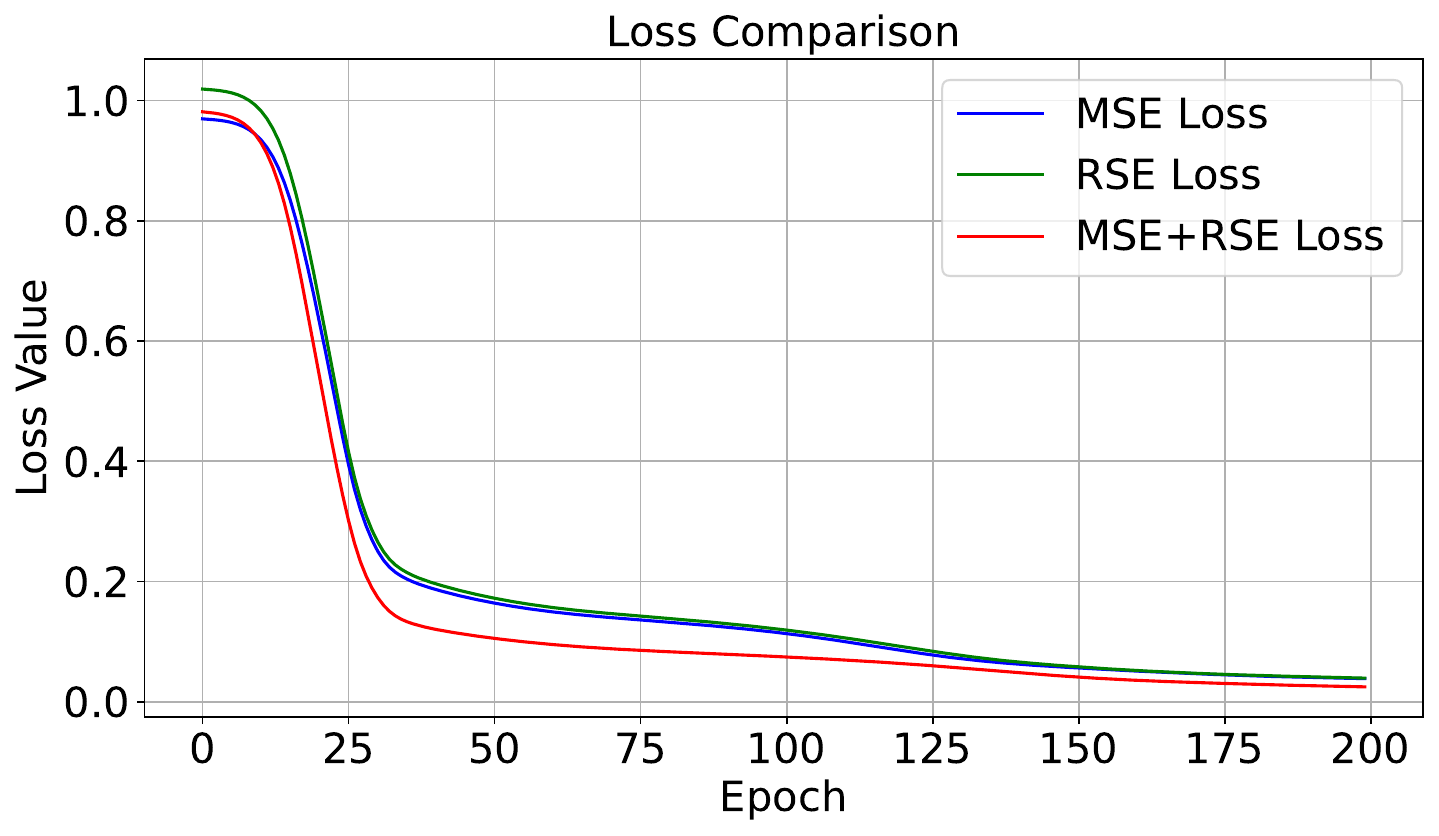}}
    \subfigure{\label{fig:tsb543}\includegraphics[width=0.49\textwidth]{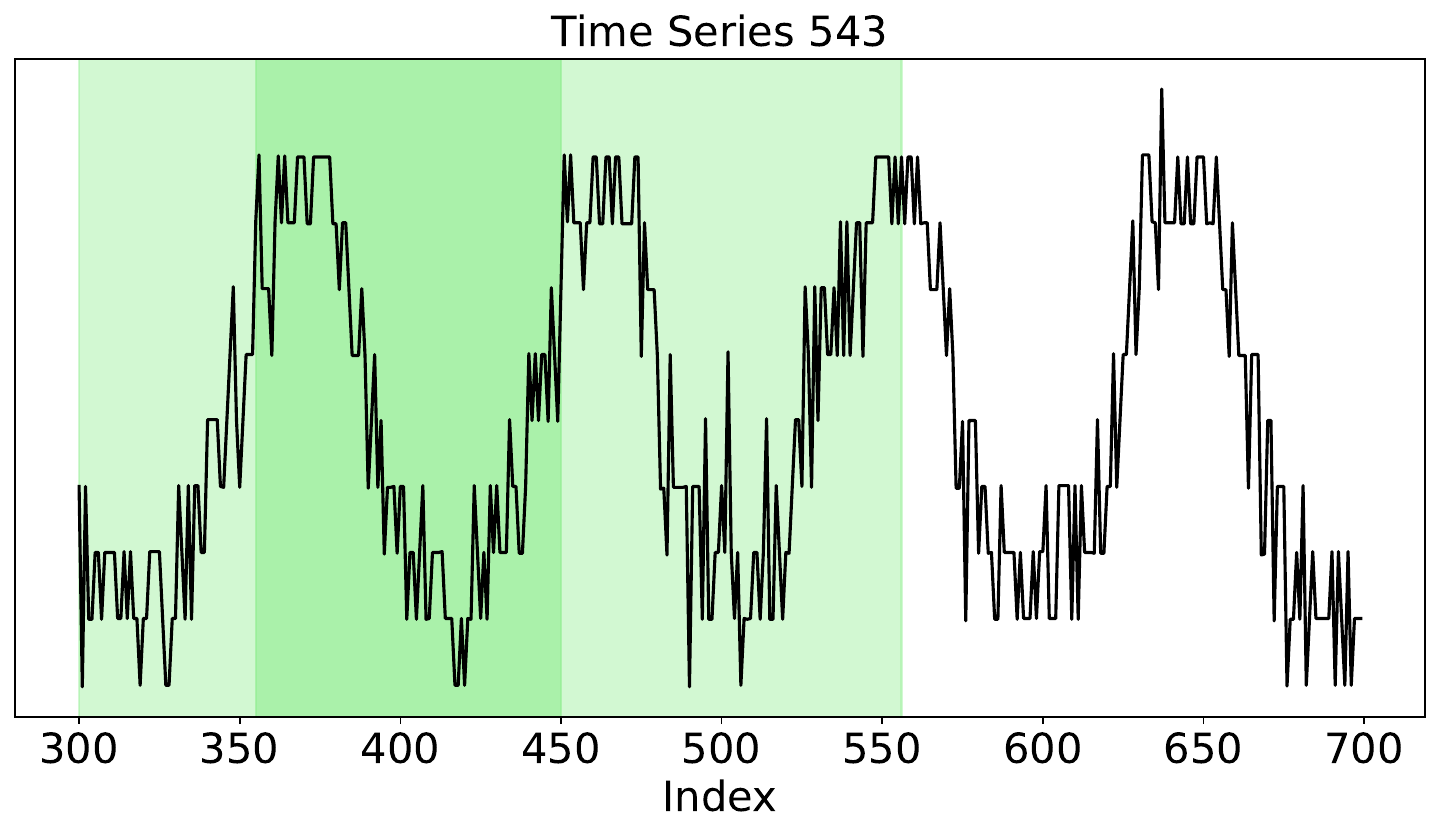}}
    \subfigure{\label{fig:tsb543conv}\includegraphics[width=0.49\textwidth]{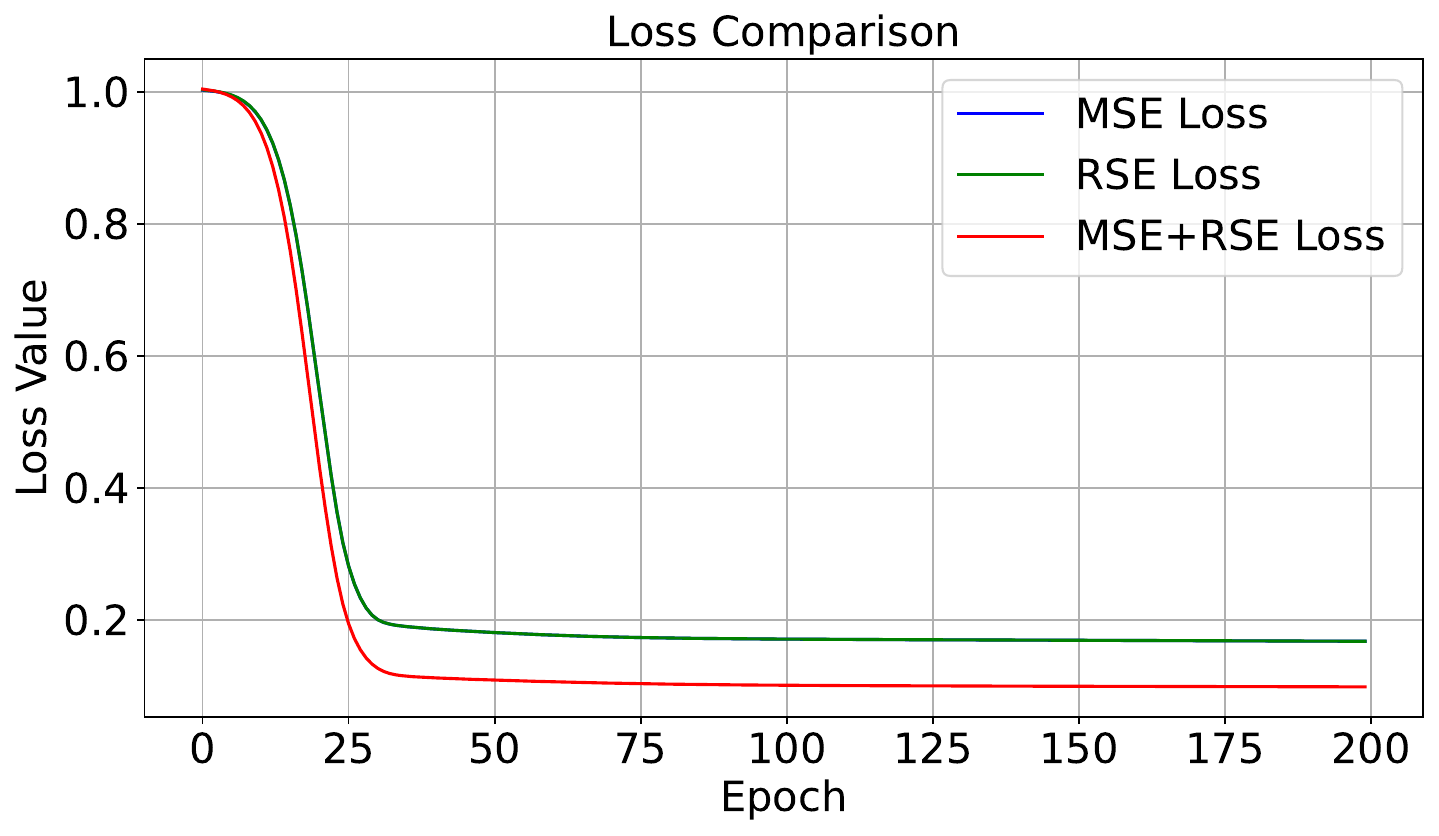}}
    \subfigure{\label{fig:tsb547}\includegraphics[width=0.49\textwidth]{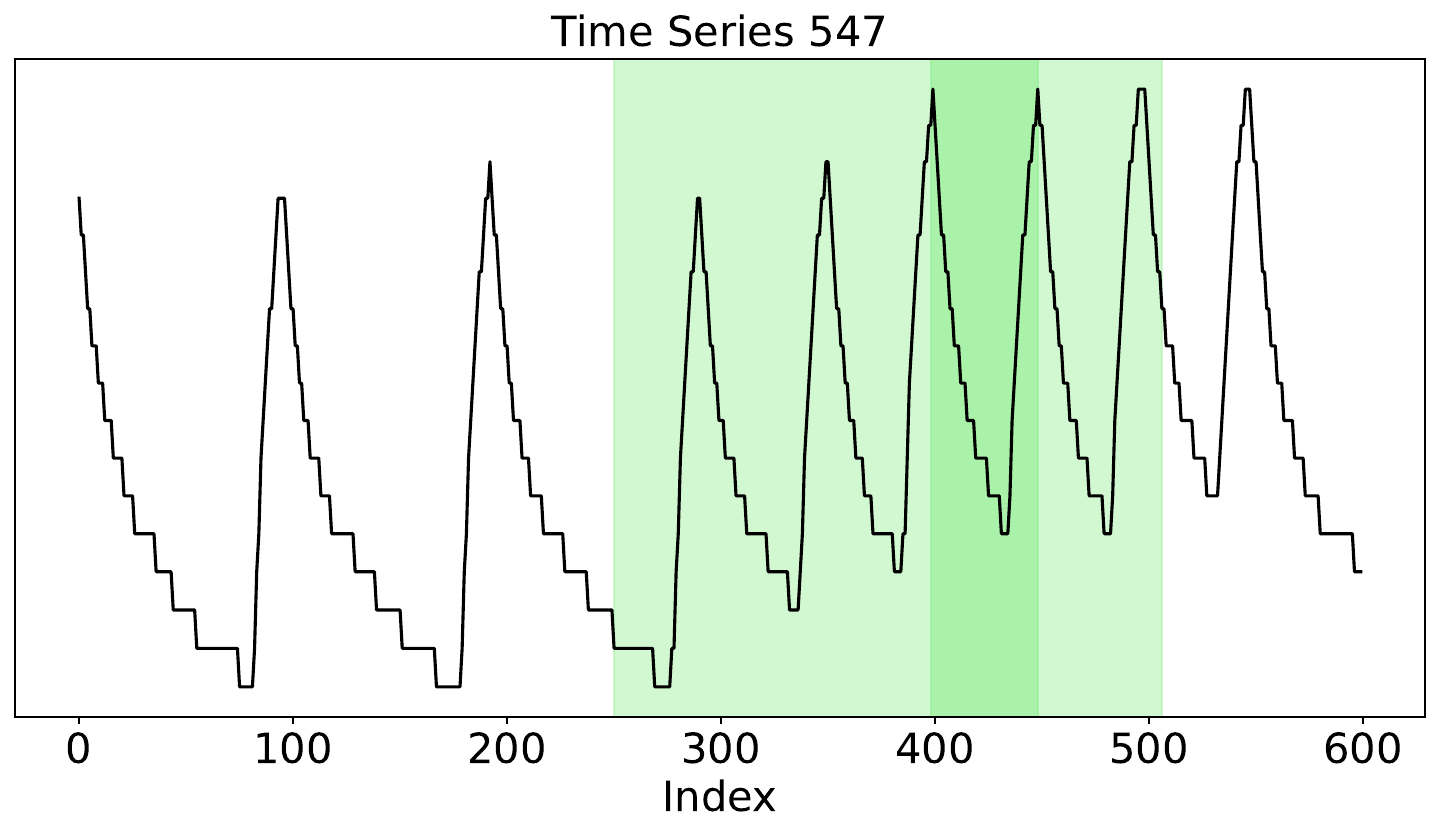}}
    \subfigure{\label{fig:tsb547conv}\includegraphics[width=0.49\textwidth]{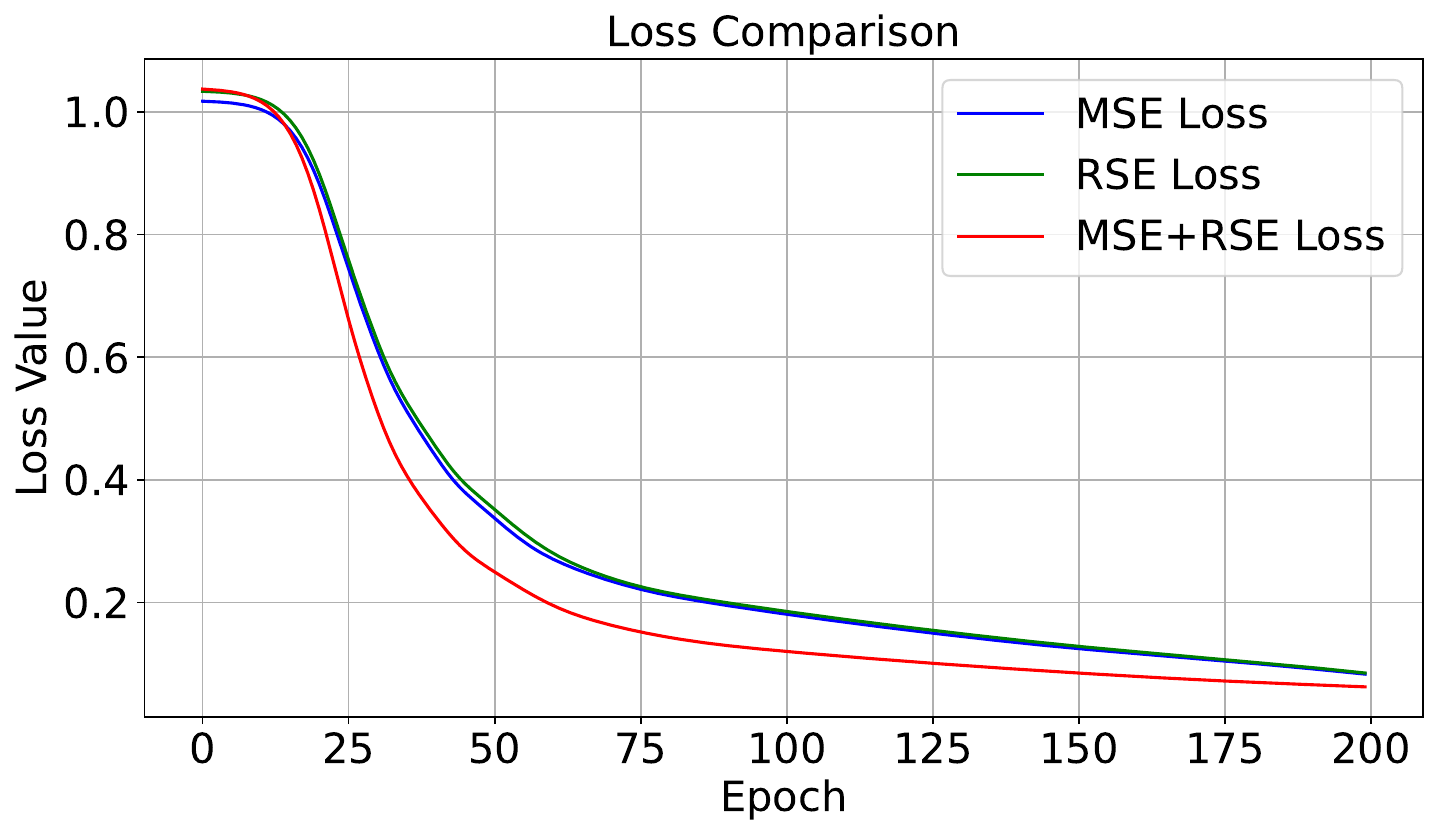}}
    \caption{
    When the window sizes are sufficiently large, we see that the composite loss function allows faster convergence for FAE.
    In each example one cycle has been highlited in green and we have also shown what a window for the model containing that cycle could look like.}
    \label{fig:fae_convergence}
\end{figure}

As previously mentioned, we use a fixed window size of 128 across all experiments. 
Based on our observations, time series that exhibit clear seasonality with periods shorter than 128 tend to benefit more from this setup. 
For these cases, the composite loss function is expected to converge better, as the model is more capable of capturing their periodic structure within the given window.
For other types of time series that do not show such seasonal behavior, we make no strong assumptions about convergence, as the theoretical justification mainly applies to periodic patterns.
As shown in Figure~\ref{fig:fae_convergence}, this combined loss converges faster and achieves lower final error, benefiting from both the absolute and relative error characteristics.

\subsection{MaxDiff Loss: Towards precise anomaly localization}

In our results, we observe no statistically significant difference between MSE and MaxDiff. The primary motivation behind introducing MaxDiff is not to outperform existing baselines in terms of aggregate scores, but to provide a foundation for more precise anomaly localization in future work.

We argue that using a large window size is beneficial: it captures broader temporal context and improves overall detection quality. Once an anomaly is detected within a large window, the next logical step is to localize it, that is, to determine where within the window the anomaly has occurred.

MaxDiff serves this purpose effectively: by substantially reducing the reconstruction error on normal points (compared to MSE), it enhances the contrast between normal and abnormal regions, making the anomalous segments stand out more clearly. This behavior directly contributes to improved point wise precision, which is a crucial metric in many practical applications.

Furthermore, MaxDiff aligns with the evolving goal in TSAD: moving beyond detection toward interpretable, fine-grained localization, especially in settings where actionable insights depend not just on the presence of an anomaly, but on knowing exactly when and where it occurs.


    \clearpage
    \section{Models}\label{sec:other-models}
In this section, we describe the baseline models used in our experiments along with their hyperparameter settings. 

\subsection{Reconstruction-Based}
Our implementations are based on the code provided by the TranAD~\citep{TranAD} repository\footnote{\url{https://github.com/imperial-qore/TranAD/}}, with minor modifications for ease of use. 
These models were chosen because they are well-established deep learning approaches that, like our method, are reconstruction-based. 
Additionally, they perform competitively on univariate time series, sometimes surpassing more recent alternatives\footnote{\url{https://thedatumorg.github.io/TSB-AD/}}.

\subsubsection{Deep AutoEncoding Gaussian Mixture Model (DAGMM)}
DAGMM is an unsupervised anomaly detection model that integrates a deep AutoEncoder with a Gaussian Mixture Model (GMM) to jointly learn representations and estimate their density \citep{DAGMM}. 
Sliding windows of univariate time series data ($w=5$) are compressed into an $8$-dimensional latent space ($c=8$), from which GMM parameters are estimated. The loss function combines reconstruction error, GMM negative log-likelihood, and covariance regularization. We optimize using the AdamW optimizer with a learning rate of $10^{-4}$ and weight decay of $10^{-5}$, along with a learning rate scheduler that decays the rate by 10\% every 5 epochs. Anomalies are detected based on the likelihood of latent representations under the fitted GMM.

\subsubsection{Multivariate Anomaly Detection with Generative Adversarial Networks (MAD-GAN)}
MAD-GAN is an unsupervised framework for multivariate anomaly detection based on GANs~\citep{goodfellow2020generative} which was introduced by \citet{MAD-GAN}. 
It utilizes multiple LSTM-based generators to model diverse temporal patterns and a shared discriminator to distinguish between real and generated sequences. Sliding windows of size $w=5$ are used to preserve temporal dependencies. The model is trained with the AdamW optimizer using a learning rate of $10^{-4}$ and weight decay of $10^{-5}$. Anomaly detection is based on a combination of reconstruction errors from all generators and the discriminator’s confidence scores, improving robustness and mitigating mode collapse.

\subsubsection{UnSupervised Anomaly Detection (USAD)}
USAD consists of two adversarially trained AutoEncoders, $\text{AE}_1$ and $\text{AE}_2$, which share a common decoder \citep{USAD}. $\text{AE}_1$ reconstructs the input (sliding windows with $w=5$), while $\text{AE}_2$ learns to differentiate real inputs from reconstructions. Training proceeds in two phases: first, both AutoEncoders are trained to reconstruct the input; second, $\text{AE}_2$ is trained adversarially to discriminate $\text{AE}_1$'s outputs while $\text{AE}_1$ attempts to fool it. We use the AdamW optimizer with a learning rate of $10^{-4}$ and weight decay of $10^{-5}$, along with a step scheduler. Anomaly scores are derived from reconstruction errors and adversarial feedback.

\subsubsection{Convolutional AutoEncoder with Memory (CAE-M)}
The CAE-M model combines convolutional layers with a memory mechanism to model temporal data \citep{CAE-M}. Input windows of size $w$ are encoded into hierarchical features using three convolutional layers with sigmoid activations. These are decoded using symmetric transposed convolutions. The model is trained to minimize mean squared error (MSE) using the AdamW optimizer with a learning rate of $10^{-3}$ and weight decay of $10^{-5}$. Anomalies are identified based on reconstruction errors.

\subsubsection{LSTM AutoEncoder (LSTM-AE)}
As a baseline for our SAE model, we use an LSTM-based AutoEncoder with an identical architecture to that of the SAE, excluding the stationary loss term. The encoder consists of three LSTM layers with hidden sizes $32$, $32$, and $16$, while the decoder mirrors this with hidden sizes $16$, $32$, and $32$. The model is trained using MSE loss and the AdamW optimizer with a learning rate of $10^{-4}$ and weight decay of $10^{-5}$.

\subsubsection{LSTM Variational AutoEncoder (LSTM-VAE)}
This model extends LSTM-AE with a Variational AutoEncoder (VAE)~\citep{kingma2013auto} formulation. 
In addition to MSE loss, it incorporates a KL-divergence loss on the latent space. The architecture and optimizer settings are the same as those used in LSTM-AE.

\subsubsection{LSTM Stationary AutoEncoder (LSTM-SAE)}
The SAE architecture mirrors LSTM-AE, using LSTM layers of sizes $32$, $32$, $16$ in the encoder and $16$, $32$, $32$ in the decoder. In addition to the MSE loss, we include a stationary loss term on the latent space to encourage invariance over time. Optimization is done using AdamW with a learning rate of $10^{-4}$ and weight decay of $10^{-5}$.

\subsubsection{Fourier AutoEncoder (FAE)}
As described in Section~\ref{sec:fae}, the FAE applies an FFT to input windows of size $w=256$, resulting in a $512$-dimensional input to the encoder. The encoder comprises fully connected layers with dimensions $512$, $256$, $256$, and $32$, while the decoder mirrors this with layers of size $32$, $256$, $256$, and $512$. ReLU activations are used after each layer except the final output. The model is trained with the AdamW optimizer (learning rate $10^{-4}$, weight decay $10^{-5}$) using the composite loss defined in Equation~\ref{eq:composite_loss}.

\subsection{Forecasting-Based}
Our implementations are based on the code provided by the FEDformer~\citep{zhou2022fedformer} repository\footnote{\url{https://github.com/MAZiqing/FEDformer}}, with minor modifications for ease of use.
\citet{lim2020tft} implement their code in TensorFlow as opposed to PyTorch.
Therefore, with the help of LLMs, their code was modified to be compatible with the rest of the baselines.
Unfortunately, there is no public implementation of PSformer by \citet{wang2025psformer}.
With the assistance of LLMs, this too was implemented.
All these models used a window size of $w=96$ alongside the AdamW optimizer with a learning rate of $10^{-4}$ and weight decay of $10^{-5}$.

\subsubsection{Temporal Fusion Transformer (TFT)}
TFT combines recurrent layers for local temporal processing with self-attention for longer-range dependencies in multi-horizon forecasting \citep{lim2020tft}. It also includes gating and variable-selection components to regulate the information used for prediction. We adapt the model to univariate time series using the shared forecasting-baseline settings described above.

\subsubsection{Informer}
Informer is a Transformer-based model designed for long-sequence time series forecasting \citep{zhou2021informer}. 
It uses ProbSparse self-attention and attention distilling to process long input sequences efficiently, together with a decoder that generates the forecast in one forward pass. 
We adapt the model to univariate time series using the shared forecasting-baseline settings described above.

\subsubsection{Autoformer}
Autoformer is a forecasting model that incorporates progressive seasonal-trend decomposition into its architecture \citep{wu2021autoformer}. Its Auto-Correlation mechanism identifies and aggregates dependencies associated with recurring temporal patterns. We adapt the model to univariate time series using the shared forecasting-baseline settings described above.

\subsubsection{FEDformer}
FEDformer combines seasonal-trend decomposition with frequency-enhanced Transformer layers for long-term time series forecasting \citep{zhou2022fedformer}. 
Its frequency-domain operations model temporal patterns while the decomposition separates trend and seasonal components. We adapt the model to univariate time series using the shared forecasting-baseline settings described above.

\subsubsection{PSformer}
PSformer is a parameter-efficient Transformer for time series forecasting introduced by \citet{wang2025psformer}. 
It uses parameter sharing to reduce the number of trainable parameters and segment attention to capture dependencies between portions of the input sequence. 
We adapt the model to univariate time series using the shared forecasting-baseline settings described above.

    \clearpage
    \section{Datasets}\label{sec:datasets}

TSB-AD was introduced by \citet{TSB} to address the biases in current benchmarking practices, as
well as the issues stemming from flawed datasets and evaluation measures. We provide the first
large-scale, manually curated dataset spanning 40 datasets, along with a collection of 40 anomaly
detection algorithms and an investigation of 10 evaluation measures. We believe that TSB-AD can
serve as a reliable testbed with high-quality datasets (TSB-AD-U/M) and an accurate evaluation
measure (VUS-PR), and we advocate for continued efforts in refining dataset creation practices.

One of the core challenges in time series anomaly detection is the absence of a well-defined evaluation framework. 
Unlike classification tasks, where unit tests produce binary outcomes and the correctness of predictions is straightforward to assess, anomalies in time series often span intervals, making it unclear how to evaluate partial correctness. 
This ambiguity affects the interpretability of standard metrics like AUC. 
To address this, we define a standardized time series unit test, a controlled benchmark scenario designed to ensure consistency and reduce noise in evaluation. 
Each unit test satisfies the following criteria:
\begin{enumerate}
    \item The training data contains no anomalies.
    \item Training sequences must be longer than a predefined threshold (in this paper: over 800 time steps).
    \item The test sequence must contain exactly one anomaly event.
    \item The test sequence must also include non-anomalous data; it should not consist of only the anomaly event.
\end{enumerate}
This setup enables a more focused assessment of model performance by reducing confounding factors that often undermine the clarity of anomaly detection evaluation.
To construct this, we curated a subset of sequences from the TSB-AD-U dataset \citep{TSB}
, selecting only those sequences that satisfied our unit test criteria.
Starting from 870 univariate sequences, we filtered out any series that violated key assumptions. 
This led to the removal of several datasets in their entirety, including the Yahoo dataset \cite{Yahoo}, which consistently failed to meet our standards. 
The resulting collection was then split into two parts: (1) a \textit{Complete} set of 311 sequences that already conformed to our requirements and were used as-is for the main evaluation; and (2) a \textit{Cropped} set consisting of additional sequences that were carefully trimmed to satisfy the defined criteria. 
For time series that did not initially satisfy our four-stage criteria for inclusion in the \textit{Complete} group, we first detect anomalous events (consecutive time points labeled anomalous without interruptions). 
We then extract segments immediately following each anomalous event, extending until just before the subsequent anomalous event, ensuring each segment contains exactly one anomalous event. 
Each extracted segment is individually re-evaluated using the four-stage criteria; segments meeting these criteria are added to the \textit{Cropped} group, while the others are discarded.

\begin{table}[ht!]
    \centering
    \small
    \caption{
              The anomaly types from left to right are \textit{Amplitude Change}, \textit{Flat}, \textit{Frequency Change}, \textit{Inverted Cycle}, \textit{Local Peak or Drop}, \textit{Missing Peak or Drop}, \textit{Noise}, \textit{Outlier or Unusual Pattern}, \textit{Reversed Cycle}, and \textit{Time Shift or Time Warping}.
    }
    \label{tab:datasets-by-type}
    \resizebox{\textwidth}{!}{
    \begin{tabular}{lccccccccccc}
        \hline
        \textbf{Model}
        & \rotatebox{0}{Amplitude} 
        & \rotatebox{0}{Flat} 
        & \rotatebox{0}{Frequency}
        & \rotatebox{0}{Inverted}
        & \rotatebox{0}{Local}
        & \rotatebox{0}{Missing}
        & \rotatebox{0}{Noise} 
        & \rotatebox{0}{Outlier} 
        & \rotatebox{0}{Reversed}
        & \rotatebox{0}{Time} 
        & \textbf{All} \\
        \hline
        {IOPS} & 0 & 0 & 0 & 0 & 0 & 0 & 0 & 1 & 0 & 0 & \textbf{1} \\
        {MSL} & 0 & 1 & 0 & 0 & 0 & 0 & 0 & 1 & 0 & 0 & \textbf{2} \\
        {NAB} & 1 & 0 & 0 & 0 & 0 & 0 & 0 & 1 & 0 & 0 & \textbf{2} \\
        {OPP} & 4 & 12 & 0 & 0 & 0 & 0 & 0 & 0 & 0 & 0 & \textbf{16} \\
        {SMAP} & 4 & 5 & 0 & 0 & 3 & 0 & 0 & 3 & 0 & 0 & \textbf{15} \\
        {SMD} & 13 & 0 & 0 & 0 & 1 & 0 & 0 & 7 & 0 & 0 & \textbf{21} \\
        {SVDB} & 0 & 0 & 0 & 0 & 0 & 0 & 0 & 1 & 0 & 0 & \textbf{1} \\
        {UCR} & 33 & 10 & 14 & 19 & 16 & 17 & 20 & 39 & 23 & 35 & \textbf{226} \\
        {WSD} & 2 & 0 & 0 & 0 & 25 & 0 & 0 & 0 & 0 & 0 & \textbf{27} \\
        \hline
        \textbf{Total} & \textbf{57} & \textbf{28} & \textbf{14} & \textbf{19} & \textbf{45} & \textbf{17} & \textbf{20} & \textbf{53} & \textbf{23} &  \textbf{35} & \textbf{311} \\
        \hline
    \end{tabular}
    }
\end{table}

In order to have a fair comparison with other methods, we report our results on the \textbf{Complete TSB-AD} dataset, leaving the proposed \textbf{Cropped TSB-AD} dataset to be further studied in future works.
Table~\ref{tab:datasets-by-type} shows the number of anomaly types per dataset.
The anomaly types are discussed further in Appendix~\ref{sec:anomaly-types}.
    \clearpage
    \section{Experiments}\label{sec:additional-experiments}

We provide our results with more details here.
First we explore the computational cost of different methods.
Next we provide the detailed results for our experiments grouped by anomaly types and different datasets in the TSB-AD.

\subsection{Settings}\label{sec:settings}

\paragraph{Datasets.}
We evaluate our models on a broad selection of datasets from the TSB-AD benchmark~\citep{TSB}. 
To enable consistent and interpretable evaluation, we define a unit test for each sequence. 
Specifically, we retain only those sequences where the training set is clean and sufficiently long, and the test set contains exactly one anomaly event with normal regions that are representative and free of contamination. 
We also discard any sequences that are artificially segmented or lack temporal continuity. 
Time series that violate the above criteria are excluded. 
All sequences are z-normalized as a preprocessing step. Further dataset details are provided in Appendix~\ref{sec:datasets}.

\paragraph{Baseline Models.}
We compare our models against reconstruction-based methods such as DAGMM \citep{DAGMM}, MAD-GAN \citep{MAD-GAN}, USAD \citep{USAD}, CAE-M \citep{CAE-M} alongside three LSTM-based AutoEncoders including LSTM-AE, LSTM-VAE and LSTM-SAE. 
In addition to FAE and FourStat, we also demonstrate the performance of forecasting-based models such as TFT~\citep{lim2020tft}, Informer~\citep{zhou2021informer}, Autoformer~\citep{wu2021autoformer}, FEDformer~\citep{zhou2022fedformer} and PSformer~\citep{wang2025psformer}.
The Adaptive FourStat model uses different window sizes based on the procedure described in Appendix~\ref{sec:adaptive-window}.
More details about these models are available in Appendix~\ref{sec:other-models}.

\paragraph{Hyperparameters.}
We train our models for 200 epochs with early stopping and use a batch size of 512.
We use the AdamW optimizer~\citep{AdamW} with a learning rate of $10^{-4}$ and weight decay of $10^{-5}$ unless the original papers suggest using different hyperparameters or schedulers.
DAGMM, MAD-GAN, USAD, and CAE-M have a window size of 5, LSTM-based AutoEncoders have a window size of 32 and FAE and FourStat have a window size of 128.
In addition to evaluating models with fixed window sizes, we also run FourStat with Adaptive Context Selection which is explained in Appendix~\ref{sec:adaptive-window}.
More details about the internal architecture of these models and their hyperparameters are provided in Appendix~\ref{sec:other-models}.

\paragraph{Evaluation Metrics.}
As discussed in Section~\ref{sec:pp_eval}, we report the performance of these models in terms of the Area Under the Precision-Recall Curve, F1-Score and Perfect Points. 
We report the number of Perfect Points per anomaly type in Table~\ref{tab:results-by-type} and the rest of results are available in Appendix~\ref{sec:additional-experiments}.

\subsection{Computational Costs}\label{sec:comp-cost}
The baselines with recurrent layers, e.g. RNNs and LSTMs, suffer from high computational costs due to the autoregressive nature of architecture.
This prohibits the use of recurrent architectures with larger window sizes.
In contrast, the FAE architecture which uses fully connected layers is more easily parallelizable.
Despite using larger window sizes, and the doubling of the input dimensions due to the Fourier transform, the FAE outperforms all autoregressive recurrent architectures.
As a result, the computational cost of the FourStat model is primarily dominated by its recurrent SAE component which performs on par with other baselines on similar window sizes.

In accordance with the original hyperparameters, our FAE has a lower training time to methods with much smaller windows due to its use of non-sequential layers even though it has two orders of magnitude more parameters. 
The LSTM-based AutoEncoders such as AE, VAE and SAE perform similarly and because of their sequential nature and are generally much more computationally expensive. 
The non-autoregressive nature of FAE provides a considerable gap against other recurrent architectures at similar or even smaller window sizes which makes it an ideal candidate for tasks where computational costs can become a bottleneck.
Moreover, the low cost of training FAEs, makes them ideal candidates for our adaptive window size selection scheme described in Appendix~\ref{sec:adaptive-window}.

While the FAE architecture is faster than autoregressive models, since it doesn't share the weights across different timesteps like recurrent layers it has a higher number of trainable parameters.
Despite using the fully connected layers, FAEs memory usage remains low due to the limited number of layers and relative low dimensionality.
It is noteworthy that unlike the time complexity of training models, which take very long even on modern accelerators, the memory demands of all methods including our FAE and FourStat are met even with modest computational resources. 
The true challenge in terms of computational complexity therefore, is not memory but time.

\subsection{Detailed results}
We report the detailed results of our models for each anomaly type in Table~\ref{tab:detailed}. 
We use the \textbf{AUC-ROC}, \textbf{F1-Score}, and \textbf{Perfect Point (PP)} metrics for evaluating our models. 
In Table~\ref{tab:detailed} we provide the Perfect Point ratio for different anomaly types, the real numbers are available in Table~\ref{tab:results-by-type}.

\begin{table*}[!ht]
    \centering 
    \caption{Detailed results based on anomaly type. The best results are in \textbf{bold} and the second best are \underline{underlined}.} 
    \resizebox{\textwidth}{!}{
    \begin{tabular}{l l ccc c ccc c ccc c ccc}
    \hline                                       
    \multirow{2}{*}{Method} &  & \multicolumn{3}{c}{Amplitude Change} &  & \multicolumn{3}{c}{Flat Region} &  & \multicolumn{3}{c}{Frequency Change} &  & \multicolumn{3}{c}{Inverted Cycle} \\
    & & AUC & F1 & PP & & AUC & F1 & PP & & AUC & F1 & PP & & AUC & F1 & PP \\
    \hline 
    
    DAGMM &     & 0.55 & 0.19 & 0.44 & & 0.73 & \underline{0.39} & 0.43 & & 0.53 & 0.06 & 0.26 & & 0.69 & 0.05 & 0.36 \\
    MAD-GAN &   & 0.54 & 0.21 & 0.4 & & 0.72 & \underline{0.39} & \underline{0.48} & & 0.53 & 0.06 & 0.26 & & 0.67 & 0.03 & 0.36 \\
    USAD &      & 0.54 & \textbf{0.28} & 0.44 & & 0.72 & \textbf{0.4} & 0.43 & & 0.51 & 0.09 & 0.26 & & 0.73 & 0.04 & 0.42 \\
    CAE-M &     & 0.58 & 0.17 & 0.44 & & 0.73 & 0.35 & 0.43 & & 0.53 & 0.06 & 0.29 & & 0.69 & 0.07 & 0.36 \\
    LSTM-AE &   & 0.58 & 0.21 & 0.44 & & \underline{0.74} & \textbf{0.4} & 0.43 & & 0.55 & \underline{0.1} & 0.31 & & \textbf{0.77} & \underline{0.17} & \underline{0.47} \\
    LSTM-VAE &  & 0.61 & 0.16 & 0.44 & & \underline{0.74} & \underline{0.39} & 0.38 & & 0.56 & 0.06 & 0.2 & & 0.59 & 0.06 & 0.05 \\
    LSTM-SAE &  & 0.61 & 0.16 & 0.44 & & \textbf{0.75} & \underline{0.39} & 0.38 & & 0.57 & 0.07 & 0.23 & & 0.43 & 0.04 & 0.05 \\
    FAE &       & \underline{0.76} & 0.17 & \underline{0.55} & & 0.71 & 0.28 & \underline{0.48} & & \textbf{0.71} & \textbf{0.14} & \underline{0.37} & & 0.71 & 0.06 & 0.42 \\
    FourStat &  & \textbf{0.78} & \underline{0.23} & \textbf{0.56} & & 0.73 & \textbf{0.4} & \textbf{0.52} & & \underline{0.67} & \textbf{0.14} & \textbf{0.57} & & \underline{0.75} & \textbf{0.22} & \textbf{0.52} \\
    
    \hline 
    \multirow{2}{*}{Method} &  & \multicolumn{3}{c}{Local Peak/Drop} &  & \multicolumn{3}{c}{Missing Peak/Drop} &  & \multicolumn{3}{c}{Noise} &  & \multicolumn{3}{c}{Outlier/Unusual}\\
    & & AUC & F1 & PP & & AUC & F1 & PP & & AUC & F1 & PP & & AUC & F1 & PP\\
    \hline 
    
    DAGMM &     & 0.58 & \textbf{0.12} & 0.46 & & 0.46 & 0.0 & 0.17 & & 0.52 & 0.01 & 0.05 & & 0.52 & 0.08 & 0.26 \\
    MAD-GAN &   & 0.56 & 0.07 & 0.39 & & 0.43 & 0.0 & 0.17 & & 0.53 & 0.01 & 0.05 & & 0.52 & 0.08 & 0.25 \\
    USAD &      & 0.54 & \underline{0.11} & 0.46 & & 0.44 & 0.0 & 0.17 & & 0.51 & 0.01 & 0.05 & & 0.52 & 0.09 & 0.27 \\
    CAE-M &     & 0.6 & 0.1 & 0.46 & & 0.44 & 0.0 & 0.17 & & 0.56 & 0.01 & 0.05 & & 0.54 & 0.08 & 0.26 \\
    LSTM-AE &   & 0.63 & 0.05 & 0.46 & & 0.48 & 0.0 & 0.17 & & 0.54 & 0.01 & 0.15 & & 0.53 & 0.08 & 0.26 \\
    LSTM-VAE &  & 0.62 & 0.08 & 0.46 & & 0.44 & 0.0 & 0.0 & & 0.58 & 0.01 & 0.05 & & 0.55 & 0.07 & 0.24 \\
    LSTM-SAE &  & 0.61 & 0.09 & 0.46 & & 0.44 & 0.0 & 0.0 & & 0.57 & 0.01 & 0.05 & & 0.55 & 0.07 & 0.25 \\
    FAE &       & \underline{0.71} & 0.03 & \underline{0.5} & & \textbf{0.63} & \underline{0.04} & \underline{0.25} & & \textbf{0.7} & \underline{0.08} & \underline{0.25} & & \textbf{0.71} & \underline{0.1} & \underline{0.36} \\
    FourStat &  & \textbf{0.72} & 0.07 & \textbf{0.57} & & \underline{0.52} & \textbf{0.05} & \textbf{0.5} & & \underline{0.61} & \textbf{0.11} & \textbf{0.3} & & \underline{0.69} & \textbf{0.12} & \textbf{0.37} \\
    
    \hline 
    \multirow{2}{*}{Method} & & & & & & \multicolumn{3}{c}{Reversed Cycle} &  &  \multicolumn{3}{c}{Time Shift/Warping}\\
    & &  & & & & AUC & F1 & PP & & AUC & F1 & PP \\
    \hline 
    
    DAGMM &     & & & & & 0.49 & 0.05 & 0.0 & & 0.52 & 0.04 & 0.0 \\
    MAD-GAN &   & & & & & 0.45 & 0.05 & 0.0 & & 0.46 & 0.01 & 0.0 \\
    USAD &      & & & & & 0.45 & 0.03 & 0.0 & & 0.45 & 0.02 & 0.0 \\
    CAE-M &     & & & & & 0.54 & 0.04 & 0.0 & & 0.55 & 0.13 & 0.15 \\
    LSTM-AE &   & & & & & 0.47 & 0.0 & 0.0 & & 0.53 & 0.14 & 0.15 \\
    LSTM-VAE &  & & & & & 0.59 & 0.06 & 0.0 & & 0.56 & 0.11 & 0.08 \\
    LSTM-SAE &  & & & & & 0.63 & 0.04 & 0.25 & & 0.59 & 0.12 & 0.23 \\
    FAE &       & & & & & \textbf{0.94} & \underline{0.23} & \underline{0.75} & & \textbf{0.76} & \underline{0.23} & \underline{0.46} \\
    FourStat &  & & & & & \underline{0.92} & \textbf{0.33} & \textbf{1.0} & & \underline{0.75} & \textbf{0.26} & \textbf{0.54} \\
    \hline 
    \end{tabular}
    }
    \label{tab:detailed}
\end{table*}

While FourStat and FAE outperform other models across all anomaly types, the highest performance jumps are realized in \textit{Frequency Change}, \textit{Outlier or Unusual Pattern}, \textit{Reversed Cycle}, and \textit{Time Shift or Time Warping} anomaly types.
In these groups, viewing the signal through the frequency-domain or through larger windows has certain advantages.
Since many models are essentially trying to use deviation from mean to classify windows as anomalies, they're not very successful when it comes to frequency changes or reversed cycles, whereas incorporating frequency features makes it easier to detect anomalies.
Similarly, when used with sufficiently large windows, frequency-based models such as FAE and FourStat can detect time shift or time warping anomalies.
Figure~\ref{fig:time_shift} illustrates one example of these anomaly types and it is easy to see why methods with small window sizes will probably fail.

\subsection{Results per dataset}
In addition to showing the results based on anomaly types, we report them based on the original datasets which they belonged to in Table~\ref{tab:detailed-dataset}.

\begin{table*}[!ht]
    \small
    \centering
    \caption{Detailed results based on dataset. 
    The best results are in \textbf{bold} and the second best are \underline{underlined}.
    The number of time series in each dataset is available in Table~\ref{tab:datasets-by-type}.\\} 
    \resizebox{\textwidth}{!}{
    \begin{tabular}{l l ccc c ccc c ccc}
    \hline                                       
    \multirow{2}{*}{Method} &  & \multicolumn{3}{c}{IOPS \citep{IOPS}} &  & \multicolumn{3}{c}{MSL \citep{SMAP}} &  & \multicolumn{3}{c}{NAB \citep{NAB}} \\
    & & AUC & F1 & PP & & AUC & F1 & PP & & AUC & F1 & PP \\
    \hline 
    DAGMM &     & 0.66 & 0.03 & \textbf{1.0} & & 0.59 & \textbf{0.1} & \textbf{0.5} & & 0.67 & 0.56 & 0.5 \\
    MAD-GAN &   & 0.73 & 0.02 & \textbf{1.0} & & 0.59 & \textbf{0.1} & \textbf{0.5} & & 0.66 & 0.56 & 0.5\\
    USAD &      & 0.61 & 0.03 & \textbf{1.0} & & 0.6 & \textbf{0.1} & \textbf{0.5} & & 0.64 & 0.56 & 0.5 \\
    CAE-M &     & 0.67 & \underline{0.04} & \textbf{1.0} & & 0.56 & \textbf{0.1} & \textbf{0.5} & & 0.68 & 0.56 & \textbf{1.0} \\
    LSTM-AE &   & 0.84 & 0.02 & \textbf{1.0} & & 0.62 & \textbf{0.1} & \textbf{0.5} & & 0.56 & 0.72 & 0.5 \\
    LSTM-VAE &  & 0.7 & 0.03 & \textbf{1.0} & & 0.6 & \textbf{0.1} & \textbf{0.5} & & 0.67 & 0.56 & 0.5 \\
    LSTM-SAE &  & 0.7 & 0.03 & \textbf{1.0} & & \textbf{0.64} & \textbf{0.1} & \textbf{0.5} & & 0.67 & 0.56 & \textbf{1.0} \\
    FAE &       & \underline{0.88} & \underline{0.04} & \textbf{1.0} & & \underline{0.63} & 0.09 & 0.1 & & \underline{0.76} & \underline{0.81} & \textbf{1.0} \\
    FourStat &  & \textbf{0.99} & \textbf{0.05} & \textbf{1.0} & & 0.62 & \textbf{0.1} & \textbf{0.5} & & \textbf{0.81} & \textbf{0.86} & \textbf{1.0} \\
    \hline 
    \multirow{2}{*}{Method} &  & \multicolumn{3}{c}{OPP \citep{OPP}} &  & \multicolumn{3}{c}{SMAP \citep{SMAP}} &  & \multicolumn{3}{c}{SMD \citep{SMD}}\\
    & & AUC & F1 & PP & & AUC & F1 & PP & & AUC & F1 & PP\\
    \hline 
    DAGMM &     & 0.89 & 0.63 & \textbf{0.62} & & 0.63 & 0.25 & \underline{0.47} & & 0.63 & 0.39 & 0.86 \\
    MAD-GAN &   & 0.88 & 0.59 & \textbf{0.62} & & 0.44 & 0.2 & 0.4 & & 0.65 & \underline{0.53} & 0.86\\
    USAD &      & 0.89 & 0.61 & \textbf{0.62} & & 0.62 & \textbf{0.31} & \underline{0.47} & & 0.61 & \textbf{0.65} & \underline{0.9} \\
    CAE-M &     & 0.9 & 0.65 & \textbf{0.62} & & 0.64 & 0.2 & \underline{0.47} & & 0.67 & 0.32 & 0.86 \\
    LSTM-AE &   & 0.9 & 0.66 & \textbf{0.62} & & 0.62 & \textbf{0.31} & \underline{0.47} & & 0.74 & 0.38 & 0.86\\
    LSTM-VAE &  & \textbf{0.93} & \textbf{0.71} & \textbf{0.62} & & 0.63 & 0.21 & 0.4 & & 0.71 & 0.28 & 0.86 \\
    LSTM-SAE &  & \textbf{0.93} & \textbf{0.71} & \textbf{0.62} & & 0.64 & 0.2 & 0.4 & & 0.7 & 0.28 & 0.86\\
    FAE &       & 0.81 & 0.37 & 0.44 & & \underline{0.7} & 0.16 & \textbf{0.6} & & \underline{0.94} & 0.35 & \textbf{0.95} \\
    FourStat &  & 0.87 & 0.53 & \textbf{0.62} & & \textbf{0.71} & 0.25 & \underline{0.47} & & \textbf{0.98} & 0.32 & \underline{0.9} \\
    \hline 
    \multirow{2}{*}{Method} &  & \multicolumn{3}{c}{SVDB \citep{SVDB}} &  & \multicolumn{3}{c}{UCR \citep{UCR}} &  & \multicolumn{3}{c}{WSD \citep{WSD}}\\
    & & AUC & F1 & PP & & AUC & F1 & PP & & AUC & F1 & PP \\
    \hline 
    DAGMM &     & 0.64 & 0.05 & \textbf{1.0} & & 0.5 & 0.02 & 0.12 & & 0.57 & 0.2 & \underline{0.85}\\
    MAD-GAN &   & 0.64 & 0.05 & \textbf{1.0} & & 0.5 & 0.02 & 0.12 & & 0.57 & \underline{0.13} & 0.78\\
    USAD &      & 0.64 & 0.05 & \textbf{1.0} & & 0.49 & 0.03 & 0.13 & & 0.54 & 0.2 & 0.81\\
    CAE-M &     & 0.64 & 0.05 & \textbf{1.0} & & 0.52 & 0.03 & 0.13 & & 0.61 & \textbf{0.19} & \underline{0.85}\\
    LSTM-AE &   & 0.61 & 0.05 & \textbf{1.0} & & 0.51 & 0.04 & 0.14 & & 0.66 & 0.11 & \textbf{0.93}\\
    LSTM-VAE &  & 0.64 & 0.05 & \textbf{1.0} & & 0.53 & 0.03 & 0.1 & & 0.64 & 0.12 & \underline{0.85} \\
    LSTM-SAE &  & 0.64 & 0.05 & \textbf{1.0} & & 0.53 & 0.03 & 0.12 & & 0.64 & \underline{0.13} & \underline{0.85}\\
    FAE &       & \underline{0.81} & \underline{0.16} & \textbf{1.0} & & \textbf{0.69} & \underline{0.09} & \underline{0.37} & & \underline{0.71} & 0.06 & 0.74 \\
    FourStat &  & \textbf{0.88} & \textbf{0.22} & \textbf{1.0} & & \underline{0.66} & \textbf{0.12} & \textbf{0.46} & & \textbf{0.73} & 0.06 & 0.78 \\
    \hline 
    \end{tabular}
    }
    \label{tab:detailed-dataset}
\end{table*}

While the lowest Perfect Point ratio, for any model, belongs to the UCR anomaly archive \citep{UCR}, we consider it to be one of the better benchmarks available.
Most of the diversity in TSB-AD in terms of anomaly types comes from UCR and all \textit{Frequency Change}, \textit{Inverted Cycle}, \textit{Missing Peak or Drop}, \textit{Noise}, \textit{Reversed Cycle}, and \textit{Time Shift or Time Warping} anomaly types belong to the UCR dataset as illustrated in Table~\ref{tab:datasets-by-type}.
This also explains the large performance gap between previous models and FourStat in the UCR dataset.
Performance of most models across every other dataset is roughly in the same neighbourhood whereas in the UCR dataset, FourStat outperforms previous models by detecting nearly four times as many Perfect Points.
    \clearpage
    \section{Perfect Points and thresholding}\label{sec:ppt}

In this section we aim to answer whether it is possible to find appropriate thresholds when we have a Perfect Point.
More specifically we utilize a variety of popular thresholding mechanisms.

\subsection{Thresholding schemes}

\textbf{Static thresholding} is a rule-based approach that determines the anomaly threshold using the empirical distribution of anomaly scores from the training set. 
It assumes that the training set contains only normal samples and calculates the mean and standard deviation of these scores. 
A data point is then classified as an anomaly if its score exceeds the mean by more than a fixed multiple of the standard deviation, typically denoted as $k$ \citep{staticthresh}.
This method is simple, interpretable, and widely used, especially when the score distribution is approximately Gaussian. 
However, it lacks robustness in the presence of skewed or heavy-tailed distributions and may fail when the assumption of normality does not hold.

\textbf{Quantile thresholding} is a non-parametric method that sets the anomaly threshold based on a specified upper quantile of the anomaly scores observed in the training data \citep{SMAP}. 
Since the training set contains only normal samples, the idea is to treat the highest few percent of these scores as potential boundary cases, and classify any test score exceeding this quantile as anomalous. 
This method does not make any assumptions about the shape of the score distribution and is particularly useful when the distribution is asymmetric or heavy-tailed.
Its simplicity and robustness to outliers in the training data make it an attractive choice in many practical settings.

\textbf{Peaks-Over-Threshold (POT)} is a statistical technique rooted in Extreme Value Theory (EVT) that models the tail behavior of anomaly scores exceeding a high initial threshold \citep{potevtthresh}. 
It begins by selecting a quantile-based cutoff to identify the most extreme values (i.e., the tail), and fits a Generalized Pareto Distribution (GPD) to these exceedances. 
Using the fitted distribution, the method estimates a higher, data-driven threshold corresponding to a user-specified false alarm probability. 
POT is particularly effective for detecting rare, high-impact anomalies and provides a theoretically grounded way to differentiate extreme outliers from normal fluctuations. 
However, it requires careful parameter tuning and a sufficient number of exceedances for accurate estimation.

\textbf{Extreme value theory (EVT) with block maxima}, also known as the Generalized Extreme Value (GEV) approach, is another EVT-based technique that fits a statistical model to the distribution of block-wise maximum anomaly scores \cite{evtbmthresh}. 
The training data is divided into non-overlapping blocks of fixed size, and the maximum score from each block is collected to form a new sample of extreme values. 
A GEV distribution is then fitted to these maxima, and the anomaly threshold is derived as the score corresponding to a desired false positive rate under the fitted distribution. 
This method is suitable when anomalies are expected to be extreme in nature and arise sporadically. 
It is especially useful when the training set is large enough to form several representative blocks, ensuring a stable estimation of the GEV parameters.

\subsection{Results}
We employ different thresholding techniques and report what percentage of the time series marked as solved via the Perfect Point metric, can earn perfect F1-scores.
The results, averaged over different time series, are shown in Table~\ref{tab:ppt}.

\begin{table}[ht!]
    \caption{
        F1-scores for time series with Perfect Points for the FourStat model.
        The best results are in \textbf{bold} and the second best are \underline{underlined}.
        The anomaly types from left to right are \textit{Amplitude Change}, \textit{Flat}, \textit{Frequency Change}, \textit{Local Peak or Drop}, \textit{Missing Peak or Drop}, \textit{Noise}, \textit{Outlier or Unusual Pattern}, \textit{Reversed Cycle}, and \textit{Time Shift or Time Warping}.
    }
    \label{tab:ppt}
    \centering
    \begin{tabular}{l c c c c c c c c c}
        \hline
        \textbf{Method}
        & \rotatebox{0}{Amp} 
        & \rotatebox{0}{Flat} 
        & \rotatebox{0}{Freq}
        & \rotatebox{0}{Local}
        & \rotatebox{0}{Missing}
        & \rotatebox{0}{Noise} 
        & \rotatebox{0}{Outlier} 
        & \rotatebox{0}{Rev}
        & \rotatebox{0}{Time} \\
        \hline
        Static ($k=2$)     & 0.23 & 0.57 & 0.14 & 0.03 & 0.04 & 0.13 & 0.14 & 0.2 & 0.27  \\
       Static ($k=3$)     & 0.32 & 0.62 & 0.2 & 0.06 & 0.1 & 0.35 & 0.22 & 0.33 & 0.38 \\
       Static ($k=4$)     & 0.4 & \underline{0.66} & 0.27 & 0.12 & 0.18 & 0.22 & 0.31 & 0.49 & 0.46 \\
       Static ($k=5$)     & 0.48 & \textbf{0.7} & 0.35 & 0.17 & 0.3 & 0.24 & 0.4 & 0.61 & 0.5  \\
       Quantile     & 0.35 & 0.65 & 0.24 & 0.08 & 0.12 & 0.19 & 0.23 & 0.4 & 0.39 \\
       POT     & \textbf{0.52} & \underline{0.66} & \underline{0.44} & \underline{0.29} & \underline{0.34} & \textbf{0.44} & \textbf{0.49} & \underline{0.68} & \textbf{0.56} \\
       EVT     & \underline{0.5} & 0.28 & \textbf{0.46} & \textbf{0.3} & \textbf{0.45} & \textbf{0.44} & \underline{0.45} & \textbf{0.77} & \underline{0.52}  \\
        \hline
    \end{tabular}
\end{table}

Based on these results we believe that thresholding remains an important and open problem in TSAD, especially in an unsupervised setting.

    \clearpage
    \section{Group Robustness}\label{sec:anomaly-types}
Our method of categorizing different anomalies into the following groups was inspired by the previous work done by \citet{IsItWorthIt} on the UCR dataset~\citep{UCR}.
We manually categorized each time series based on their plots, which can be found in our code repository.
We decided to merge groups of peaks and drops since they're the same phenomenon only in reverse, peaks are sudden and sharp increases in values, whereas drops are sudden and sharp decreases.
We will discuss each anomaly type and provide examples for them from the TSB-AD dataset~\citep{TSB}.

\subsection{Amplitude Change}
\begin{figure}[h]
    \centering
    \subfigure[Increased Amplitude]{\label{fig:tsb305}\includegraphics[width=0.49\textwidth]{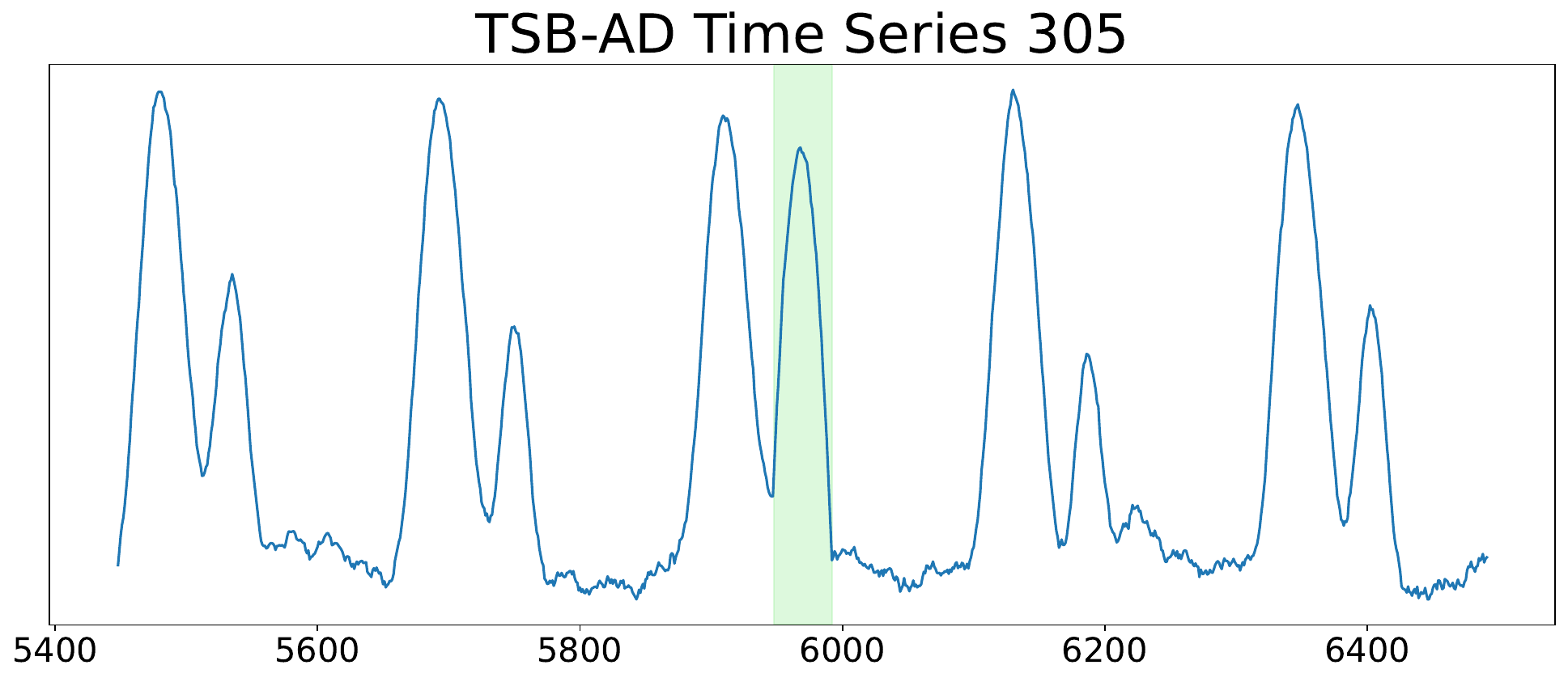}}
    \subfigure[Decreased Amplitude]{\label{fig:tsb497}\includegraphics[width=0.49\textwidth]{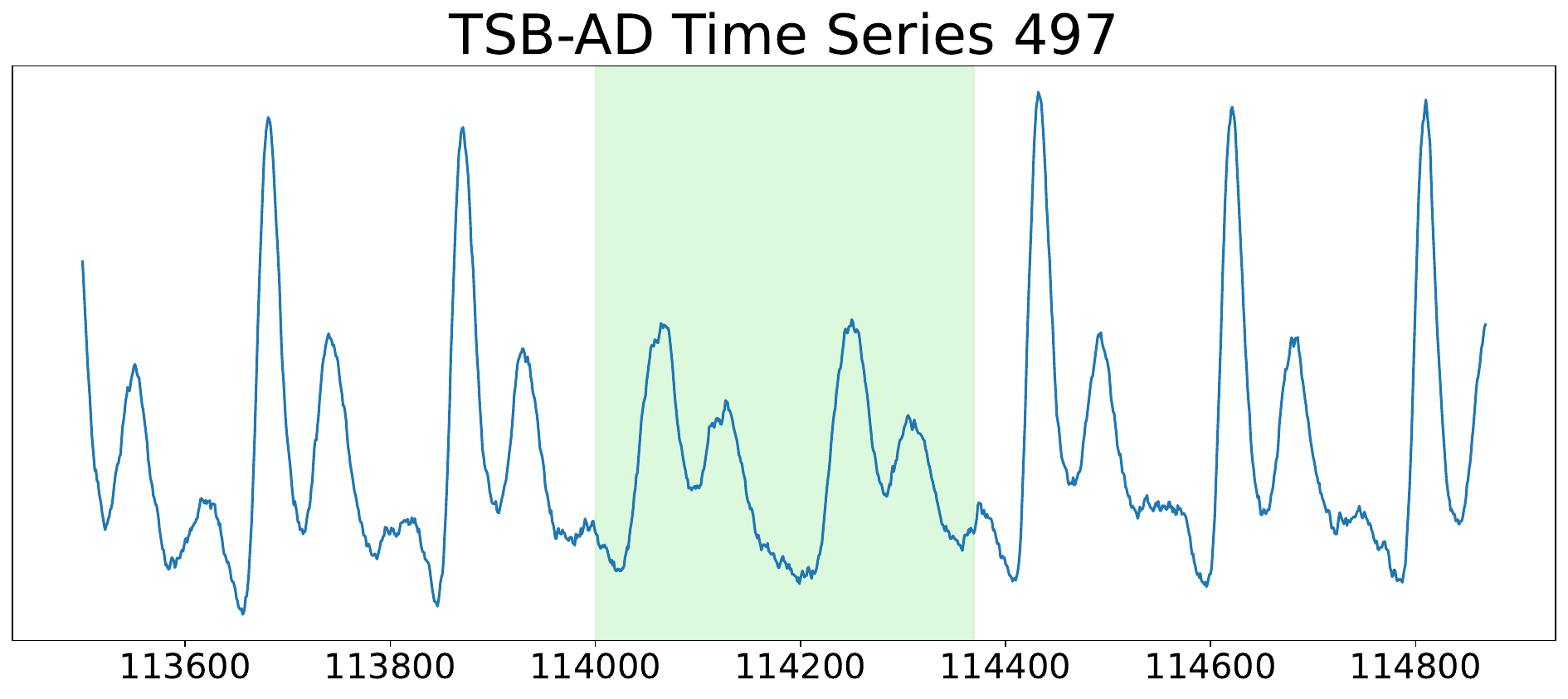}}
    \caption{
    Two examples of the \textit{amplitude change} anomaly type.
    Anomalous regions have been highlighted.
    It is not difficult to see that small window sizes might struggle to detect the anomalous region in  \subref{fig:tsb305} because it resembles the other peaks.
    }
    \label{fig:amp}
\end{figure}

In this category of anomalies the structure and seasonality of the signal remains what it was but the amplitude gets either increased or decreased, resulting in a different pattern.
While steep increases or decreases might be quite easy to detect, there are certain examples which can be tricky for models with small window sizes.
One such example is illustrated in Figure~\ref{fig:amp}.

\subsection{Flat Region}
\begin{figure}[h]
    \centering
    \subfigure[Average Maintained]{\label{fig:tsb458}\includegraphics[width=0.49\textwidth]{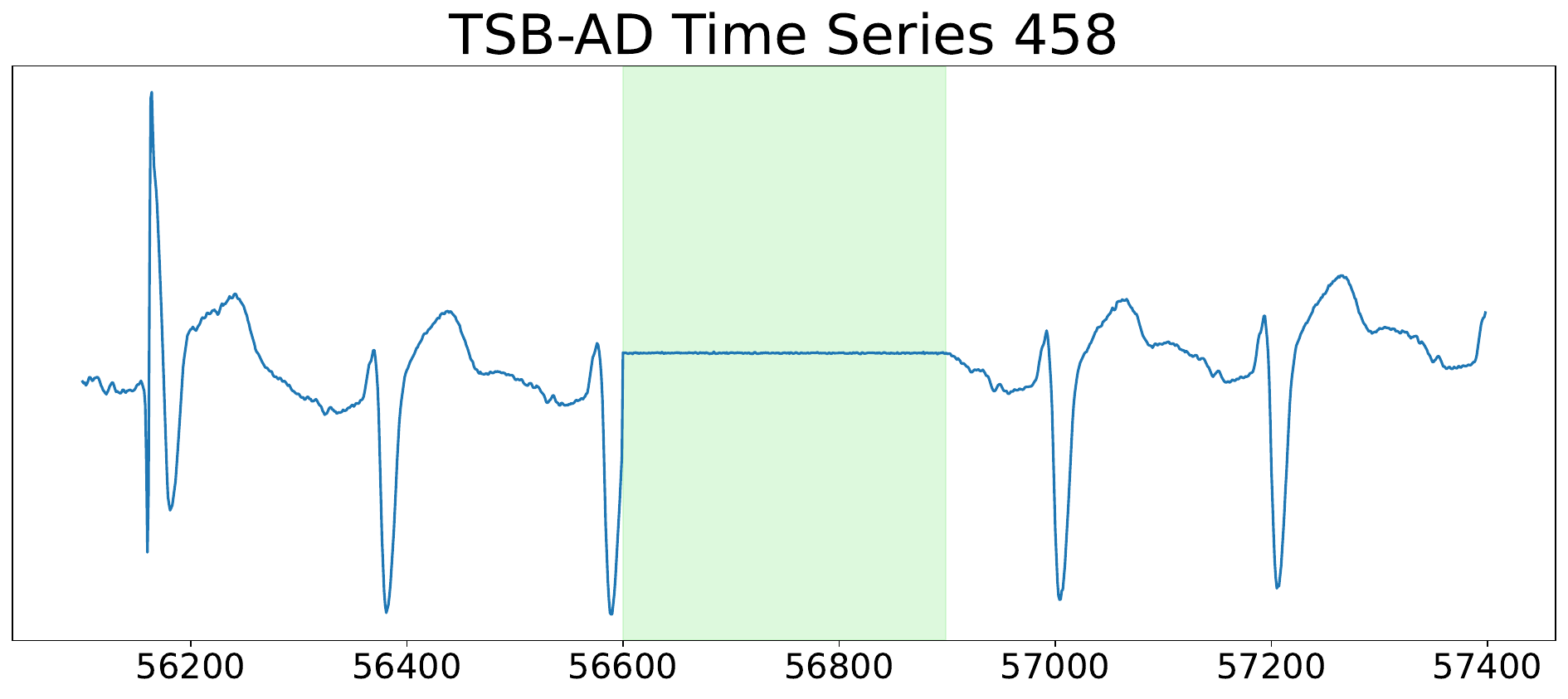}}
    \subfigure[Average Changed]{\label{fig:tsb459}\includegraphics[width=0.49\textwidth]{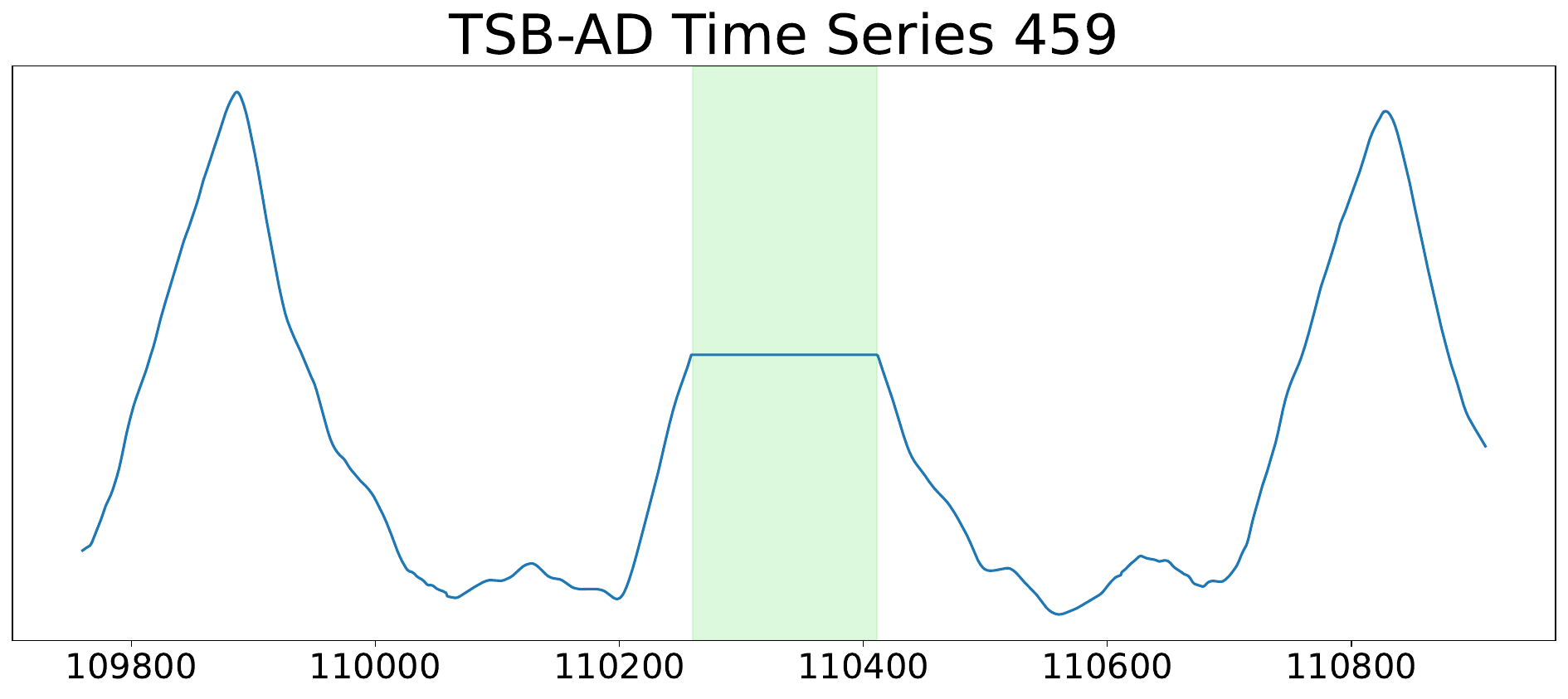}}
    \caption{
    Two examples of the \textit{flat region} anomaly type.
    Anomalous regions have been highlighted.
    Instead of seeing the seasonal pattern in \subref{fig:tsb458} the entire cycle has been replace with a flat region with the same mean as the signal. 
    In \subref{fig:tsb459} the peak has been replace with a flat region. 
    }
    \label{fig:flat}
\end{figure}

As the name suggests, when a region of the time series becomes constant.
Assuming the model is capable of learning the underlying structure of signal during training, detecting this anomaly type is not very difficult assuming other general problems such as identity mapping don't cause any problems.
This is one of the categories where our FourStat and FAE models perform slightly better than previous models and we don't see major boost to performance.
You can see what these anomalies look like in Figure~\ref{fig:flat}.

\subsection{Frequency Change}
For signals with seasonality, sometimes we a change in their frequency.
This often manifests as either the cycles getting shorter as a result of increasing the frequency, or cycles getting longer as a result of decreasing the frequency.
As it is expected, the FAE and FourStat models have an easier time detecting these anomalies because they view the signal via its frequency-domain.

\begin{figure}[h]
    \centering
    \subfigure[Increased Frequency]{\label{fig:tsb380}\includegraphics[width=0.49\textwidth]{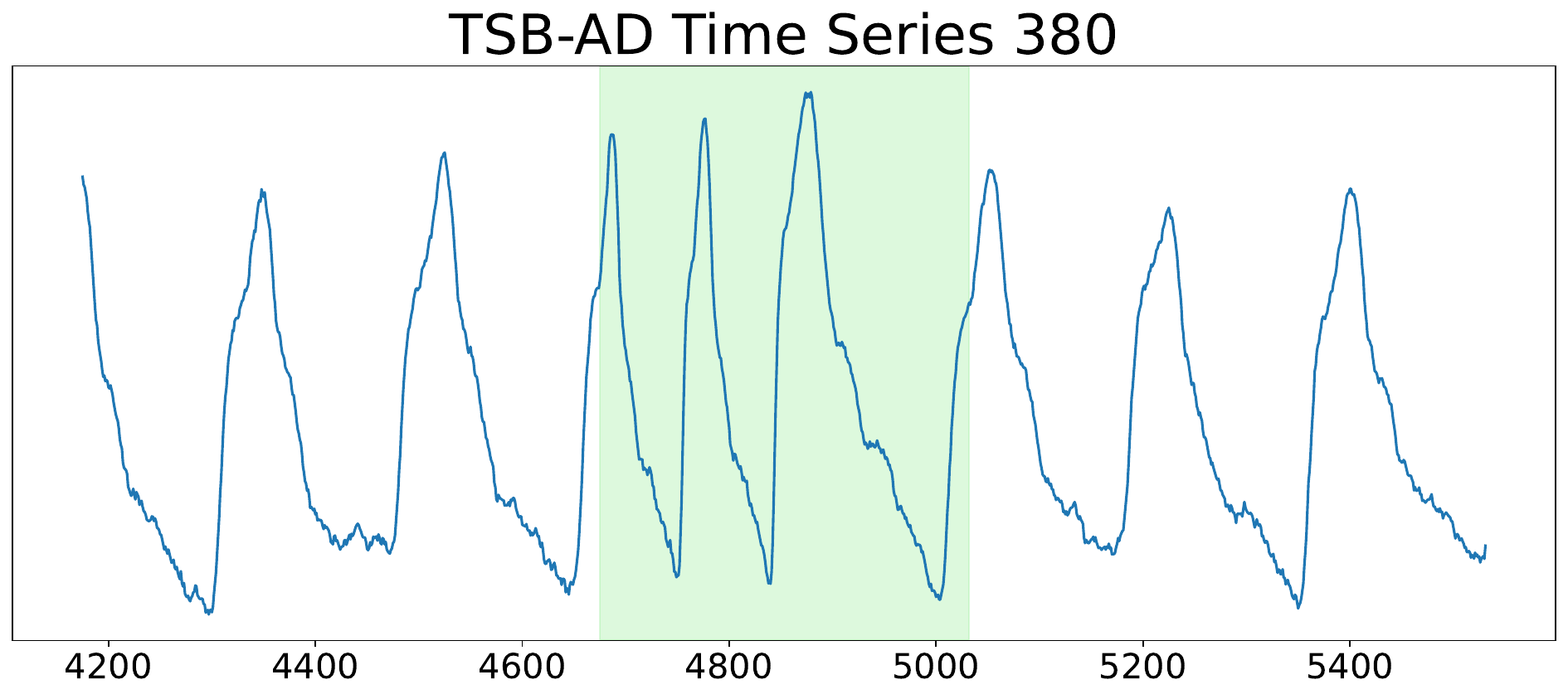}}
    \subfigure[Decreased Frequency]{\label{fig:tsb466}\includegraphics[width=0.49\textwidth]{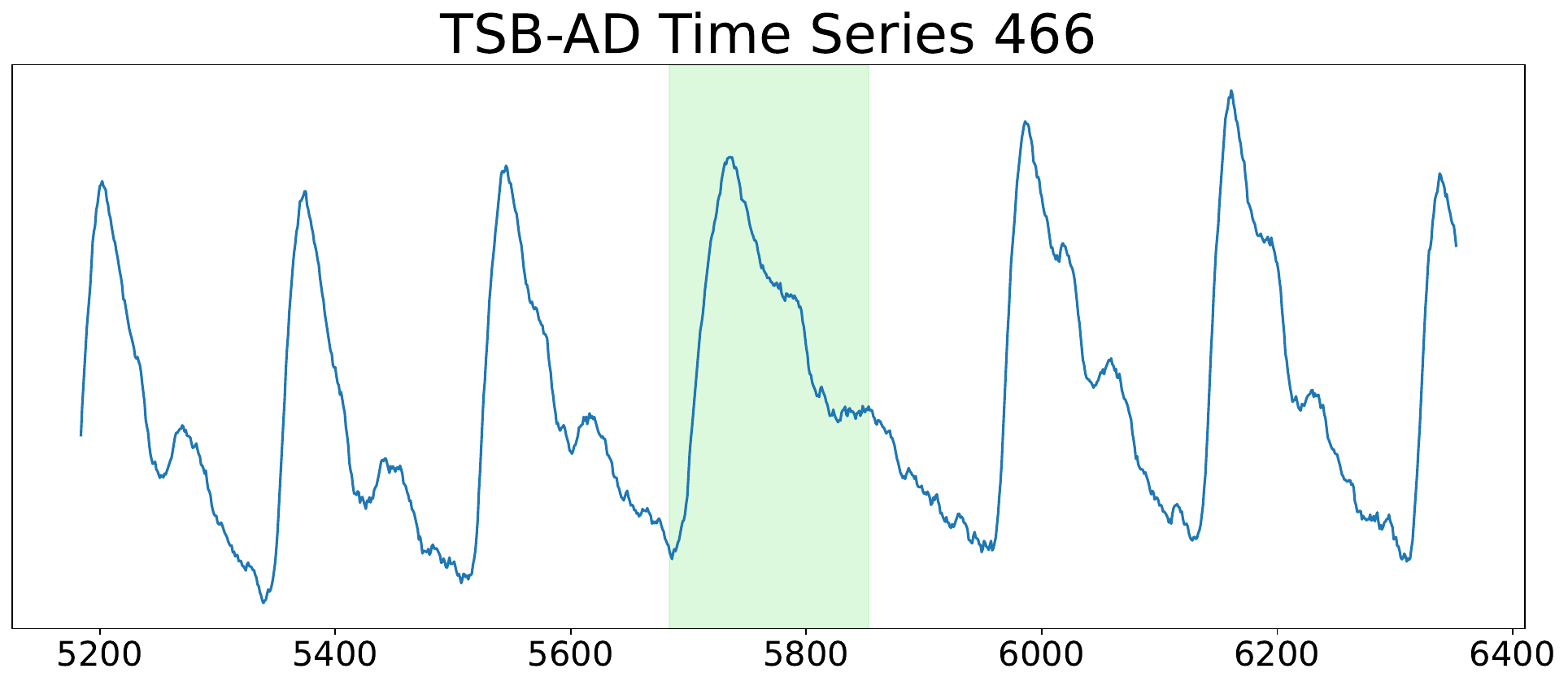}}
    \caption{
    Two examples of the \textit{frequency change} anomaly type.
    Anomalous regions have been highlighted.
    }
    \label{fig:freq}
\end{figure}

\subsection{Inverted Cycle}
This category of anomalies, a cycle in the signal gets inverted with respect to the signal mean.
Since we either have an increase or decrease in values, it is not challenging for previous models to detect.

\begin{figure}[h]
    \centering
    \subfigure[Increased Value]{\label{fig:tsb370}\includegraphics[width=0.49\textwidth]{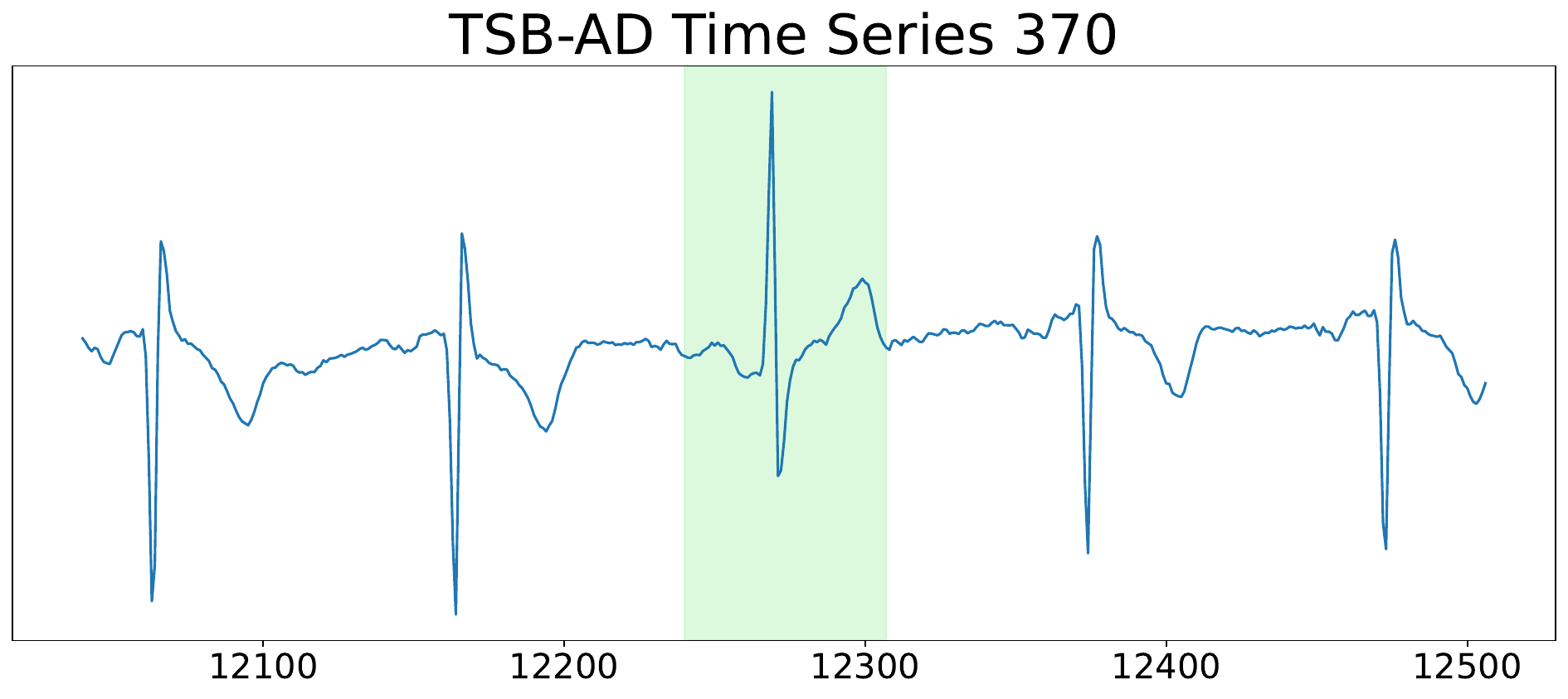}}
    \subfigure[Decreased Value]{\label{fig:tsb450}\includegraphics[width=0.49\textwidth]{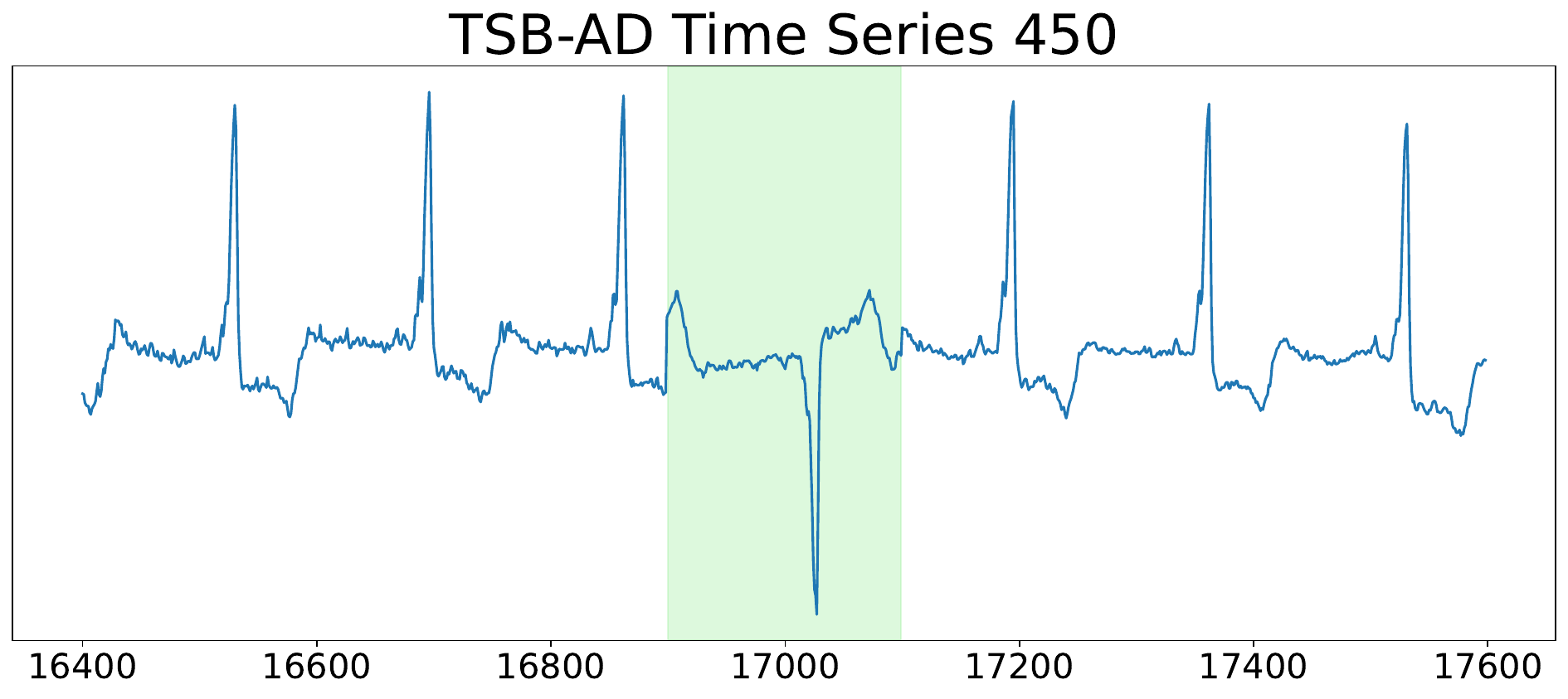}}
    \caption{
    Two examples of the \textit{inverted cycle} anomaly type.
    Anomalous regions have been highlighted.
    }
    \label{fig:inv}
\end{figure}

\clearpage
\subsection{Local Peak or Drop}
Sudden spikes in values where the we see radical change in the values are categorized as local peaks or drops.
These anomalies could be easily detected, assuming the model is capable of accurately predicting the mean, since they elicit high errors.

\begin{figure}[h]
    \centering
    \subfigure[Peak]{\label{fig:tsb410}\includegraphics[width=0.49\textwidth]{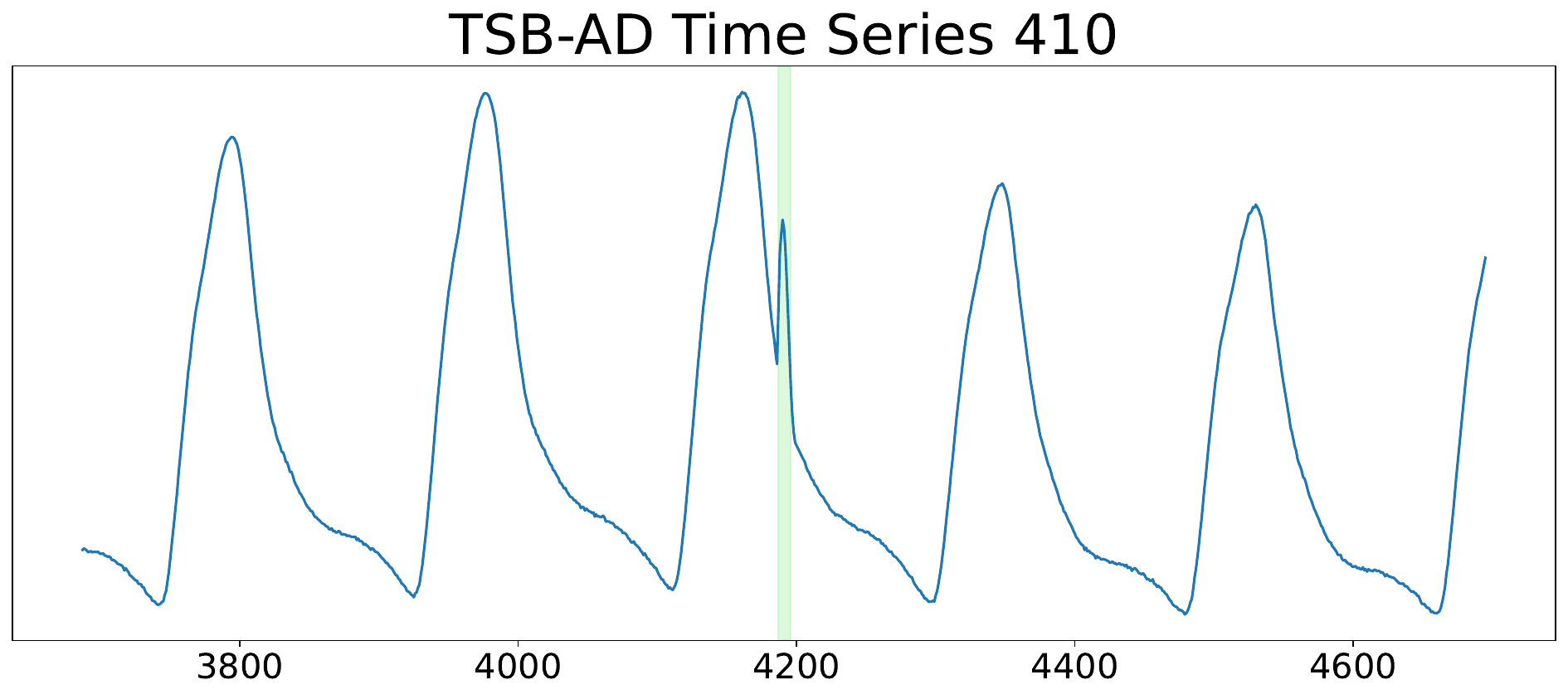}}
    \subfigure[Drop]{\label{fig:tsb439}\includegraphics[width=0.49\textwidth]{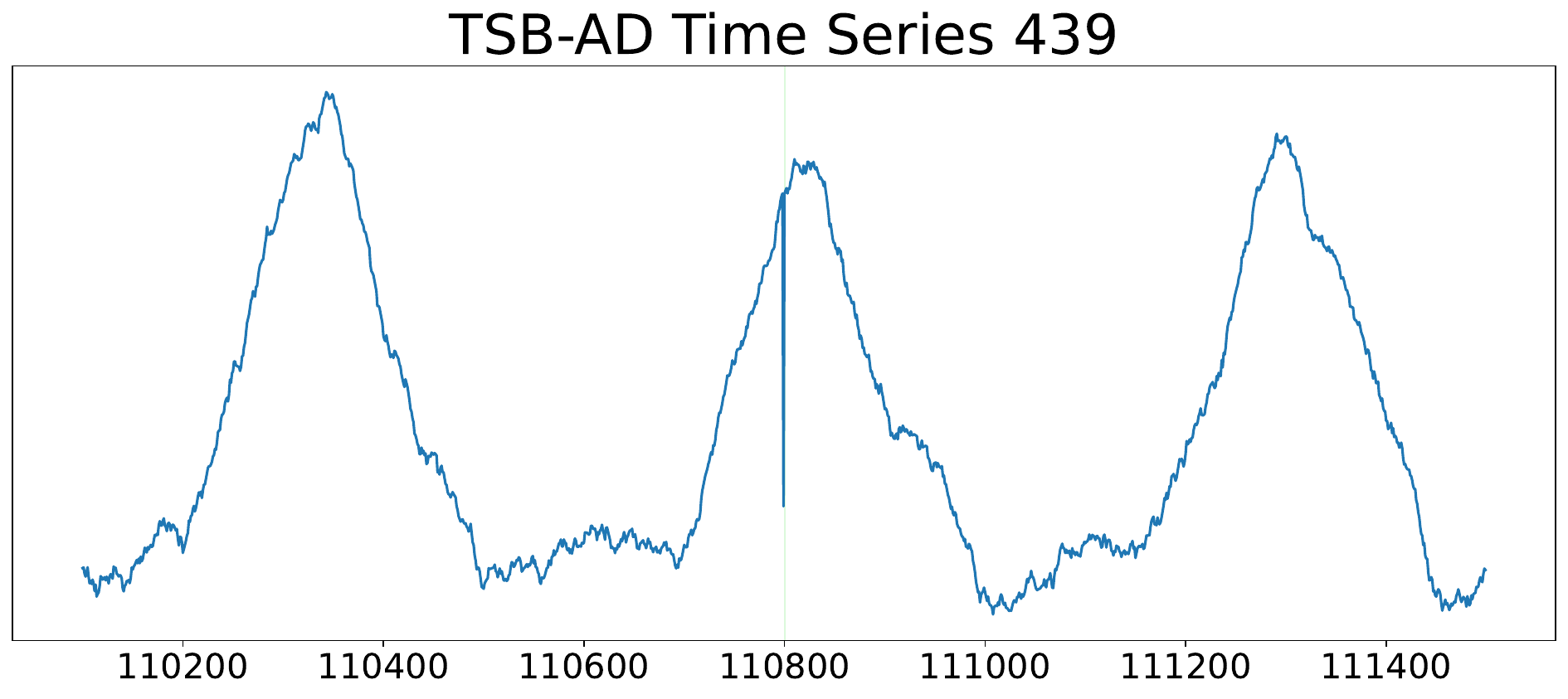}}
    \caption{
    Two examples of the \textit{local peak or drop} anomaly type.
    Anomalous regions have been highlighted.
    }
    \label{fig:local}
\end{figure}

\subsection{Missing Peak or Drop}
In this category of anomalies, a peak or drop present in other cycles is removed.

\begin{figure}[h]
    \centering
    \subfigure[Peak]{\label{fig:tsb325}\includegraphics[width=0.49\textwidth]{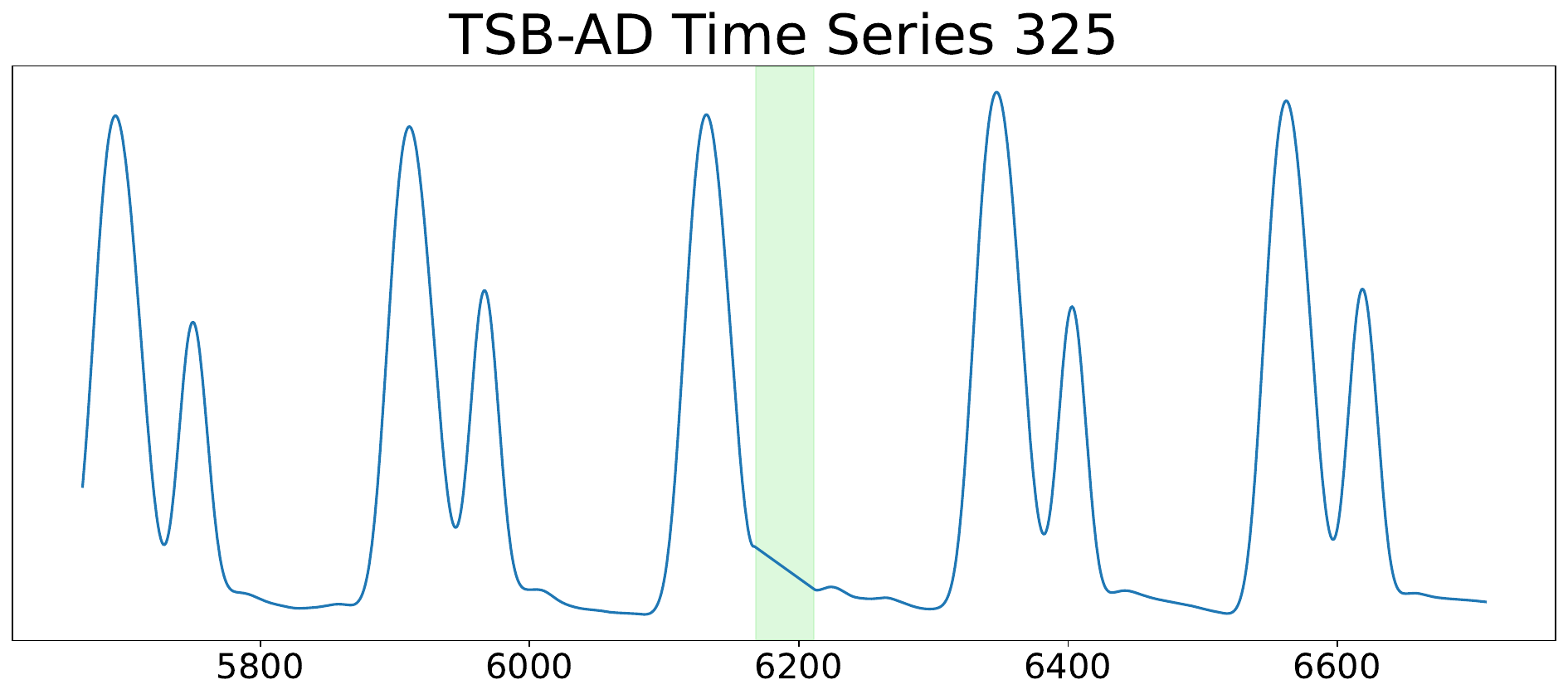}}
    \subfigure[Drop]{\label{fig:tsb367}\includegraphics[width=0.49\textwidth]{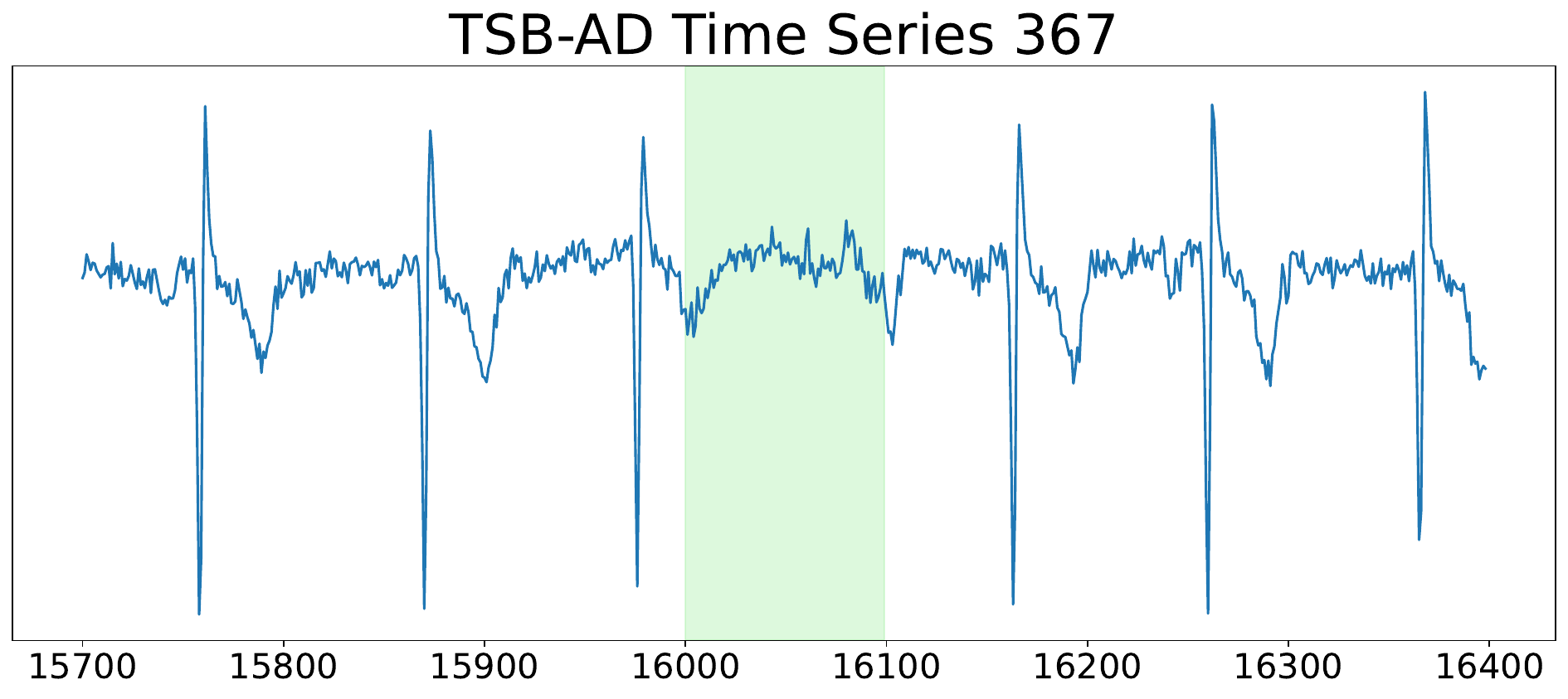}}
    \caption{
    Two examples of the \textit{missing peak or drop} anomaly type.
    Anomalous regions have been highlighted.
    }
    \label{fig:missing}
\end{figure}

\pagebreak
\subsection{Noise}
In this category a section of the signal becomes more noisy or gets replaced with noise altogether.
Viewing these anomalies through the frequency-domain is quite helpful since noise often emerges as high-frequency.
This is why we can remove noise using a low-pass filter.

\begin{figure}[h]
    \centering
    \subfigure[Entire Cycle]{\label{fig:tsb321}\includegraphics[width=0.49\textwidth]{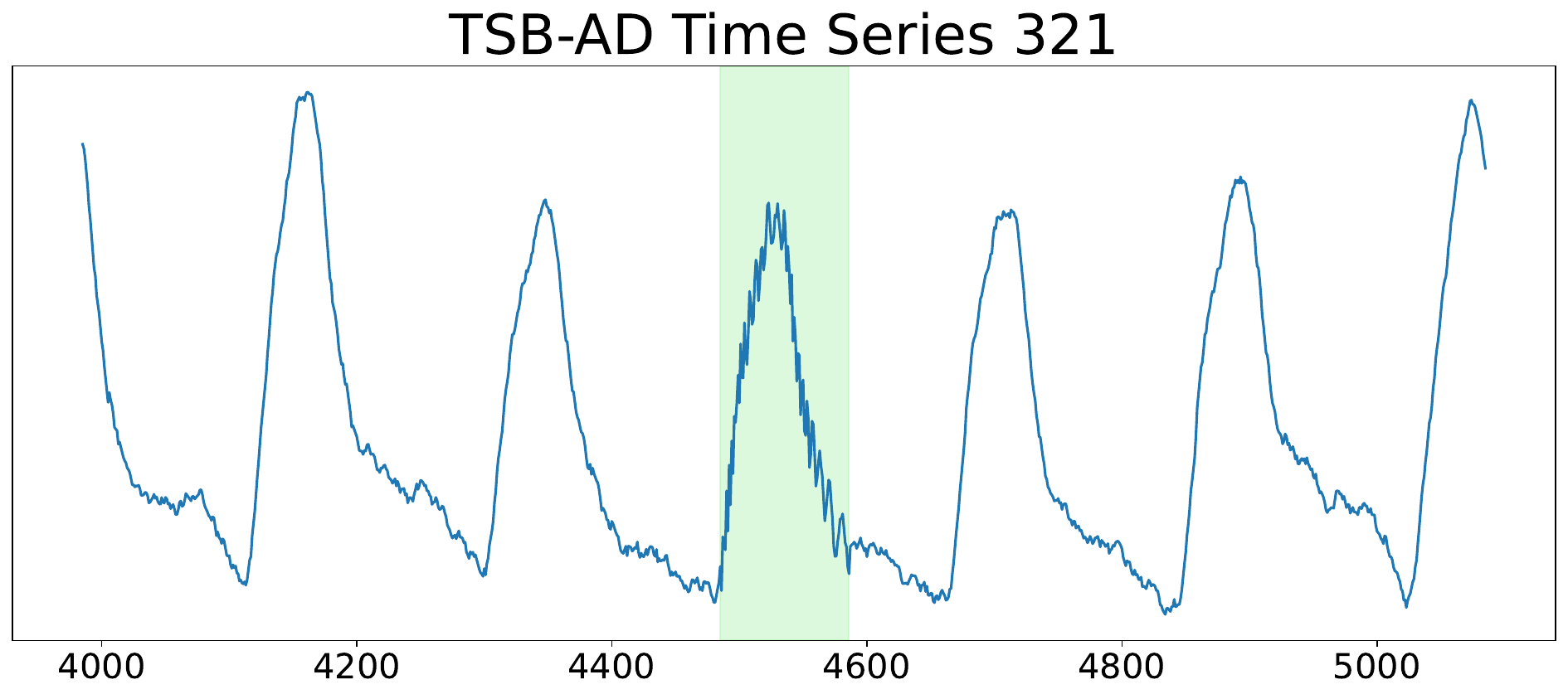}}
    \subfigure[Part of the Cycle]{\label{fig:tsb326}\includegraphics[width=0.49\textwidth]{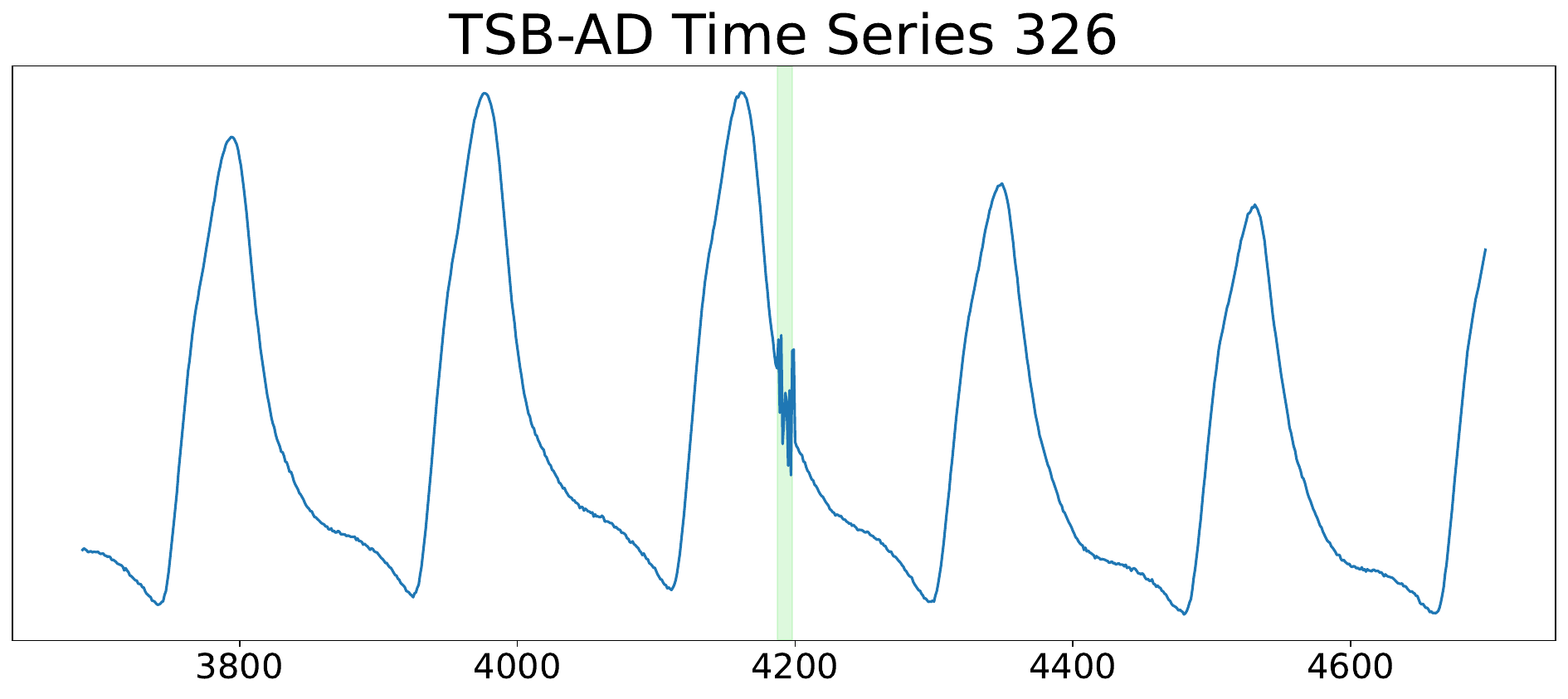}}
    \caption{
    Two examples of the \textit{noise} anomaly type.
    Anomalous regions have been highlighted.
    In \subref{fig:tsb321} some noise has been added to the entire cycle, whereas in \subref{fig:tsb326} a smaller portion of the cycle has been augmented with noise.
    }
    \label{fig:noise}
\end{figure}

\subsection{Outlier or Unusual Pattern}
In this category, one or more cycles get replaced with an unusual pattern.
\begin{figure}[h]
    \centering
    \subfigure[Completely  Different]{\label{fig:tsb312}\includegraphics[width=0.49\textwidth]{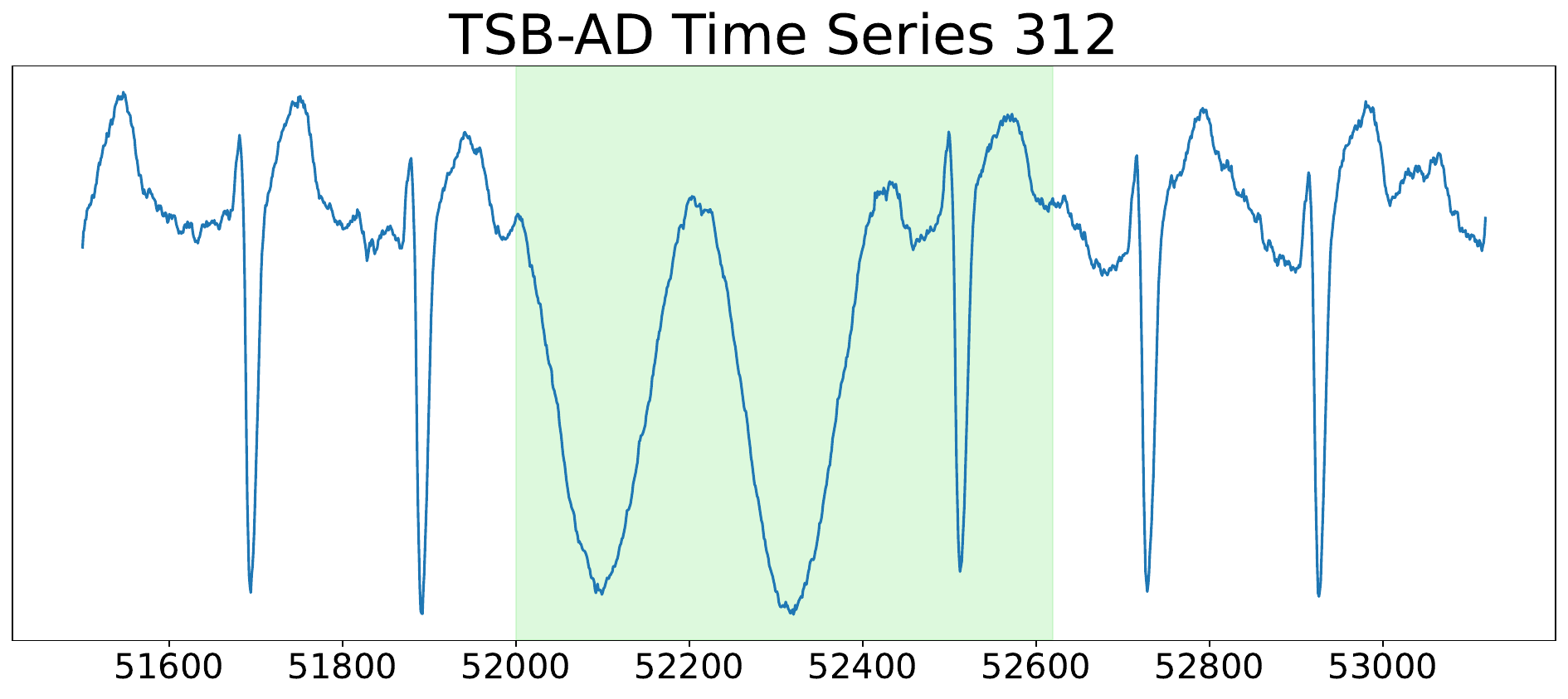}}
    \subfigure[Modified]{\label{fig:tsb317}\includegraphics[width=0.49\textwidth]{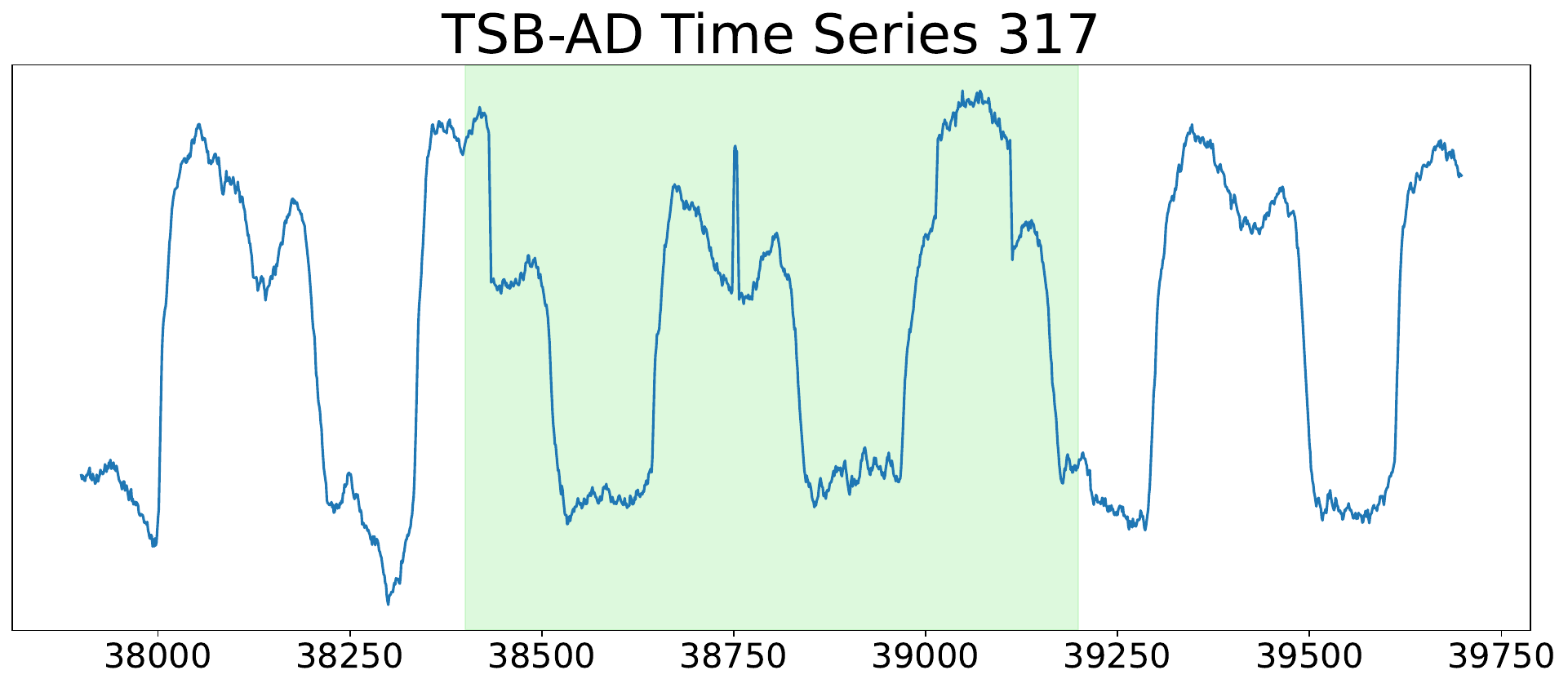}}
    \caption{
    Two examples of the \textit{outlier or unusual pattern} anomaly type.
    Anomalous regions have been highlighted.
    }
    \label{fig:outlier}
\end{figure}

\pagebreak
\subsection{Reversed Cycle}
This category of anomalies, a cycle in the signal gets reversed with respect to the cycle center.
In order to correctly classify these anomalies, the model needs to be able to reconstruct the original signal well so that the error attributed to the anomalous region stands out.

\begin{figure}[h]
    \centering
    \subfigure[]{\label{fig:tsb345}\includegraphics[width=0.49\textwidth]{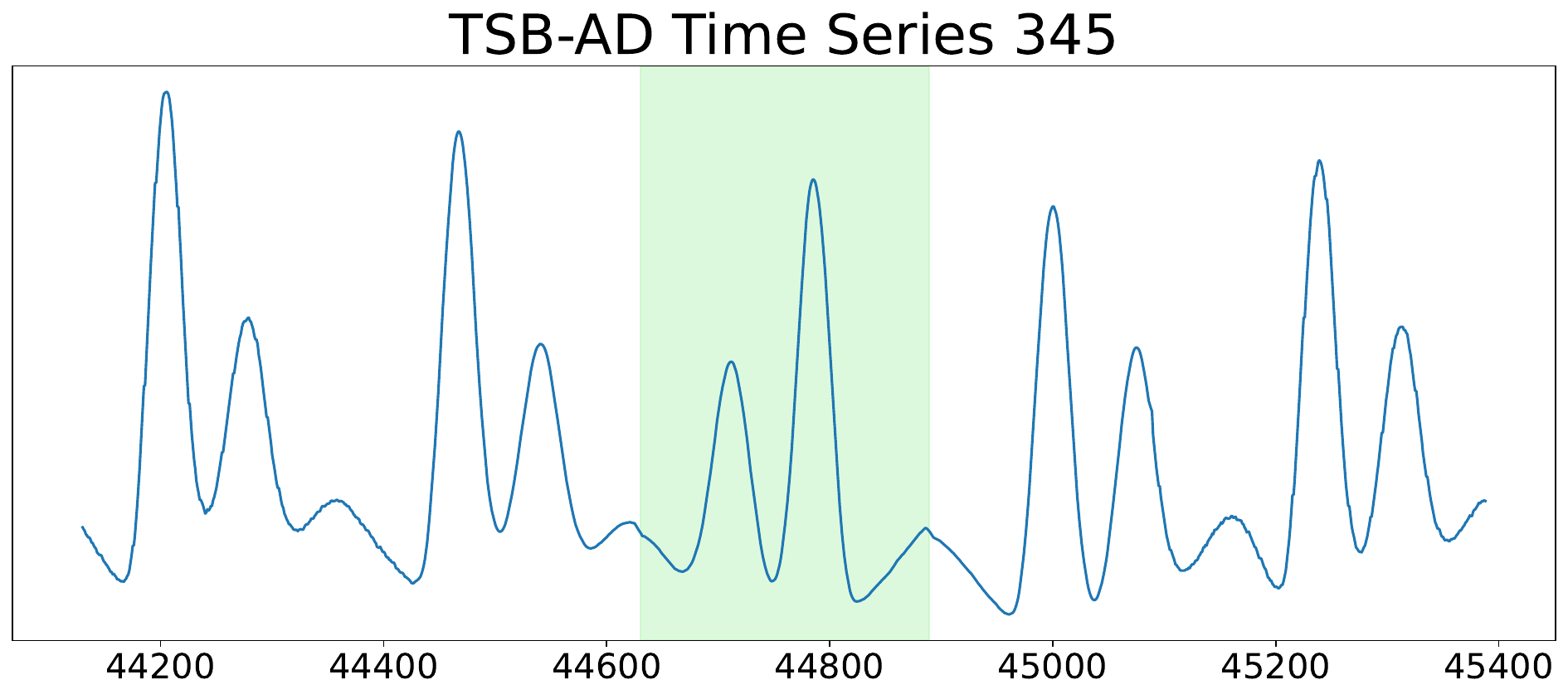}}
    \subfigure[]{\label{fig:tsb374}\includegraphics[width=0.49\textwidth]{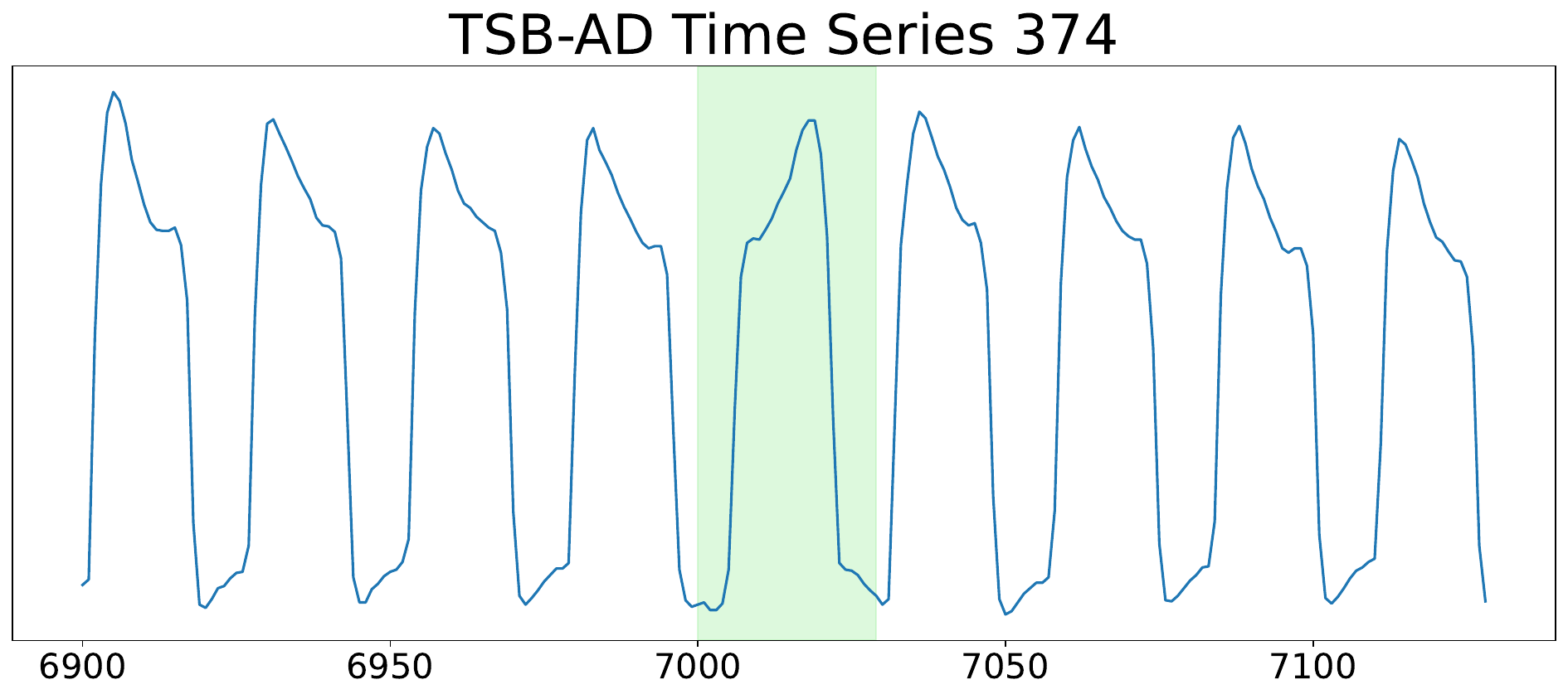}}
    \caption{
    Two examples of the \textit{reversed cycle} anomaly type.
    Anomalous regions have been highlighted.
    }
    \label{fig:reversed}
\end{figure}

\subsection{Time Shift or Time Warping}
Time shift refers to when the pause between two peaks or drops gets increased or decreased.
Time warping refers to when the cycle peak gets moved but the cycle length remains unaltered.

\begin{figure}[h]
    \centering
    \subfigure[Time  Shift]{\label{fig:tsb327}\includegraphics[width=0.49\textwidth]{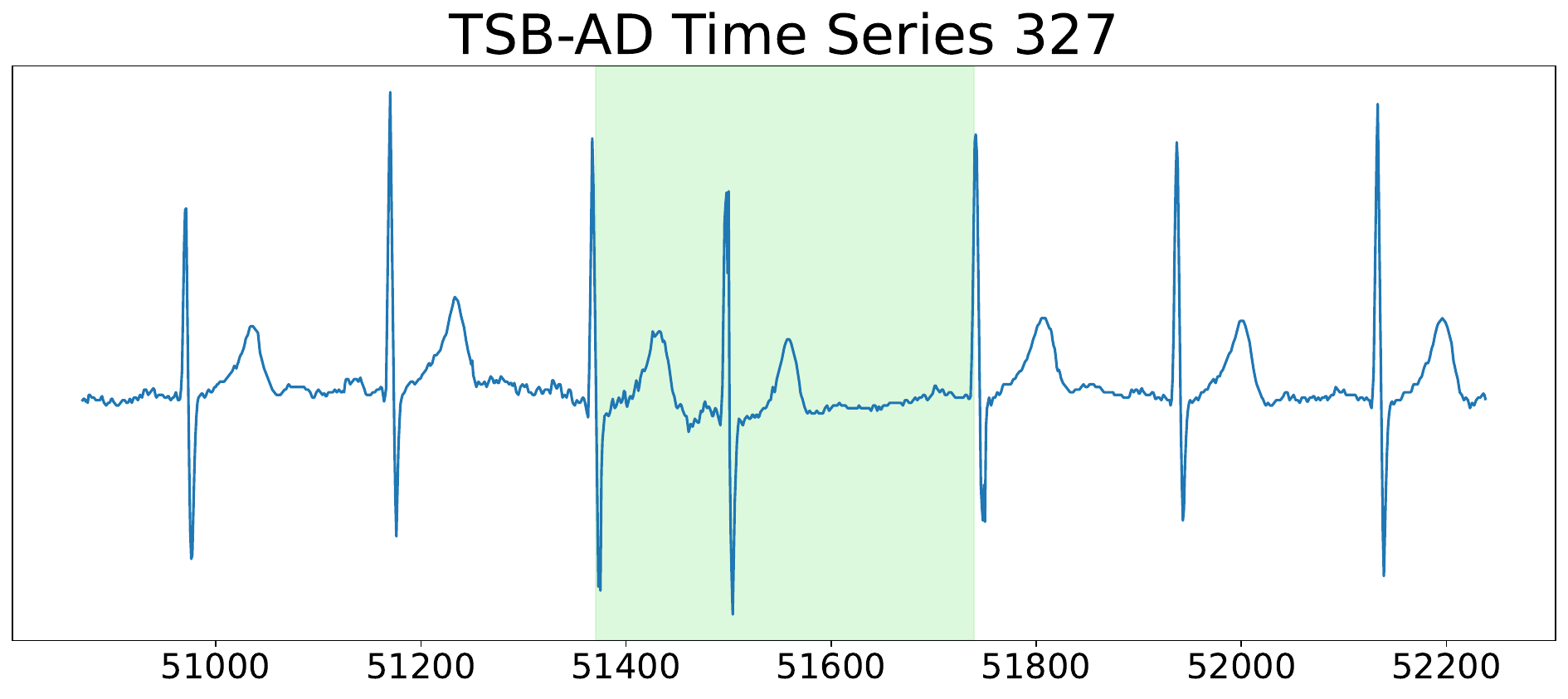}}
    \subfigure[Time Warping]{\label{fig:tsb421}\includegraphics[width=0.49\textwidth]{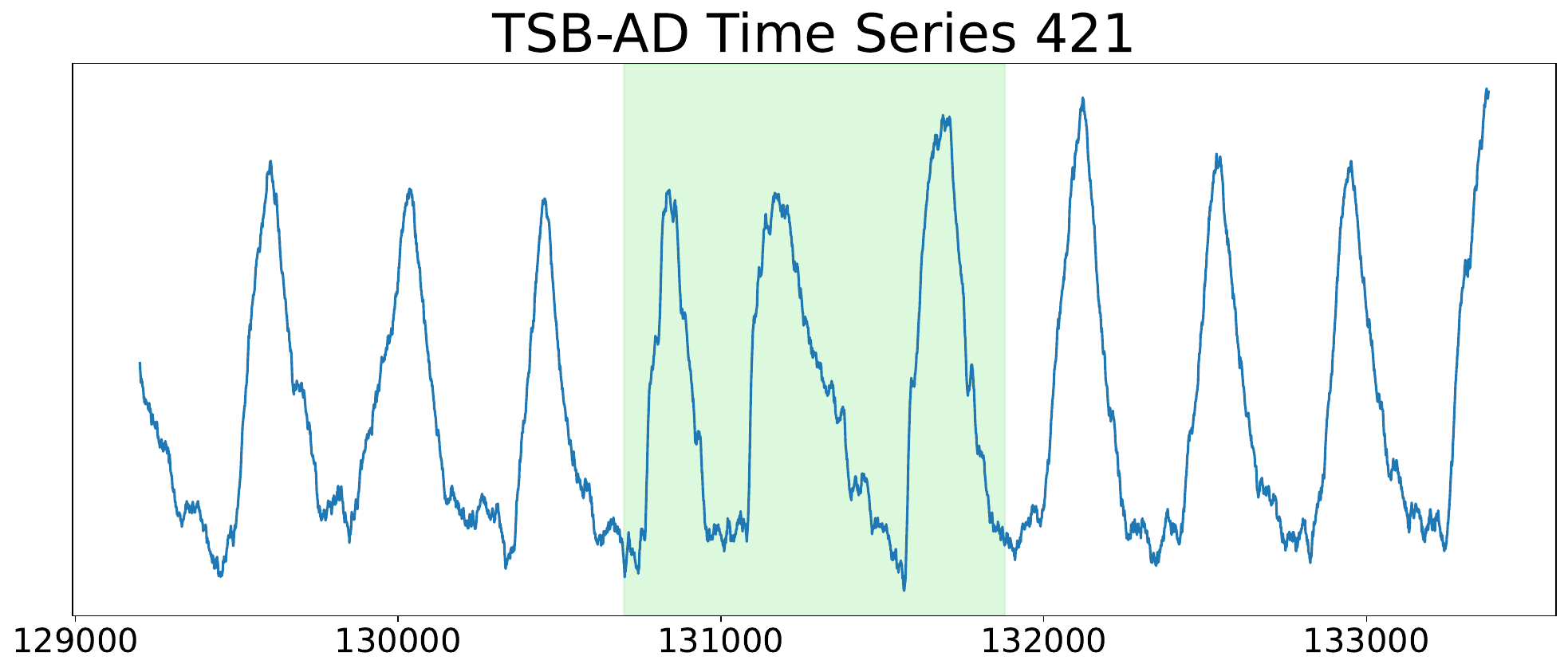}}
    \caption{
    Two examples of the \textit{time shift or time warping} anomaly type.
    Anomalous regions have been highlighted.
    }
    \label{fig:time}
\end{figure}

    \clearpage
    \section{Adaptive Window Sizes}\label{sec:adaptive-window}

\subsection{Motivation and Design Requirements}

The objective of the adaptive window size selection procedure is not merely to identify an optimal window length, but rather to determine whether a given temporal context is sufficient for the Fourier AutoEncoder (FAE) to meaningfully capture the underlying periodic structure of the time series.

In particular, the effectiveness of FAE strongly depends on the relationship between the selected window size and the dominant seasonal length of the signal. 
IF the input window is significantly shorter than the true seasonality period, the frequency-domain representation becomes incomplete, preventing the model from learning a stable spectral structure.

For example, when a time series exhibits a seasonal pattern with a period of 1000 time steps, using a window size of 128 does not provide adequate coverage of even a single cycle. 
As a result, FAE fails to observe a consistent frequency signature, and the reconstruction loss does not exhibit meaningful early convergence behavior.

In contrast, when the window size approaches or exceeds the dominant seasonal scale (e.g., one, one-and-a-half, or two full seasonal cycles), the spectral structure becomes stable. 
Under such conditions, FAE consistently demonstrates a rapid initial decrease in reconstruction loss, indicating that the underlying periodic pattern has been successfully captured.

To validate this behavior, we conduct controlled experiments on synthetic sinusoidal time series with known periodic lengths. 
These experiments confirm that the proxy learning dynamics of FAE reliably reflect the adequacy of the selected temporal context, providing a practical mechanism for inferring suitable window sizes in the presence of long-range seasonal structures.

These observations motivate the design of an adaptive procedure that explicitly probes the adequacy of different temporal contexts before model training.

\subsection{Adaptive Window Selection Algorithms}

Several key considerations underlie the proposed adaptive window size selection procedure. 
First, the training budget of the proxy Fourier AutoEncoder is intentionally kept limited and fixed across all candidate window sizes. 
This stage is not intended to achieve full convergence, but rather to probe the early learning dynamics of the model in a controlled and comparable manner.




The proposed adaptive window selection procedure is summarized in Algorithm~\ref{alg:adaptive_window}. Starting from a maximum candidate window size, we progressively reduce the window length and train a standalone Fourier AutoEncoder for a limited proxy training budget at each candidate size. We then measure the relative reduction in reconstruction loss during early training. A candidate window is considered suitable when its relative loss improvement exceeds a predefined threshold $\delta$. The procedure retains the smallest candidate satisfying this criterion, thereby seeking a temporal context that supports effective reconstruction without unnecessarily increasing the window size.

\begin{algorithm}[t]
\caption{Adaptive Window Selection via Proxy Fourier AutoEncoder}
\label{alg:adaptive_window}
\begin{algorithmic}[1]
\STATE \textbf{Input:} Time series $x$, minimum window size
$w_{\min}$, maximum window size $w_{\max}$,
loss tolerance $\delta$, proxy training budget $U$
\STATE \textbf{Output:} Selected window size $w^*$

\STATE $w \leftarrow w_{\max}$
\STATE $w^* \leftarrow w_{\max}$

\WHILE{$w \geq w_{\min}$}
    \STATE Construct sliding windows of size $w$ from $x$
    \STATE Initialize a standalone Fourier AutoEncoder
    \FOR{$u = 1, \ldots, U$}
        \STATE Perform one optimization step
        \STATE Record reconstruction loss $\mathcal{L}_u(w)$
    \ENDFOR

    \STATE Compute relative loss improvement:
    \STATE $\displaystyle
    \Delta(w) \leftarrow
    \frac{\mathcal{L}_1(w) -
    \min_{1 \leq u \leq U}\mathcal{L}_u(w)}
    {\mathcal{L}_1(w) + \epsilon}$

    \IF{$\Delta(w) \geq \delta$}
        \STATE $w^* \leftarrow w$
    \ENDIF

    \STATE $w \leftarrow \lfloor w/2 \rfloor$
\ENDWHILE

\STATE \textbf{Return} $w^*$
\end{algorithmic}
\end{algorithm}

\subsection{Illustrative Example}

\begin{figure}[h]
    \centering
    \includegraphics[width=\linewidth]{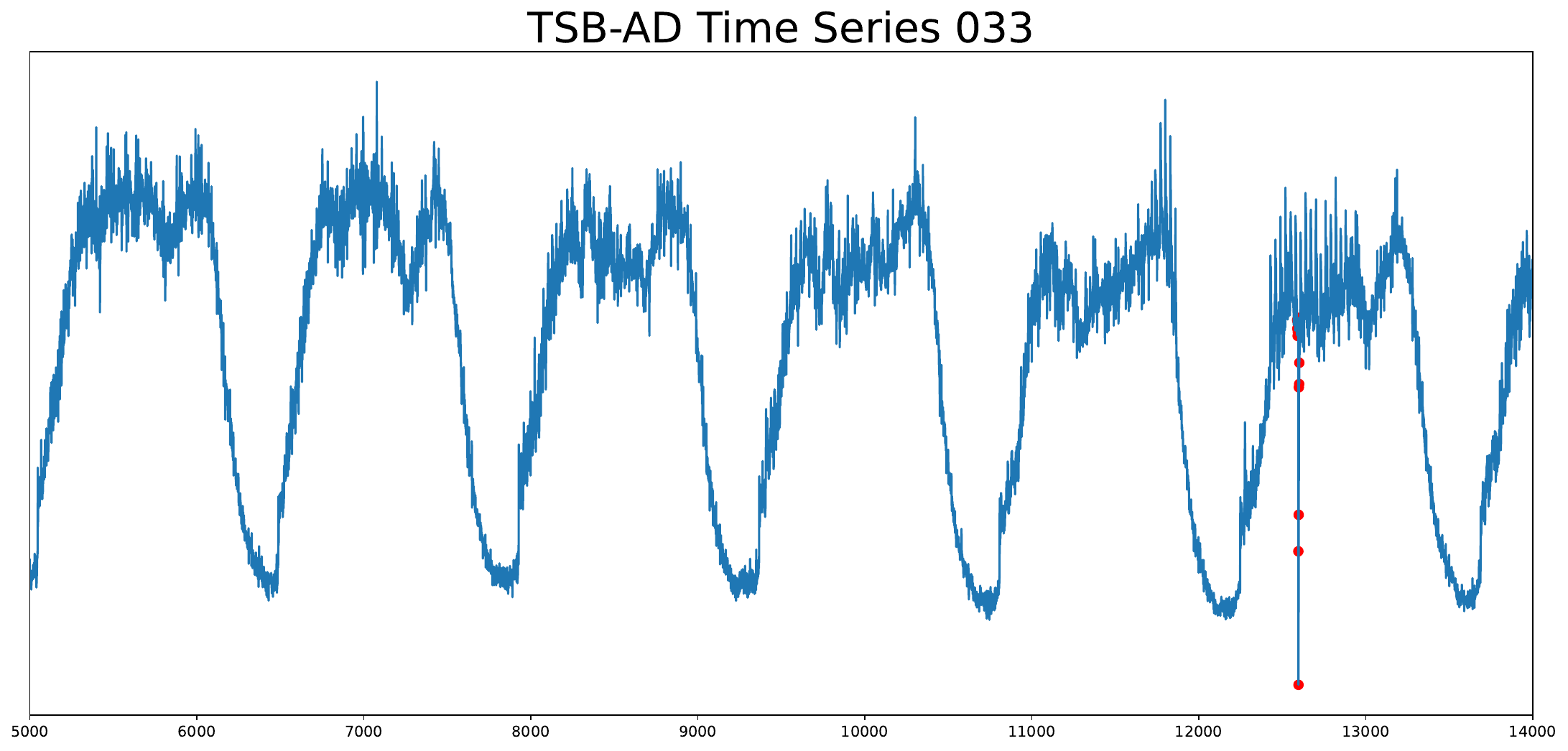}
    \caption{
    Illustrative example of the adaptive window size selection mechanism.
    A time-series segment between approximately time steps 5000 and 14000 is shown, within which multiple candidate window sizes are evaluated.
    Approximately six seasonal cycles are observed over this interval, corresponding to a dominant seasonal period of 1517.
    The proposed method selects a window length of 3034, which is approximately two times the underlying seasonal scale, demonstrating its ability to identify appropriate temporal context for frequency-domain modeling.
    }
    \label{fig:adaptive_window}
\end{figure}

To further illustrate this behavior, Figure~\ref{fig:adaptive_window} presents a representative example of the adaptive window size selection process.
The figure shows a segment of a time series spanning approximately from time step 5000 to 14000, within which multiple candidate window sizes are visualized.

Within this interval, approximately six repeating seasonal cycles can be observed over a span of about 9000 time steps, suggesting an underlying seasonal period of roughly 1500.
Based on this structure, the proposed method selects a window size of 3034, corresponding to approximately two multiples of the dominant seasonal length.
A more precise analysis of the signal confirms that the true seasonal period is 1517, indicating that the selected window size closely aligns with the intrinsic temporal scale of the series.

Such behavior demonstrates that the adaptive procedure does not arbitrarily favor a fixed window length, but instead identifies temporal contexts that are consistent with meaningful repetitions of the underlying seasonal structure.
This observation is consistent with our sinusoidal validation experiments and further supports the ability of the proxy FAE learning dynamics to reflect long-range periodic patterns in real-world time series.

    \clearpage
    \section{Related Work}\label{sec:related}
Traditional TSAD methods fall into three categories: distance-based, density-based, and prediction-based approaches \citep{Review}. 
Among these, prediction-based methods have gained prominence with the rise of deep learning \citep{Review, EncDec-AD, USAD, CAE-M, MAD-GAN, TranAD, DAGMM}. These methods aim to learn normal temporal behavior from training data and identify anomalies as deviations from this learned pattern. 
They are typically divided into forecasting-based and reconstruction-based approaches \citep{Review}.

\subsection{Reconstruction-Based Methods}
Reconstruction-based methods rely on encoding time series subsequences into a compact latent representation and reconstructing the original sequence from this encoding. 
The reconstruction error is used to flag potential anomalies. 
These models are usually trained on normal data, under the assumption that anomalies will be harder to reconstruct and thus exhibit higher reconstruction error.

AutoEncoders (AEs) have become especially popular for this task due to their ability to learn efficient data representations. 
EncDec-AD \citep{EncDec-AD} introduced an LSTM-based encoder-decoder architecture that uses reconstruction error to detect anomalies. 
Later extensions, such as LSTM-VAE \citep{LSTM-VAE}, combined the representational power of Variational AutoEncoders (VAEs) \citep{kingma2013auto} with LSTM \citep{LSTM} networks to better model temporal dependencies.

Despite their effectiveness, AEs face several well-documented limitations in TSAD. 
First, they are susceptible to learning a trivial identity mapping \(F(x) = x\), especially when overparameterized. 
This issue can be diagnosed by their ability to reconstruct noise and is typically mitigated by constraining the latent dimensionality. 
Second, AEs may fail to detect anomalies that do not induce a sufficiently high reconstruction error. 
This relates to the concept of group robustness: while AEs are sensitive to strong outliers, they often miss more subtle deviations such as trend shifts or pattern inversions, an issue exacerbated by large window sizes. 
Third, although AEs allow point-wise anomaly scoring, windows containing multiple anomalies can result in diffuse or misleading anomaly scores across the window. 
Although this may be seen as a limitation, it is still preferable to methods like Generative Adversarial Networks (GANs) \citep{goodfellow2020generative}, which typically lack such granularity altogether.
While all generative models including GANs can be used to derive point-wise anomaly scores, the canonical GAN architecture only utilizes a single discriminator network which provides a total anomaly score on the entire window.

Beyond time-domain approaches, the Fourier Transform has been widely used to analyze time series data in the frequency-domain \citep{bloomfield2004fourier}. 
This transformation unveils periodic structures, hidden oscillations, and energy distributions that may be difficult to detect otherwise \citep{wrinch2012anomaly, shi2010real, lappas2021fourier, yokkampon2020AutoEncoder}. 
Integrating frequency-domain analysis into deep learning models has proven effective for enhancing anomaly detection. 
For instance, the F-SE-LSTM model \citep{lu2024f} combines the Fast Fourier Transform (FFT) with Squeeze-and-Excitation (SE) modules \citep{hu2018squeeze} and LSTMs to extract frequency-aware features across temporal windows. 
Such hybrid models demonstrate the benefits of incorporating global frequency patterns alongside local temporal modeling.

Recent TSAD methods have also explored frequency-aware and multi-scale modeling.
\citet{chen2024learning} learn multiple normal patterns in the frequency domain, while \citet{nam2024breaking} combine time- and frequency-domain information at different granularities.
\citet{zhong2025multiresolution} use multi-resolution decomposition to address nonstationary time series.
Other approaches include association-based anomaly detection \citep{xu2022anomaly} and adaptive bottlenecks with adversarial decoders \citep{shentu2025towards}.
These studies establish that frequency representations, decomposition, and the combination of complementary temporal views are not, by themselves, new to TSAD. 
FourStat studies a specific univariate reconstruction design: a long-window Fourier AutoEncoder models broader structure, while a short-window Stationary
AutoEncoder contributes to the reconstruction of the final portion of each window.

\subsection{Forecasting-Based Methods}
Related approaches to long-range temporal modeling have been developed for forecasting.
Informer uses sparse attention to process long sequences \citep{zhou2021informer}; Autoformer combines progressive decomposition with periodicity-based Auto-Correlation \citep{wu2021autoformer}; and FEDformer incorporates Fourier or wavelet operations into a decomposition-based Transformer \citep{zhou2022fedformer}.
PSformer uses segment attention and parameter sharing \citep{wang2025psformer}, whereas the Temporal Fusion Transformer combines recurrent local processing with attention over longer dependencies \citep{lim2020tft}.
These models were originally designed to predict future values rather than produce anomaly scores. We evaluate their adaptations to univariate anomaly detection using a common data and evaluation protocol, while distinguishing their forecasting objectives from FourStat's reconstruction objective.

\subsection{Evaluation}
Evaluation choices also affect comparisons between TSAD methods.
Prior work has documented problems with commonly used benchmarks and metrics \citep{sarfraz2024position,TSB}.
Our Perfect Point metric complements continuous, threshold-independent metrics: it uses ground-truth labels to determine whether \emph{some} threshold would attain complete event recall with no point-wise false positives. It does not select a threshold for deployment.

The proposed Perfect Point (PP) metric still needs real labels. 
This is impossible to get the best threshold for Perfect Point (PP) with labels in real applications~\citep{sarfraz2024position}.


\end{document}